%% file: main.tex
\documentclass{article}

\PassOptionsToPackage{numbers,compress}{natbib}
\usepackage{synmandex}

\usepackage[utf8]{inputenc}
\usepackage[T1]{fontenc}
\usepackage{graphicx}
\usepackage{float}
\usepackage{booktabs}
\usepackage{amsmath}
\usepackage{amssymb}
\usepackage{mathtools}
\usepackage{multirow}
\usepackage{colortbl}
\usepackage{diagbox}
\usepackage{algorithm}
\usepackage{algorithmic}
\usepackage{xcolor}
\usepackage[skins]{tcolorbox}
\usepackage{url}
\usepackage{hyperref}
\usepackage[capitalize,noabbrev]{cleveref}

\definecolor{synblue}{HTML}{0072B2}
\definecolor{manorange}{HTML}{D55E00}
\definecolor{dexteal}{HTML}{009E73}
\definecolor{synfill}{HTML}{EAF3F8}
\definecolor{manfill}{HTML}{FCF1E8}
\definecolor{dexfill}{HTML}{E9F7F2}
\definecolor{proposefill}{HTML}{FFF7D6}
\definecolor{proposeborder}{HTML}{D8A63A}

\newcommand{\method}{SynManDex}

\newcommand{\qbi}{\mathcal{Q}_{\text{bi}}}
\newcommand{\qsingle}{\mathcal{Q}}
\DeclareMathOperator{\logm}{Log}
\DeclareMathOperator{\FK}{FK}

\definecolor{coverink}{HTML}{18212B}
\definecolor{covermuted}{HTML}{4F5E6D}
\definecolor{covertheme}{HTML}{26323D}
\definecolor{coverrule}{HTML}{D4DADF}
\definecolor{coverabstract}{HTML}{F6F8FA}
\definecolor{coverabstractframe}{HTML}{D5DDE4}
\definecolor{coverauthor}{HTML}{334553}
\newcommand{\coverabstracttext}{
Human hand-object interactions encode functional intent, but direct transfer to robotic hands often fails under morphology, contact, and reachability constraints. We present SynManDex, a synthetic pipeline that uses generated human pre-grasps as affordance-aware proposals and resolves the final contacts with robot-native optimization. SynManDex samples object-conditioned digital human pre-grasps, retargets them to dexterous robotic hand poses, optimizes force-closure contacts on the target embodiment, and admits trajectories that pass checks from each step. The resulting keyframes support both grasp-and-lift demonstrations and various prehensile manipulation tasks such as tea pouring, photo taking, and flute playing, designed via VLM agents. As a result, SynManDex combines high grasp quality (86.4\% grasp stability) with 4.67/5 human-likeness (93.4\%). It achieves 80.7\% successes in simulation and 25/30 (83.3\%) real-robot successes when applied to a 36-DOF bimanual dexterous robotic platform.}
\newcommand{\coverauthorsfirst}{
Yanming Shao$^{1}$ \quad
Zanxin Chen$^{1,2}$ \quad
Wenwei Lin$^{3}$ \quad
Mingjie Zhou$^{4}$
}

\newcommand{\coverauthorssecond}{
Tianxing Chen$^{5}$ \quad
Xiaokang Yang$^{2}$ \quad
Yichen Chi$^{6,\dagger}$ \quad
Yao Mu$^{2,\dagger}$
}
\newcommand{\coveraffiliations}{
$^{1}$Shanghai AI Lab \quad
$^{2}$Shanghai Jiaotong University \quad
$^{3}$Shenzhen University \quad
$^{4}$Fudan University \quad
$^{5}$University of Hong Kong \quad
$^{6}$ZTE Corporation
}
\newcommand{\coverauthornotes}{
$^\dagger$Corresponding authors.
}
\newcommand{\covertitlefont}{\fontfamily{ptm}\bfseries\fontsize{18.2}{21.2}\selectfont}

\begin{document}

\thispagestyle{empty}
\begin{center}
\vspace*{-0.58in}
{\covertitlefont
\textcolor{coverink}{SynManDex: Synthesizing Human-like Dexterous Grasps}\par
\vspace{0.006in}
\textcolor{coverink}{from Synthetic Human Pre-Grasps}\par}
\vspace{0.062in}
{\sffamily\fontsize{8.15}{9.8}\selectfont\textcolor{coverauthor}{\coverauthorsfirst}\par}
\vspace{0.032in}
{\sffamily\fontsize{8.45}{10.1}\selectfont\textcolor{coverauthor}{\coverauthorssecond}\par}
\vspace{0.020in}
{\sffamily\fontsize{7.2}{8.5}\selectfont\textcolor{covermuted}{\makebox[\linewidth][c]{\coveraffiliations}}\par}
\vspace{0.014in}
{\sffamily\tiny\textcolor{covermuted}{\coverauthornotes \quad Project page: \url{https://tsunami-kun.github.io/SynManDex/}}\par}
\vspace{0.030in}
{\color{coverrule}\rule{0.66\linewidth}{0.8pt}}
\end{center}

\vspace{0.015in}
\begin{tcolorbox}[
  enhanced,
  width=\linewidth,
  colback=coverabstract,
  colframe=coverabstractframe,
  boxrule=0.7pt,
  arc=5pt,
  outer arc=5pt,
  boxsep=0pt,
  left=10pt,
  right=10pt,
  top=5pt,
  bottom=5pt]
  {\sffamily\footnotesize\bfseries\textcolor{covertheme}{Abstract}}\par\vspace{0.08em}
  {\footnotesize\linespread{0.96}\selectfont\textcolor{covermuted}{\coverabstracttext}}
\end{tcolorbox}

\vspace{0.045in}
\begin{center}
  \includegraphics[width=\linewidth]{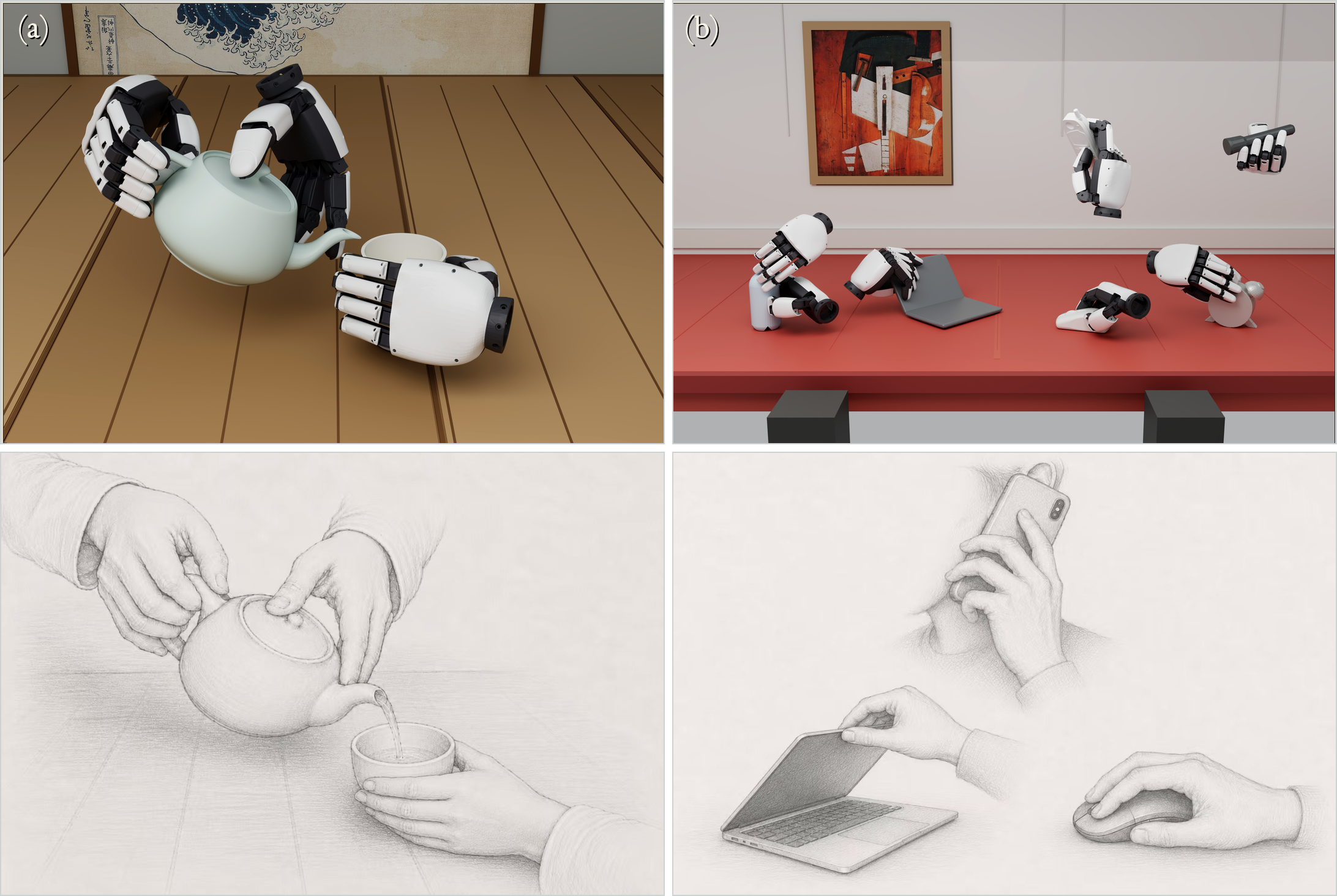}
\end{center}
\vspace{-0.01in}
\captionof{figure}{\textbf{Human-prior-guided bimanual dexterous manipulation generated by \method{}.} Top: accepted XHand grasp keyframes for (a) collaborative tea pouring and (b) office-object interactions, including bottle opening, laptop use, phone holding, mouse use, alarm-clock grasping, and flashlight holding. Bottom: human sketches illustrate the functional hand-object affordances that the robot grasps aim to preserve. This teaser is qualitative; physical, policy, and hardware results are evaluated in \cref{sec:experiments}.}
\label{fig:teaser}
\clearpage

\section{Introduction}
\label{sec:introduction}

Humans develop dexterity from an early age, yet endowing robots with such capabilities remains a fundamental challenge. Multi-fingered hands provide a natural embodiment for general-purpose robots: within a single morphology, they can wrap, pinch, and encircle everyday objects to support diverse functional uses. The core difficulty, however, lies in coordinating fingers with object affordances in service of task intent, rather than merely forming stable contacts. Human-level dexterity therefore requires both semantic awareness of where and how an object should be used and fine-grained physical validity under the kinematic and contact constraints of a high-DOF hand. Unlike parallel-jaw grasping, which often admits compact geometric simplifications, dexterous grasp synthesis must reason over embodiment-specific finger coordination and non-smooth contact dynamics~\cite{billard2019trends,andrychowicz2020learning}.

Existing methods typically trade off physical validity against functional intent. Physics-first grasp synthesis grounds contacts through optimization-based objectives, but often remains brittle for high-DOF hands and lacks human-semantic awareness~\cite{wang2022dexgraspnet,xu2023unidexgrasp,chen2025bodex,zurbrugg2025graspqp}. Conversely, semantic-first approaches, including those based on vision-language affordances~\cite{tang2025affordgrasp,wei2025afforddexgrasp}, can capture task intent but are usually too coarse to specify physically precise finger-level contacts. Grasp taxonomies~\cite{feix2015grasp, chen2025dexonomy} offer knowledge-based templates, yet they do not directly align those templates with object-specific affordances. Retargeting methods attempt to bridge this gap by transferring collected human hand data to robotic hands; however, direct mappings often suffer from severe embodiment mismatch, leading to mis-contact, interpenetration, or physically invalid grasps~\cite{qin2022dexmv,handa2020dexpilot,mandikal2022dexvip,zhao2024dexh2r,mu2026deximit}. Learning-based retargeting further depends heavily on specific hand embodiments and trajectory distributions~\cite{mandi2025dexmachina,li2025maniptrans}. What is missing, therefore, is a unified interface that preserves human functional intent while delegating physical validity to the target robotic embodiment.

To bridge this gap, we introduce a human-to-robot grasp synthesis framework that separates functional proposal generation from robot-specific physical grounding. Rather than directly transferring human grasps to robotic hands, we first use diffusion-based digital human models trained on large-scale hand-object interaction data~\cite{romero2017embodied,taheri2020grab,christen2024diffh2o,taheri2022goal} to generate object-conditioned human pre-grasps. These pre-grasps are used as pre-contact priors, providing affordance-aware approach directions, wrist orientations, and finger coordination patterns, without determining the final robot contact state. The physically valid grasp is instead resolved by a force-closure-based optimization pipeline defined directly on the target robotic embodiment.

\begin{figure}[H]
    \centering
    \includegraphics[width=\textwidth]{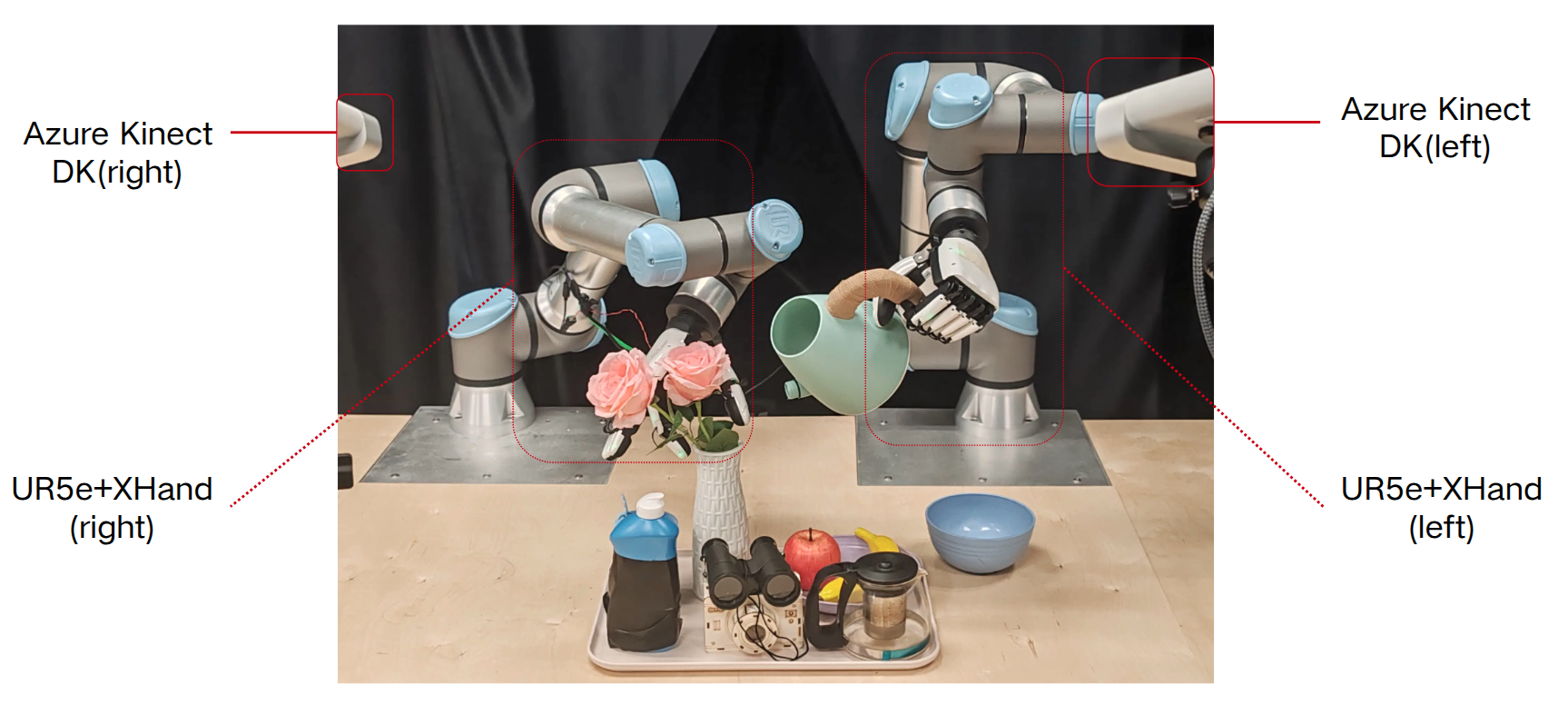}
    \caption{\textbf{Real-world robot platform.}
    The bimanual system operates over a tabletop workspace observed by two Azure Kinect cameras. The fused point cloud and robot proprioception form the same policy interface used in simulation.}
    \label{fig:hardware_platform}
\end{figure}

\begin{figure}[H]
    \centering
    \includegraphics[width=\textwidth]{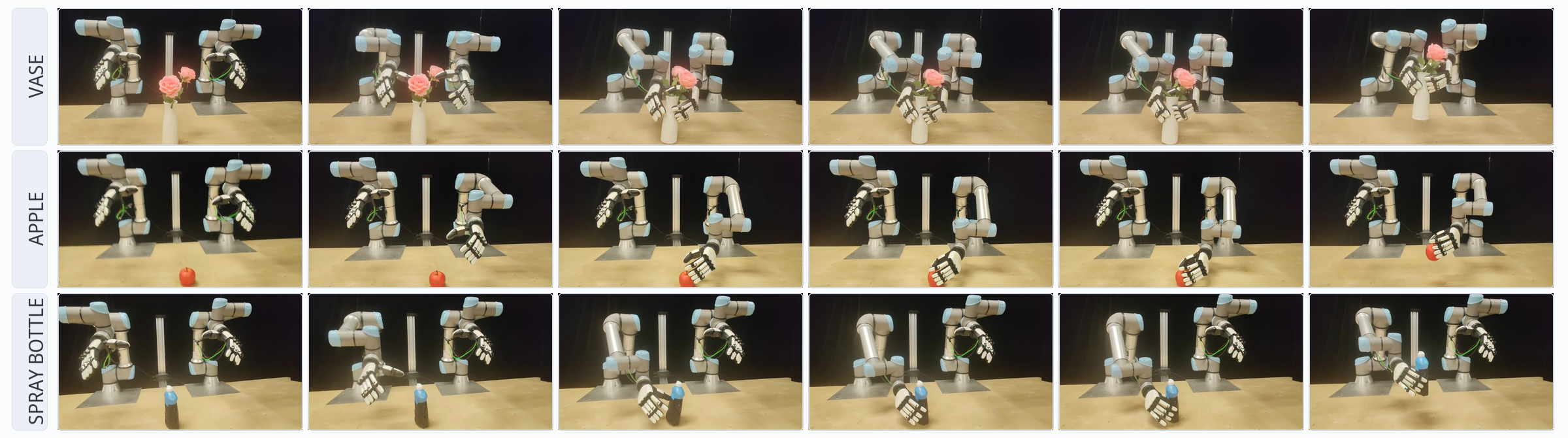}
    \caption{\textbf{Real-world grasping.} Rows show successful vase, apple, and spray-bottle trials from the three-object tabletop benchmark; columns are time-ordered execution frames. The corresponding 30-trial success rates are reported in \cref{tab:hardware}.}
    \label{fig:hardware_benchmark}
\end{figure}

We instantiate this framework as \textbf{\method{}}, a three-stage pipeline for bimanual dexterous prehensile manipulation (\cref{fig:pipeline}). First, an object-conditioned diffusion model generates digital human pre-grasps that encode functional human-object intent. Second, a robot-native optimizer retargets and refines these proposals into force-closure grasp keyframes with low penetration. Third, a task-specific planner admits grasp keyframes into executable arm-hand manipulation trajectories. Across 312 objects from 25 classes, \method{} achieves 86.4\% force-closure success with maximal penetration of 0.6 mm, together with a 4.67/5 human-likeness score. Policies trained only on the generated synthetic demonstrations achieve 80.7\% simulated zero-shot success and 25/30 successes on a tabletop manipulation real-world benchmark. We further validate the decomposition through grasp-quality analysis (\cref{tab:grasp_quality}), policy-data ablations (\cref{tab:policy_ablation}), object-centric transition, and real-system experiment (\cref{fig:hardware_platform,fig:hardware_benchmark}).

\paragraph{Contributions.}
\begin{itemize}
\setlength{\itemsep}{0.18em}
\setlength{\parsep}{0pt}
\setlength{\topsep}{0.18em}
\item We propose a staged human-prior-to-robot formulation for dexterous grasp synthesis, where generated digital human pre-grasps serve as affordance-aware pre-contact proposals rather than directly executable robot demonstrations.

\item We instantiate this formulation as \method{}, a pipeline that couples object-conditioned pre-grasp synthesis, human-to-robot retargeting, force-closure optimization, and VLM-based task description to produce grasp keyframes and manipulation trajectories.

\item We conduct a comprehensive grasp evaluation across physical feasibility, embodiment grounding, human-likeness, and downstream policy learning, showing that human-prior proposals improve robot-native grasp synthesis without being treated as final contact labels.

\item We extend the admitted grasp keyframes into diverse prehensile manipulation demonstrations, including grasp-and-lift, tea pouring, photo taking, and flute playing, illustrating how functional human-object priors can support diverse prehensile manipulation data beyond static grasp generation.
\end{itemize}

\section{Related Work}
\label{sec:related_work}

\subsection{Human Priors for Dexterous Manipulation}

Human hand-object data provides useful priors for hand pose, contact regions, and object affordances. HOI datasets are often parameterized with MANO~\cite{romero2017embodied}; GRAB~\cite{taheri2020grab}, ContactPose~\cite{brahmbhatt2020contactpose}, and DexYCB~\cite{chao2021dexycb} encode complementary views of hand pose, object geometry, and contact. The question for robot learning is how the collected human data should enter a robot pipeline. Behavior cloning and retargeting systems can imitate human hand behavior~\cite{qin2022dexmv,handa2020dexpilot,mandikal2022dexvip,zhao2024dexh2r}, but direct imitation is sensitive to distribution shift and embodiment mismatch. Functional retargeting methods specify object- or task-centric objectives~\cite{mandi2025dexmachina,li2025maniptrans}, yet they still require task rewards and may produce local motions that do not preserve the intended contact when the robot morphology cannot realize the human pose.

Generative HOI models synthesize diverse digital human interactions~\cite{christen2024diffh2o,zhang2024graspxl,taheri2022goal}. They can produce object-conditioned task semantics without collecting new robot data, but their outputs remain MANO-space motions rather than robotic trajectories. If those motions are retargeted directly, contacts can move, fingers can penetrate, and wrists can become unreachable. In contrast, \method{} places the human prior earlier by sampling pre-contact frames as semantic seeds and then obtain physically valid contacts via force-closure-based optimizations.

\subsection{Dexterous Grasp Synthesis}

Dexterous grasps are the keyframe of dexterous prehensile manipulation. They should satisfy both physical stability and semantic intents. Online reinforcement learning can discover dexterous behaviors~\cite{billard2019trends,andrychowicz2020learning,handa2023dextreme}, but contact rewards alone do not guarantee force closure or arm-hand reachability. Physics-aware grasp generators prioritize stability against gravity~\cite{wang2022dexgraspnet,chen2025bodex,zurbrugg2025graspqp}, and geometric retargeting improves correspondence across hands~\cite{yin2025geort}; however, neither directly chooses object-function contacts. Recent optimization systems and generative models scale data and improve pose diversity~\cite{yang2025ultradexgrasp,mu2026deximit,li2023gendexgrasp,zhang2024dexgraspnet2,xu2024dexgrasp_transformer,jiang2021grasptta}, but their are limited to stable grasping without semantics.

Vision-language and affordance-driven methods capture task intent via language instructions or open-world region matching~\cite{li2025dhagrasp,tang2025affordgrasp,wei2025afforddexgrasp}. Although such methods are useful for object-function reasoning, but they are usually too coarse to determine precise finger-level contacts. Taxonomy-based approaches provide human factor-oriented grasp templates~\cite{feix2015grasp, chen2025dexonomy}, but still require manually object-specific annotations. The missing interface is not a stronger prior alone or a stronger optimizer alone, but a staging that gives each one the right validity criterion. \method{} differs by placing the human model before contact and the robot optimizer after retargeting: the human model proposes a basin, and the robot optimizer tests contact, IK, and rollout feasibility. \Cref{tab:capability_stack} summarizes this positioning, and \Cref{app:extended_related_work} provides an expanded comparison.

\Cref{tab:capability_stack} separates capability strength from evidence type, highlighting the missing intersection between human/task priors and bimanual action data.

\input{table_capability}

\section{Method}
\label{sec:method}

\begin{figure}[htbp]
    \centering
    \includegraphics[width=\linewidth]{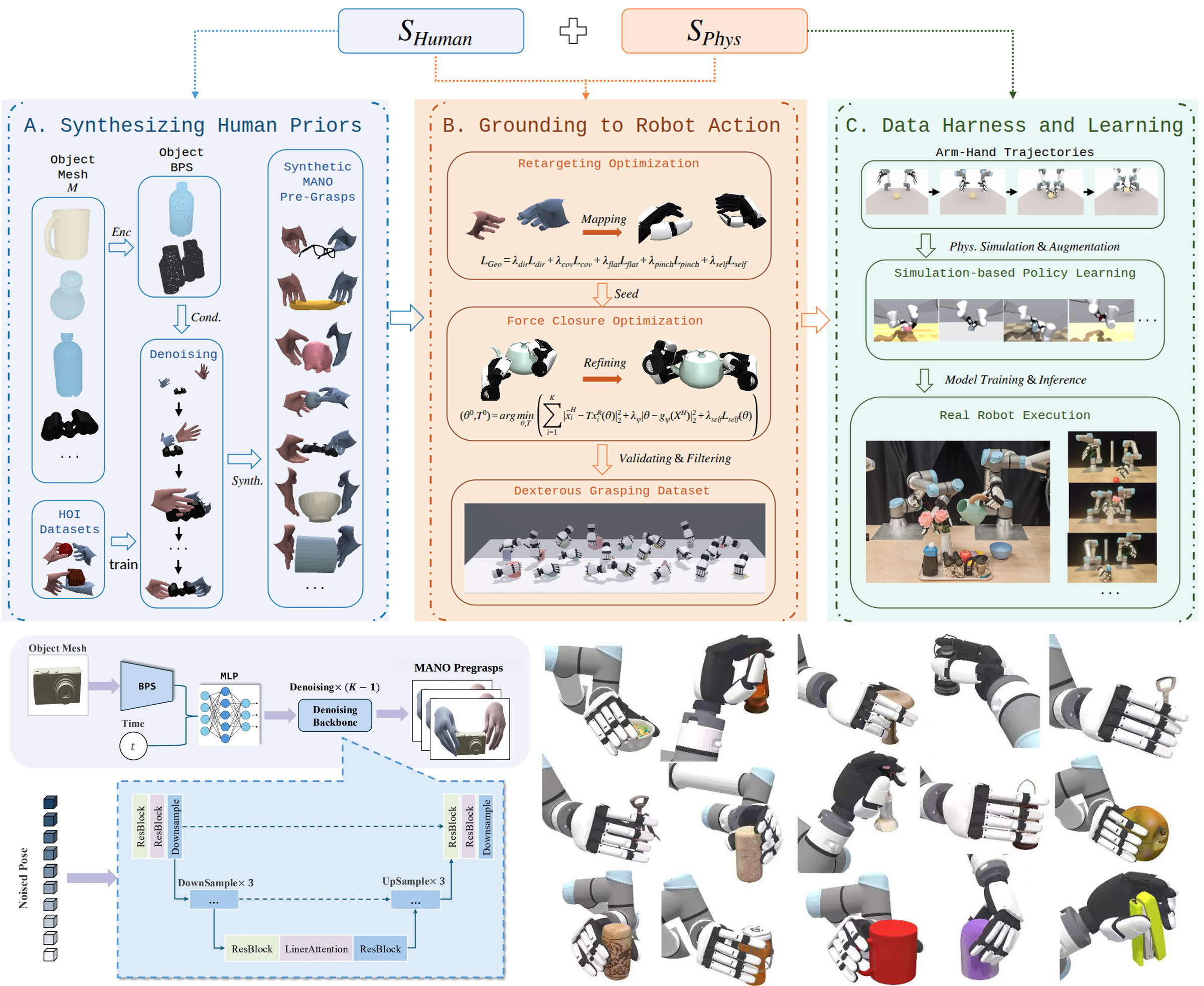}
    \caption{\textbf{\method{} pipeline.} (A) An object-conditioned diffusion model samples digital human pre-grasps as affordance-aware pre-contact proposals. (B) Geometric retargeting maps human hand keypoints to robotic hand seeds, and robot-native optimization refines contacts under collision and force-closure objectives. (C) After validations, we collect dexterous manipulation trajectories for policy learning and real-robot evaluation. The lower strip shows representative outputs from the corresponding stages; model and protocol details are in \cref{app:full_method_details,app:full_experiment_protocols}.}
    \label{fig:pipeline}
\end{figure}

\subsection{Problem Definition}
\label{sec:problem}

We formalize bimanual dexterous grasp synthesis as a staged proposal-grounding-admission problem. For each side $l\in\{L,R\}$, let $\mathbf{q}_l=(\mathbf{T}_l,\boldsymbol{\theta}_l)$ denote wrist pose and hand joints, giving the bimanual configuration space $\qbi=\qsingle_L\times\qsingle_R$. The problem is searching for grasp poses that is both physically valid and close to the human interaction prior:

\begin{equation}
    \mathbf{q}^* \leftarrow \arg\max_{\mathbf{q} \in \qbi} \; S_\text{phys}(\mathbf{q}, \mathcal{M}) + S_\text{human}(\mathbf{q}),
    \label{eq:dual_obj}
\end{equation}

which is a 2-objective optimization, where $S_\text{phys}$ denotes physical grasp quality, related to well discussed metrics like collision avoidance and force-closure quality~\cite{li2003grasp,murray1994mathematical}, and $S_\text{human}$ denotes the human-likeness or proximity to human hand-object data such as GRAB~\cite{taheri2020grab}, implicitly related to affordance, grasp taxonomy and grasp direction.

\subsection{Human-to-Robot Optimization Framework}
\label{sec:retarget_opt}

\Cref{fig:pipeline} shows the full pipeline: diffusion samples object-conditioned human pre-grasps, geometric retargeting converts them into open XHand seeds, contact optimization grounds the grasp, and IK/rollout checks admit executable demonstrations.

Through the pipeline, we continuely improve the sum score defined in \cref{eq:dual_obj}. Digital human pre-grasps provide semantic search proposals, retargeted seeds match the robot morphology, grounded keyframes satisfy contact and collision constraints, and admitted trajectories must be reachable and dynamically executable. Given an object mesh $\mathcal{M}$, \method{} can be seem as 4-steps:

\begin{align}
    \mathbf{h}_0 &\sim p_\theta(\mathbf{h}\mid \mathcal{M}), \nonumber\\
    \mathbf{q}_{\mathrm{init}} &= R_\psi(\mathbf{h}_0,\mathcal{M}), \nonumber\\
    \mathbf{q}^{\star} &= \Pi_{\mathrm{phys}}(\mathbf{q}_{\mathrm{init}},\mathcal{M}), \nonumber\\
    \boldsymbol{\tau} &= \Pi_{\mathrm{exec}}(\mathbf{q}^{\star},\mathcal{M}).
    \label{eq:dual_obj}
\end{align}

\begin{figure}[H]
    \centering
    \includegraphics[width=\textwidth]{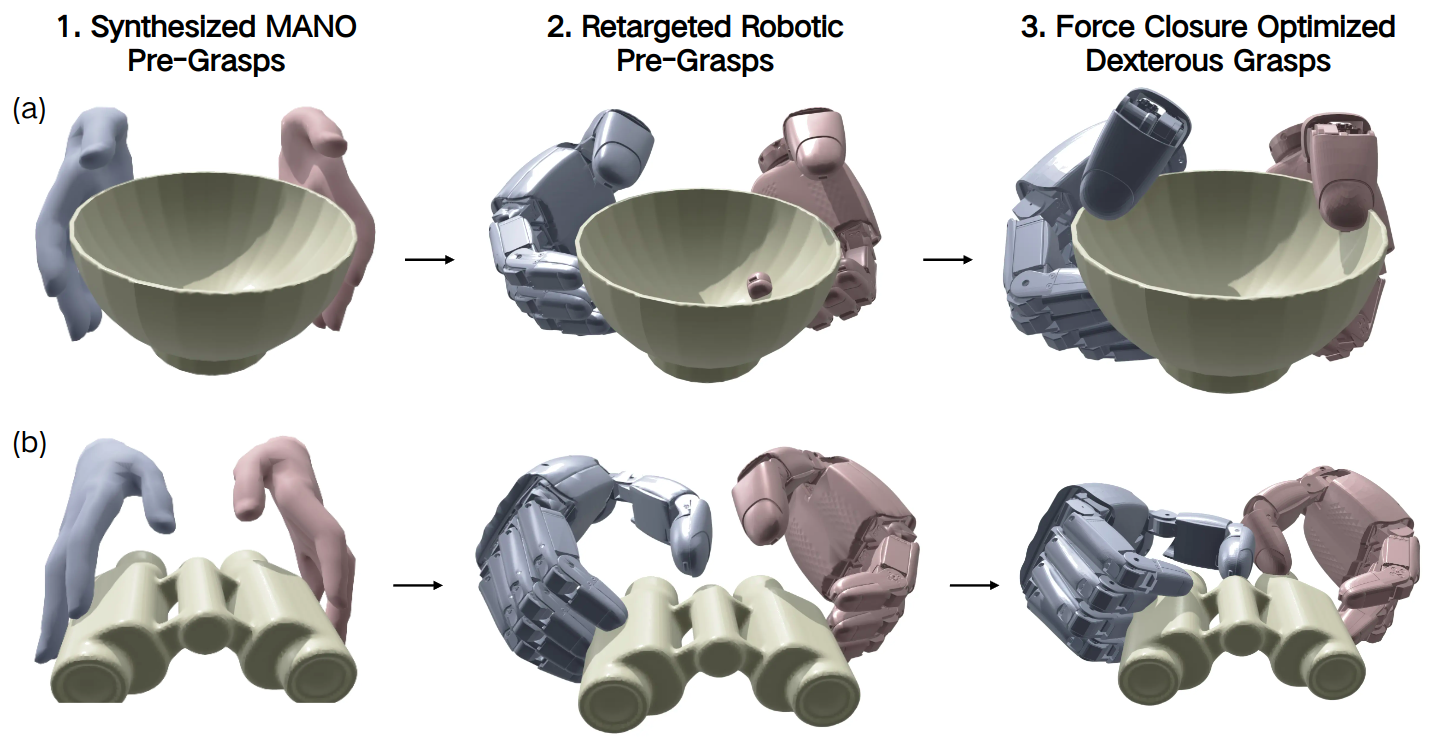}
    \caption{\textbf{Grasp generation stages.}
    This figure illustrates the generated digital human pre-grasp, robotic pre-grasp, and refined grasps. Left: digital human pre-grasps encode human-like approach direction and coarse finger coordination. Middle: geometric retargeting preserves the pre-grasp intents but can leave wrist offsets or object penetration. Right: force-closure optimization resolves contact on the robotic hand geometry while staying near the human-prior seed.}
    \label{fig:stage_refinement}
\end{figure}

Here $\mathbf{h}_0$ is a human hand pre-grasp proposal, $\mathbf{q}_{\mathrm{init}}$ is the retargeted robot seed, $\mathbf{q}^{\star}$ is the grounded keyframe, and $\boldsymbol{\tau}$ is the admitted trajectory. The maps $R_\psi$, $\Pi_{\mathrm{phys}}$, and $\Pi_{\mathrm{exec}}$ respectively perform retargeting, robot-native contact grounding, and arm-hand execution testing. We algorithm describe the algorithm in~\cref{app:problem_algorithm}. A trajectory is admitted only when
\begin{equation}
    A(\boldsymbol{\tau}) =
    A_{\mathrm{coll}}\land A_{\mathrm{FC}}\land A_{\mathrm{IK}}\land A_{\mathrm{lift}} .
    \label{eq:admission}
\end{equation}
The admission gate keeps the human prior out of the final validity test: contact, collision, IK, and rollout success are evaluated on the robot model~\cite{li2003grasp,murray1994mathematical}. Below \Cref{fig:stage_refinement} shows the intended progression: MANO proposes approach and coordination, retargeting transfers the basin, and force-closure refinement fixes the robot contacts.

\paragraph{Human Prior Synthesis.}
\label{sec:pregrasp}

SynManDex-Human is an object-conditioned diffusion model trained on hand-object resources such as GRAB and ContactPose~\cite{romero2017embodied,taheri2020grab,brahmbhatt2020contactpose}. For temporal sequences, we identify first contact by minimum hand-object distance and supervise the frame 0.2\,s earlier; static grasps are used as pose priors. This pre-contact target captures approach direction and coarse coordination without asking the robot to reproduce a human contact state. \Cref{app:diffusion_details} details the architecture.

\paragraph{Mapping to Robot Pre-Grasps.}
\label{sec:retargeting}

Following denoising, the model yields a digital human hand seed $\mathbf{X}^{H}$ with $H=21$ hand keypoints. Because human and robot hands differ in morphology and joint space, retargeting solves for an open robot pre-grasp rather than a closed contact pose.

Using geometric calibration principle~\cite{yin2025geort}, we train robotic hand maps $g_\psi=(g_{\psi_L}^{L},g_{\psi_R}^{R})$ from object-free bimanual calibration data and robot keypoint clouds generated by forward kinematics. To be more specific, we use hand tracking gloves (\cref{fig:retargeting_gloves}) to calibrate the human keypoint manifold. With robotic keypoints $\mathbf{x}^{R}_i(\boldsymbol{\theta})$, $g_\psi$ preserves local motion geometry, spatial coverage, pinch correspondence, and self-collision margins. To avoid undesirable finger-bending local minima, we relax the retargeted pose toward a taxonomy-compatible finger-open configuration before subsequent grasp optimization.

\begin{align}
    \mathcal{L}_{\text{GeoRT}}
    &= \lambda_{\text{dir}}\mathcal{L}_{\text{dir}}+\lambda_{\text{cov}}\mathcal{L}_{\text{cov}}+\lambda_{\text{flat}}\mathcal{L}_{\text{flat}}+\lambda_{\text{pinch}}\mathcal{L}_{\text{pinch}}+\lambda_{\text{self}}\mathcal{L}_{\text{self}},
    \label{eq:geort_train}\\
    (\boldsymbol{\theta}^0,\mathbf{T}^0)
    &= \arg\min_{\boldsymbol{\theta},\mathbf{T}}
    \sum_i\|\bar{\mathbf{x}}^{H}_i-\mathbf{T}\mathbf{x}^{R}_i(\boldsymbol{\theta})\|_2^2
    +\lambda_\psi\|\boldsymbol{\theta}-g_\psi(\mathbf{X}^{H})\|_2^2
    +\lambda_{\text{self}}\mathcal{L}_{\text{self}}(\boldsymbol{\theta}).
    \label{eq:retarget}
\end{align}

This yields the initial robot configuration for force-closure refinement.

\paragraph{Force-Closure Optimization.}
\label{sec:grasp_opt}

Initialized with $\mathbf{q}_{\text{init}}=(\mathbf{T}^0,\boldsymbol{\theta}^0)$, robot-native optimization reduces interpenetration, anchors local contacts, and improves force closure:

\begin{align}
    \mathbf{q}^*
    &= \arg\min_{\mathbf{q}}\; w_c C_{\text{coll}}(\mathbf{q})+w_f\mathcal{L}_{\text{FC}}(\mathbf{q})+w_r\|\mathbf{q}-\mathbf{q}_{\text{init}}\|^2,
    \label{eq:opt}\\
    Q_{\text{FC}}(\mathbf{q})
    &= \min_{\|\mathbf{w}\|_2=1}
    \max_{\mathbf{f}\in\mathcal{F}(\mathbf{q}),\,\|\mathbf{f}\|_1\le 1}
    \mathbf{w}^{\top}\mathbf{G}(\mathbf{q})\mathbf{f}.
    \label{eq:fc}
\end{align}

where $\mathcal{L}_{\text{FC}}=-Q_{\text{FC}}$, $\mathbf{G}$ maps contact forces to the object wrench, and the $\ell_1$ bound makes the wrench-margin test finite~\cite{li2003grasp,murray1994mathematical}. We use this discretized friction-cone metric as an admission score, not a physical guarantee; contact extraction and display scaling details are in \Cref{app:optimization_details}. Surviving candidates are checked in Isaac Sim~\cite{mittal2025isaac} and retained only if they support the object against gravity.

\paragraph{Physical Grounding.}
\label{sec:traj_policy}
\label{sec:geo_ik}

Grounded wrist-hand candidates still need arm reachability and collision-free execution. The admitted sample is a grasp-and-lift trajectory with approach, closure, squeeze, and lift phases. cuRobo~\cite{sundaralingam2023curobo} solves arm-hand IK and motion seeds, Isaac Sim checks isolated floating-hand stability, and SAPIEN~\cite{sapien3} rolls out the full UR5e+XHand trajectory following the UltraDexGrasp-style protocol~\cite{yang2025ultradexgrasp}. The IK solver optimizes:

\begin{align}
    (\mathbf{a}^{\star}_l,\boldsymbol{\eta}^{\star}_l)
    &= \arg\min_{\mathbf{a}_l,\boldsymbol{\eta}_l}
    \mathcal{L}_{SE(3)}+\lambda_{\text{kpt}}\mathcal{L}_{\text{kpt}}
    +\lambda_{\text{coll}}\mathcal{L}^{\text{IK}}_{\text{coll}}
    +\lambda_{\text{reg}}\mathcal{L}^{\text{IK}}_{\text{reg}},
    \label{eq:geort_ik}\\
    \mathcal{L}_{\text{kpt}}
    &= \sum_i \rho\!\left(\|\mathbf{T}^{A}_l(\mathbf{a}_l)\mathbf{x}^{R}_i(\boldsymbol{\theta}_l)-\mathbf{y}^{*}_{l,i}\|_2\right),
    \label{eq:ik_keypoint}\\
    y &= \mathbf{1}\!\left[\max_{t\ge t_{\text{lift}}}(z_t-z_0)>\tau_z\right].
    \label{eq:lift}
\end{align}

Only IK-valid trajectories that satisfy the vertical lift test ($y=1$) enter the imitation dataset. Extended tasks in \cref{sec:extended_evidence} reuse these keyframes as possession states.

\subsection{Dexterous Manipulation Data Pipeline}
\label{sec:data_utilization}

Admitted rollouts train a closed-loop point-cloud policy, and selected keyframes initialize extended object-centric manipulation.

\paragraph{Data.}
\label{sec:dataset}
For each admitted rollout, we store the object identifier, hand assignment, phase boundaries, optimized keyframe, validation metrics, and synchronized arm-hand states. These rollouts form the imitation dataset for a closed-loop, receding-horizon point-cloud policy operating on the union of scene geometry and rendered robot proprioceptive points:

\begin{align}
    \mathcal{D}
    &= \{(\mathbf{P}_t,\mathbf{a}_{t:t+H-1})\},\qquad
    \mathbf{P}_t=\mathbf{P}^{\mathrm{scene}}_t\cup\mathbf{P}^{\mathrm{robot}}_t,
    \label{eq:policy_dataset}
\\
    p_\varphi(\mathbf{a}_{t:t+H-1}\mid \mathbf{P}_t)
    &= \prod_{\tau=0}^{H-1}\operatorname{TN}(\mathbf{a}_{t+\tau};\boldsymbol{\mu}_{t+\tau},\boldsymbol{\sigma}_{t+\tau},\mathbf{a}_{\min},\mathbf{a}_{\max}),
    \label{eq:point_policy}\\
    \mathcal{L}_{\text{policy}}
    &= -\sum_{\tau=0}^{H-1}\log p_\varphi(\mathbf{a}_{t+\tau}\mid\mathbf{P}_t).
    \label{eq:policy_loss}
\end{align}

Each object receives the same 240 candidate budget before admission. The policy split, rollout horizon, and training hyperparameters are fixed before evaluation and detailed in \Cref{sec:appendix_policy}; held-out evaluation uses object instances not seen during policy training. Visual features extracted via PointNet++~\cite{qi2017pointnetplusplus} are combined with action queries to predict bounded 36-DoF joint-trajectory chunks, ordered deterministically across the bimanual system (left-arm, left-hand, right-arm, right-hand). Rather than executing open-loop trajectories, the policy replans at each short horizon based on the most recent point-cloud observation.

\paragraph{Manipulation.}
\label{sec:manip}

For extended object-centric manipulation, a task specification defines object-relative motion, hand roles, and release conditions. We reuse the IK, collision, force-closure, and possession checks to admit transitions for in-grasp reconfiguration, bimanual handover, and pick-and-place.

\subsection{VLM Agent for Task Specification}
\label{sec:vlm_agent_method}

The final stage uses the admitted grasp keyframe as an interface between robot-grounded contact and open-ended task specification. Given a validated keyframe $k=(\mathcal{M},T_o^0,q_L^\star,q_R^\star,\mathcal{C}_L,\mathcal{C}_R,m)$, where $T_o^0$ is the object pose, $q_l^\star$ are the optimized hand states, $\mathcal{C}_l$ are contact regions and normals, and $m$ records the admission metrics, we query a VLM agent with multi-view renders, object metadata, contact annotations, hand-role candidates, and a library of admissible task primitives to prevent from out-of-distributions. The agent outputs a task proposal
\[
    u = (g, r_L, r_R, \Xi_o, \rho_{\mathrm{rel}}, \rho_{\mathrm{succ}}),
\]
where $g$ is the functional goal, $r_L,r_R\in\{\text{active},\text{stabilizing},\text{receiving},\text{released}\}$ are hand roles, $\Xi_o=\{T_o^t\}_{t=1}^{T}$ is an object-relative waypoint sequence, and $\rho_{\mathrm{rel}}$ and $\rho_{\mathrm{succ}}$ define release and terminal-possession predicates. The VLM agent is separate from the VLM-H evaluator in \cref{sec:metric_protocol}: VLM-H scores human-likeness, whereas the VLM agent proposes task specifications that are later checked by the executor. It does not directly command joint torques or replace physical validation. Instead, it proposes semantically meaningful object motions, such as tilting a teapot around its spout axis, aligning a camera toward a target view, or releasing selected flute fingers while preserving support.

For each hand that maintains possession, we convert the proposed object motion into wrist targets by preserving the keyframe contact transform,
\[
    T_{w,l}^t = T_o^t (T_o^0)^{-1} T_{w,l}^0,
\]
with optional finger-joint deltas only when the agent selects a release or regrasp primitive. These targets are passed to the same validations used for admitted trajectories in \cref{sec:data_utilization}. Similar to \cref{eq:admission}, the VLM proposal is retained only when
\[
    A_{\mathrm{agent}}(\tau)=A_{\mathrm{coll}}\wedge A_{\mathrm{IK}}\wedge A_{\mathrm{poss}}\wedge A_{\mathrm{task}},
\]
where $A_{\mathrm{poss}}$ checks stable possession by at least one non-released hand and $A_{\mathrm{task}}$ checks the terminal predicate. Failed proposals are returned with failure type labels such as wrist unreachable, approach collision, release slip, or terminal overshoot. \Cref{app:vlm_agent} gives the context packet, prompt, output schema, and retry protocol.

\section{Experiments}
\label{sec:experiments}

The experiments follow four questions. RQ1 evaluates the effect of human pre-grasp priors on grasp quality and human-likeness. RQ2 measures executable admission under robot-native force-closure, inverse kinematics, and rollout filters against taxonomy, optimization, and retargeting baselines. RQ3 tests whether admitted trajectories improve closed-loop point-cloud policy learning. RQ4 studies downstream utility through selected manipulation transitions, hardware transfer, a cross-embodiment seed diagnostic, and agent-generated task proposals. \Cref{tab:rq_map} maps each question to its evidence.

\begin{table}[H]
\centering
\scriptsize
\setlength{\tabcolsep}{3pt}
\renewcommand{\arraystretch}{1.08}
\caption{\textbf{Research questions and evidence.} Each RQ lists the figures and tables used as primary evidence. Appendix entries provide protocols or extended diagnostics rather than additional main claims.}
\label{tab:rq_map}
\begin{tabular*}{\textwidth}{@{\extracolsep{\fill}}p{0.055\textwidth}p{0.24\textwidth}p{0.22\textwidth}p{0.26\textwidth}p{0.13\textwidth}@{}}
\toprule
RQ & Tested mechanism & Primary metrics & Main evidence & Appendix details \\
\midrule
RQ1 & Human pre-grasp prior & G1, Pen., Contact, FC, H-score, PCD & \Cref{tab:grasp_quality,fig:stage_refinement,fig:fc_refinement,fig:diversity} & \Cref{fig:retargeting_gloves,tab:diversity_human,sec:metric_protocol} \\
RQ2 & Robot-native grounding and matched baselines & FC, IK/lift, task match, benchmark success & \Cref{tab:admission,tab:baseline_scope,tab:embodiment_grasp_success,tab:grasp_benchmark,fig:bimanual_consistency,fig:bottle_direction,fig:flute_modes} & \Cref{sec:main_experiment_protocols} \\
RQ3 & Admitted trajectory data for policy & rollout success, action L2 & \Cref{tab:policy_ablation,fig:admitted_demonstrations} & \Cref{tab:policy_protocol,tab:geort_ik_ablation} \\
RQ4 & Downstream manipulation, hardware, and diagnostics & transition success, hardware success, cross-embodiment seed gains, executor admission & \Cref{tab:ingrasp,tab:handover_pick_place,tab:hardware,tab:shadow_seed_diagnostic,fig:hardware_benchmark,fig:shadow_seed_diagnostic,sec:vlm_agent_exp} & \Cref{sec:main_experiment_protocols,app:shadow_cross_embodiment,app:vlm_agent} \\
\bottomrule
\end{tabular*}
\end{table}

\subsection{Setup}
\label{sec:exp_setup}

We instantiate this evaluation across a progressive embodiment stack: MANO space~\cite{romero2017embodied} $\rightarrow$ floating-base XHand $\rightarrow$ fully actuated 36-DOF bimanual UR5e-XHand system ($\qbi$). Each optimization query uses 240 parallel cuRobo seeds~\cite{sundaralingam2023curobo}, followed by dynamic floating-hand stability checks in Isaac Sim. Unless a table states otherwise, success requires the same admission chain: feasible contact, low penetration, positive force closure, kinematic validation, and a 10\,cm lift or task-specific terminal-possession check. This convention makes the reported numbers stricter than static visual plausibility.
We report five distinct success notions. \emph{FC success} is the percentage of optimized grasp candidates with positive discretized force-closure margin under the contact model. \emph{IK/lift admission} is the percentage of candidates that pass arm-hand IK and the 10\,cm dynamic lift rollout. \emph{Policy success} is closed-loop rollout success on held-out simulated objects. \emph{Transition success} applies to in-grasp, handover, and pick-and-place protocols and requires terminal possession. \emph{Hardware success} is the real-robot three-object tabletop success rate. Unless otherwise specified, the headline 86.4\% grasp-quality number refers to FC success on the 312-object manifest, while 65.8\% is the lift-admitted trajectory rate. The headline 4.67/5 human-likeness score is the combined blinded audit score defined in \Cref{sec:metric_protocol}.

\begin{table}[H]
\centering
\scriptsize
\setlength{\tabcolsep}{3pt}
\renewcommand{\arraystretch}{1.08}
\caption{\textbf{Admission funnel for the data.} Rates are reported under the fixed 240-candidate-per-object budget. Generator-level rates use the optimized candidate denominator for the corresponding stage; lift admission is reported separately from policy and hardware success.}
\label{tab:admission}
\begin{tabular*}{\textwidth}{@{\extracolsep{\fill}}p{0.23\textwidth}p{0.22\textwidth}p{0.18\textwidth}p{0.25\textwidth}@{}}
\toprule
Stage & Validity criterion & Pass signal & Main rejection mode \\
\midrule
MANO pre-grasp proposals & Object-conditioned semantic basin & 240 seeds / object, 312 objects & none at proposal stage \\
Retargeted robot seeds & Morphology-consistent open-hand seed & self-collision filter & gross self-collision / wrist offset \\
Grounded keyframes & Penetration, contact, bounded FC score & 86.4\% FC among optimized XHand candidates & low FC / excessive penetration \\
IK-valid trajectories & Arm-hand reachability and collision-aware path & 82.3\% IK valid among grounded candidates & wrist unreachable / approach collision \\
Lift-admitted demonstrations & 10\,cm dynamic lift in rollout & 65.8\% lift-admitted among grounded candidates & slip / object loss \\
\bottomrule
\end{tabular*}
\end{table}

\begin{figure}[H]
    \centering
    \includegraphics[width=\textwidth]{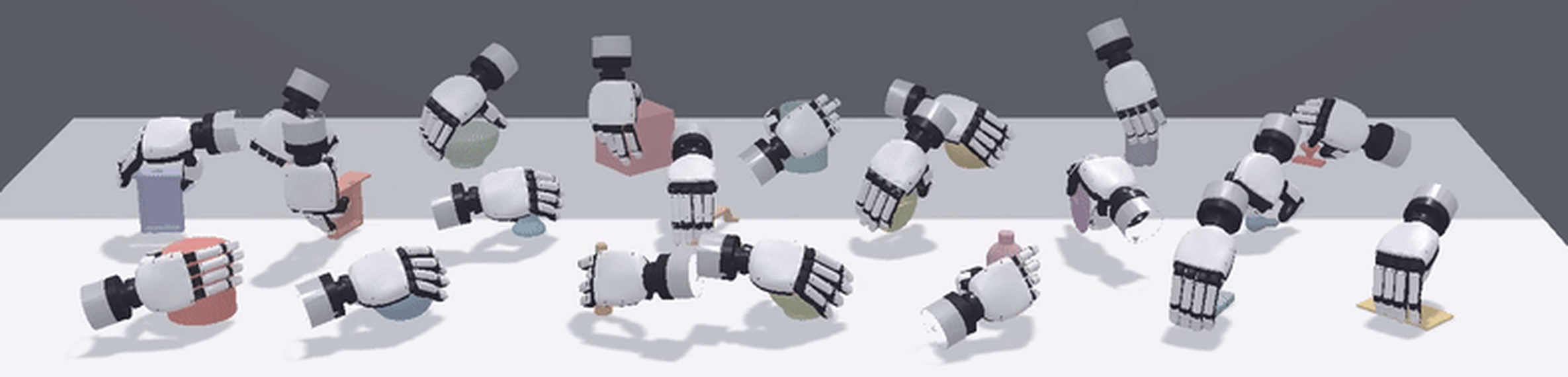}
    \caption{\textbf{Qualitative coverage of generated bimanual candidates.}
    The gallery shows representative object-conditioned grasp configurations before policy training. Candidate admission is determined by the FC, penetration, IK, and lift gates summarized in \cref{tab:admission} and detailed in \cref{sec:main_experiment_protocols}; this figure illustrates coverage rather than reporting success rates.}
    \label{fig:dataset_gallery}
\end{figure}

\paragraph{Dataset Illustrations.}
\Cref{fig:dataset_gallery} shows the spatial and semantic coverage of generated bimanual candidates. Quantitative claims use the admission criteria in \Cref{sec:data_utilization,sec:main_experiment_protocols}.

\subsection{Human Priors and Optimizations}
\label{sec:qualitative}

\begin{figure}[t]
    \centering
    \includegraphics[width=\textwidth]{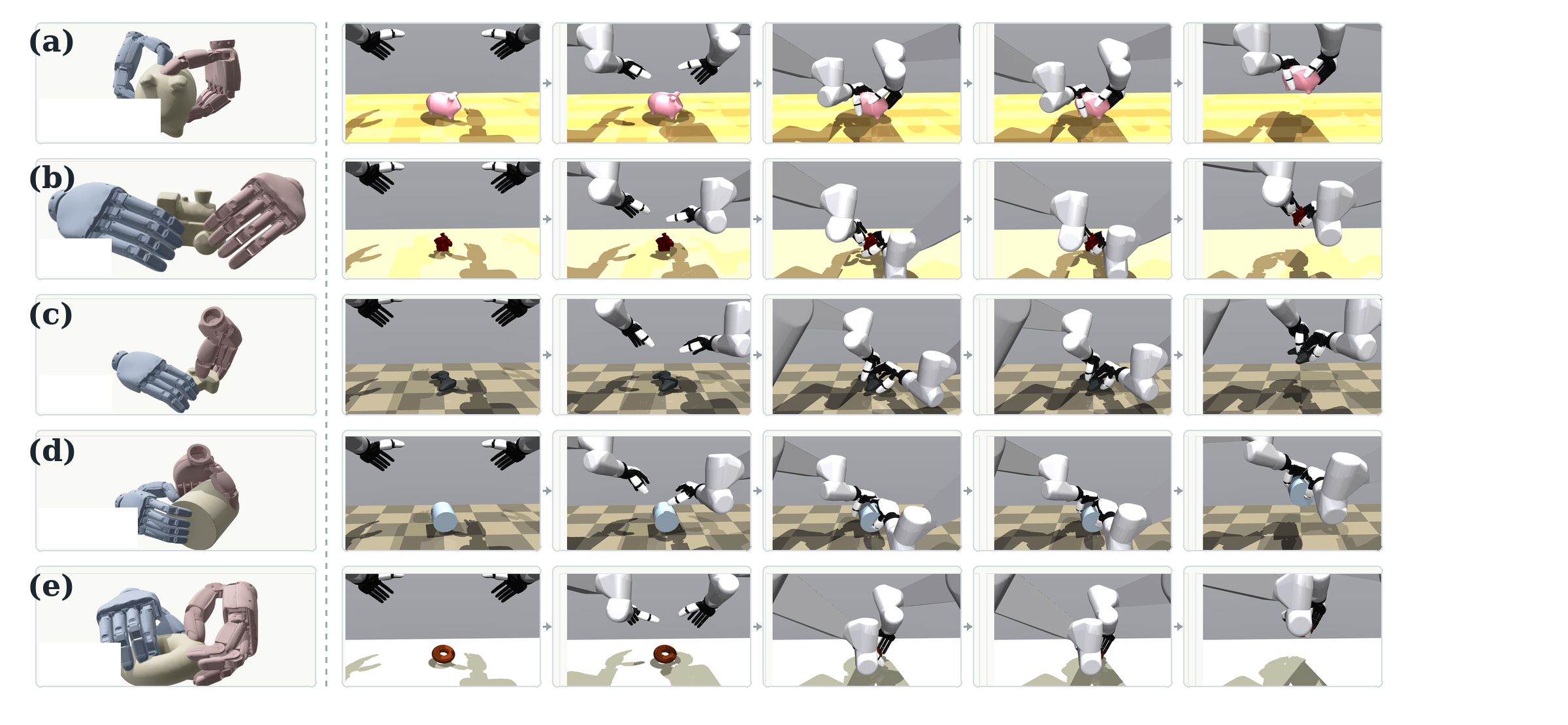}
    \caption{\textbf{Generating trajectories.}
    Rows show piggy-bank, rose, duck, cylinder, and donut examples. In each row, the left panel is the optimized goal keyframe, followed by the executed rollout sequence: start, approach, pre-grasp, grasp, and lift. A rollout enters the imitation dataset only if it passes physical checks and the 10\,cm lift test in \cref{eq:lift}; aggregate admission rates are summarized in \cref{tab:admission}.}
    \label{fig:admitted_demonstrations}
\end{figure}

\begin{figure}[t]
    \centering
    \includegraphics[width=\textwidth]{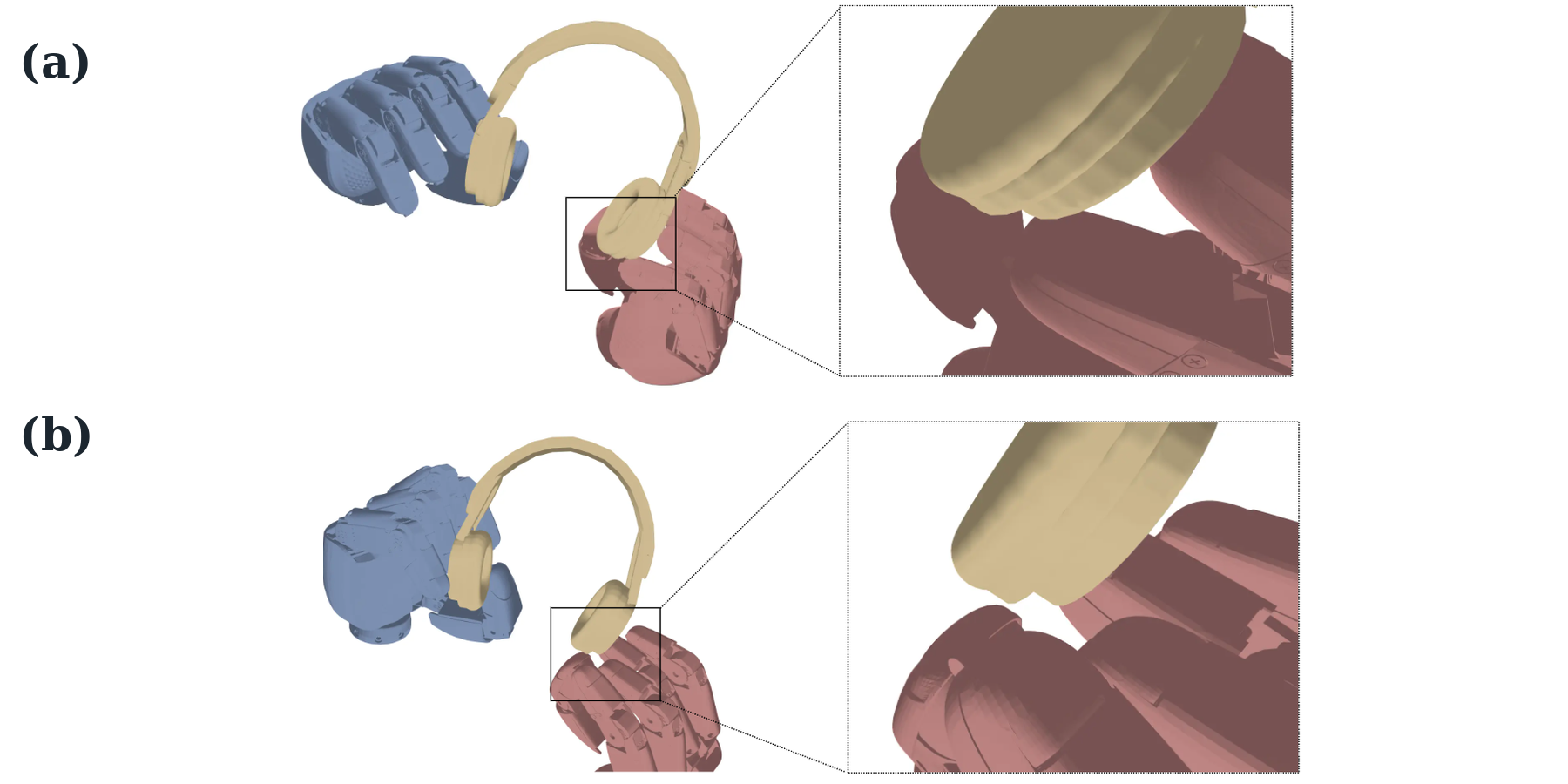}
    \caption{\textbf{Force-closure refinement corrects retargeted contact failures.}
    (a) Full \method{} refinement grounds local fingertip-object contacts on the XHand geometry. (b) Retarget-only output remains visually plausible but misses the load-bearing contact patch. This diagnostic explains the gap between retarget-only and full-pipeline FC rates in \cref{tab:grasp_quality}.}
    \label{fig:fc_refinement}
\end{figure}

For RQ1, we compare human-prior seeding, retarget-only transfer, and full robot-native refinement. \Cref{fig:stage_refinement,fig:fc_refinement,fig:admitted_demonstrations} separate proposal, contact grounding, and executable admission. Retarget-only preserves a plausible hand shape but leaves contact unresolved; the full pipeline keeps the intended grasp role while reducing penetration and raising FC success in \Cref{tab:grasp_quality}.

\Cref{fig:admitted_demonstrations} evaluates accepted keyframes as trajectories: after contact refinement, each sample must still survive approach, closure, and lift on the policy embodiment. \Cref{fig:fc_refinement} shows the local contact correction imposed by \cref{eq:opt}.

\begin{figure}[t]
    \centering
    \includegraphics[width=\textwidth, trim={0 0 0 100}, clip]{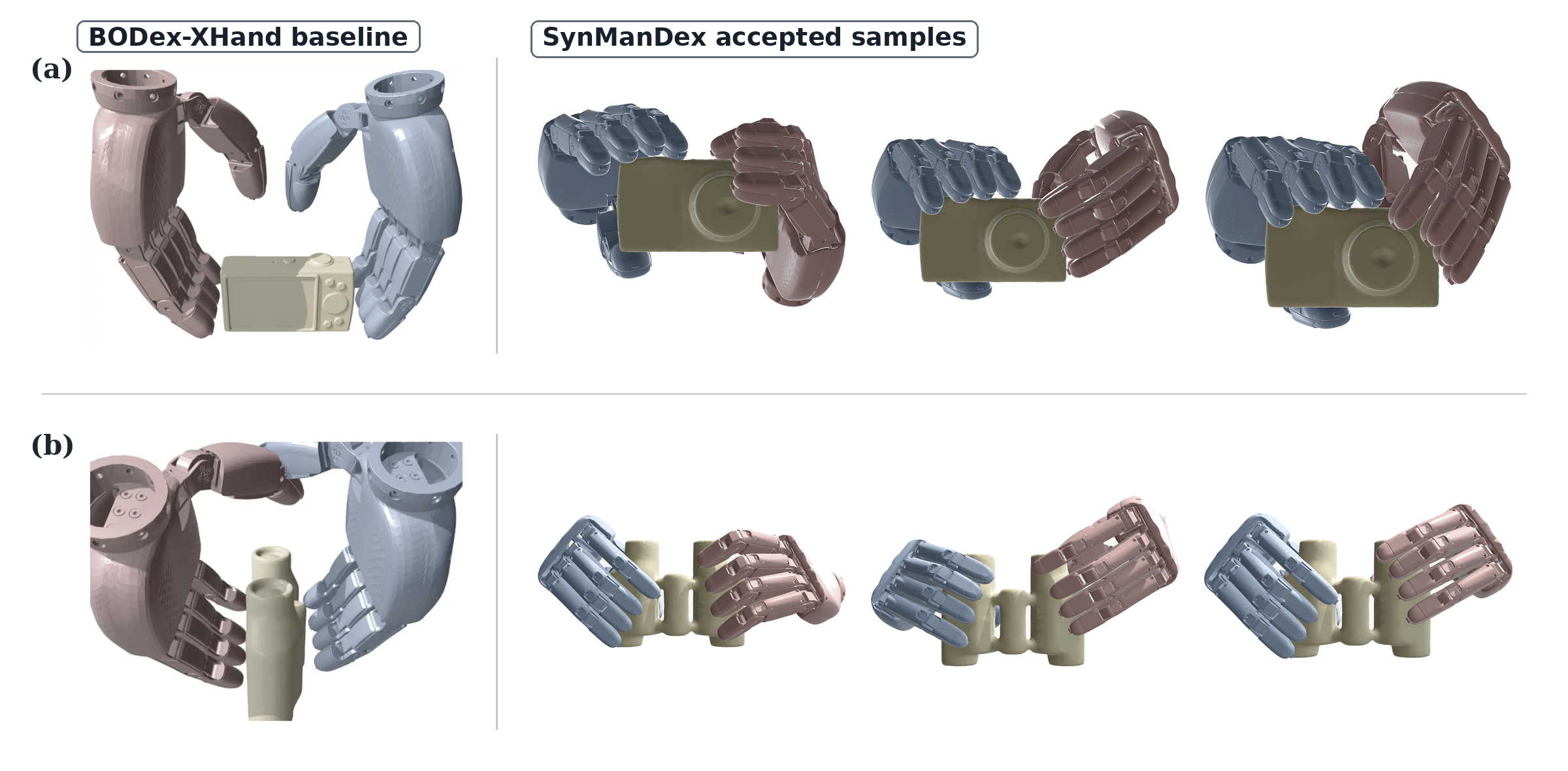}
    \caption{\textbf{Grasping camera and binoculars.}
    Rows (a) and (b) show camera and binoculars. In each row, the image left of the divider is the bimanual version of BODEX optimization baseline from UltraDexGrasp, while the three images to the right are accepted \method{} grasps generated from human-prior seeds. The figure illustrates how human-prior basins preserve object-specific bimanual roles; aggregate matched-baseline results are reported in \cref{tab:embodiment_grasp_success,tab:grasp_benchmark}.}
    \label{fig:bimanual_consistency}
\end{figure}

\begin{figure}[t]
    \centering
    \includegraphics[width=\textwidth]{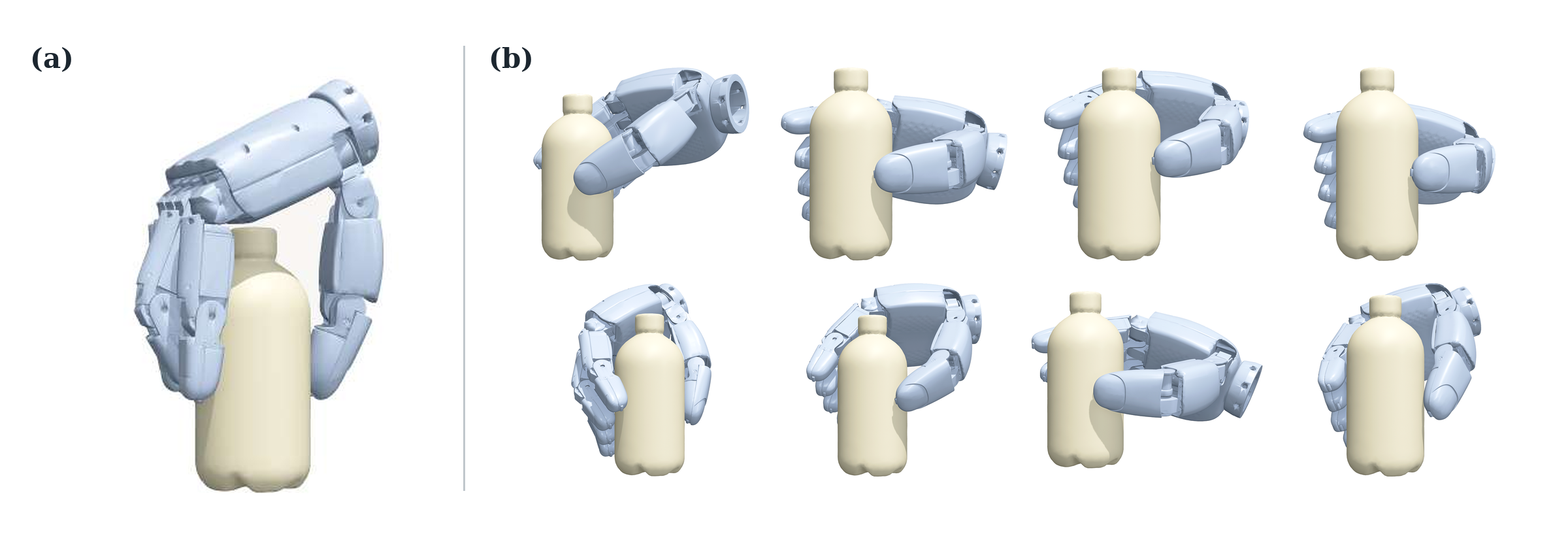}
    \caption{\textbf{Bottle grasping.}
    (a) The baseline converges to a stable wrap that does not preserve the intended side-oriented grasp prior. (b) \method{} samples retain side-approach directions while satisfying robot-native contact constraints. This qualitative stress case complements the task-match comparison in \cref{tab:embodiment_grasp_success}.}
    \label{fig:bottle_direction}
\end{figure}

\begin{figure}[t]
    \centering
    \includegraphics[width=\textwidth]{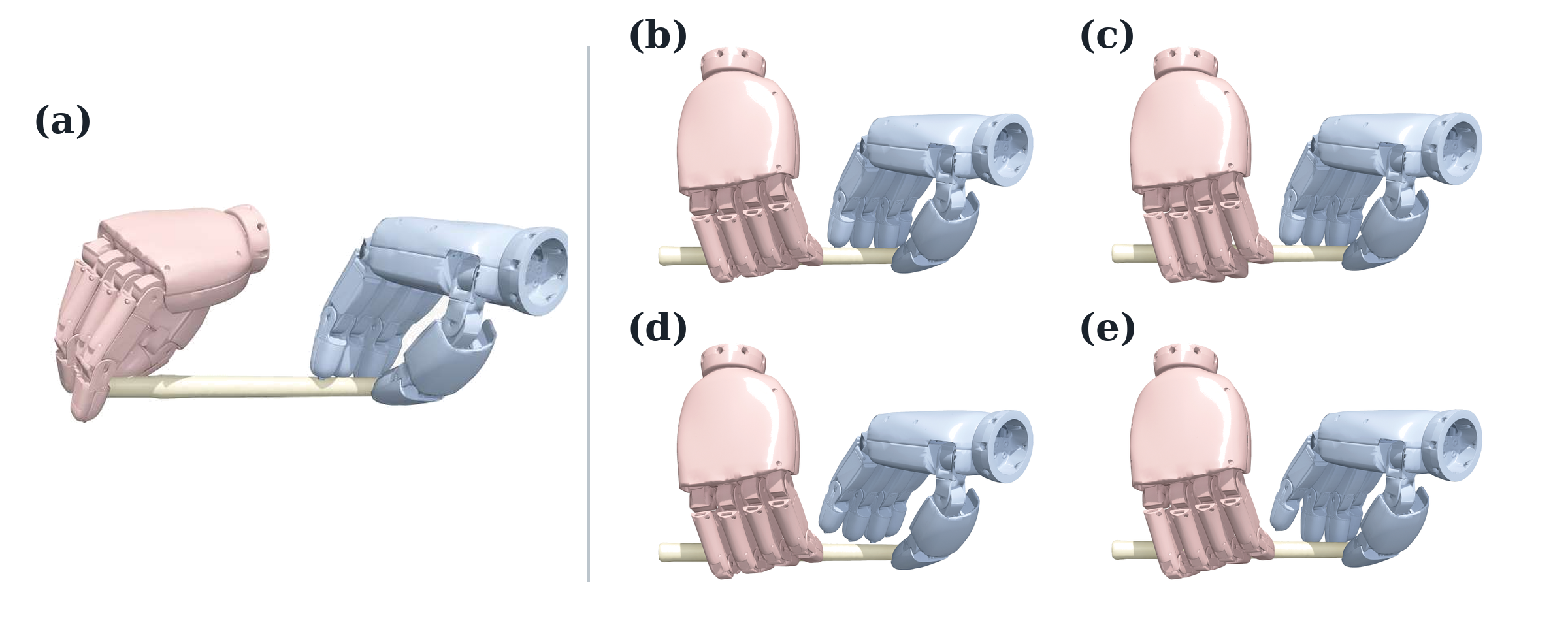}
    \caption{\textbf{Fine-grained flute-holding contact modes.}
    (a) A generic bimanual grasp can support the flute but does not preserve the task-motivated finger-contact configuration. (b)--(e) \method{} grounds variants from the same pose family, covering all-finger support, left-hand release, right-hand release, and cross-hand release. \Cref{sec:supp_flute_taxonomy} expands this example into the full finger-release taxonomy.}
    \label{fig:flute_modes}
\end{figure}

\paragraph{Functional grasp experiment.}
Aggregate metrics can hide task-intent errors, so \Cref{fig:bimanual_consistency,fig:bottle_direction,fig:flute_modes} pair representative baseline failures with \method{} recoveries. The cases test bimanual role assignment on cameras and binoculars, side-grasp direction on bottles, and finger-contact modes for flute holding. The appendix expands this case into a flute-playing taxonomy (\cref{sec:supp_flute_taxonomy}).

\begin{table}[H]
\centering
\scriptsize
\setlength{\tabcolsep}{2.4pt}
\renewcommand{\arraystretch}{1.08}
\caption{\textbf{Grasp quality and human-likeness on the 312-object, 25-class grasp-quality manifest.} All methods use the same candidate budget and XHand evaluator. G1 is the scaled Ferrari--Canny-style wrench margin under the discretized contact model; Pen. is hand-object penetration in mm; Contact and FC are percentages of optimized candidates satisfying the corresponding contact-region and positive-margin gates. Human-likeness reports method-blinded VLM and human audits on a 1--5 scale; Comb.-H is the headline 4.67/5 score, and PCD is defined in \cref{sec:metric_protocol}.}
\label{tab:grasp_quality}
\begin{tabular*}{\textwidth}{@{\extracolsep{\fill}}lcccccccc@{}}
\toprule
& \multicolumn{4}{c}{Physical quality} & \multicolumn{4}{c}{Human-likeness / diversity} \\
\cmidrule(lr){2-5}\cmidrule(lr){6-9}
Method & G1 $\uparrow$ & Pen. (mm) $\downarrow$ & Contact (\%) $\uparrow$ & FC (\%) $\uparrow$ & VLM-H (1--5) $\uparrow$ & Human-H (1--5) $\uparrow$ & Comb.-H (1--5) $\uparrow$ & PCD $\uparrow$ \\
\midrule
\rowcolor{dexfill} \textbf{\method{} (full)} & \textbf{7.2} & 0.6 & \textbf{89.2} & \textbf{86.4} & \textbf{4.72} & \textbf{4.59} & \textbf{4.67} & \textbf{0.41} \\
Optimization-only & 4.6 & \textbf{0.6} & 71.6 & 79.1 & 2.86 & 2.74 & 2.81 & 0.11 \\
Retarget-only & 0.4 & 8.3 & 34.7 & 12.3 & 4.28 & 4.03 & 4.18 & 0.09 \\
\bottomrule
\end{tabular*}
\end{table}

Compared with \emph{Optimization-only}, \method{} improves G1 stability by 56.5\%, decreases penetration by 45.5\%, and raises the combined human-likeness score to 4.67/5.0. The ablations expose two failure modes: \emph{Retarget-only} preserves visual intent but lacks load-bearing contacts, while \emph{Optimization-only} finds stable contacts that are often unnatural or task-inappropriate. Among full-pipeline candidates, 82.3\% satisfy IK reachability and 65.8\% execute the dynamic lift. Entropy, VLM, human-audit, and PCD protocols are detailed in \Cref{sec:metric_protocol,tab:diversity_human}.

\begin{figure}[H]
    \centering
    \setlength{\tabcolsep}{1.5pt}
    \renewcommand{\arraystretch}{0.8}
    \begin{tabular}{ccc}
        \includegraphics[width=0.315\linewidth]{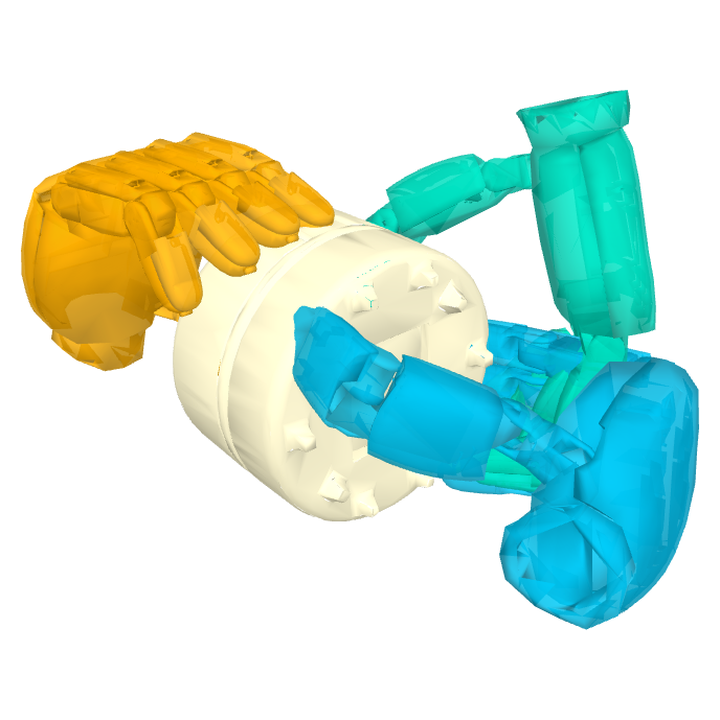} &
        \includegraphics[width=0.315\linewidth]{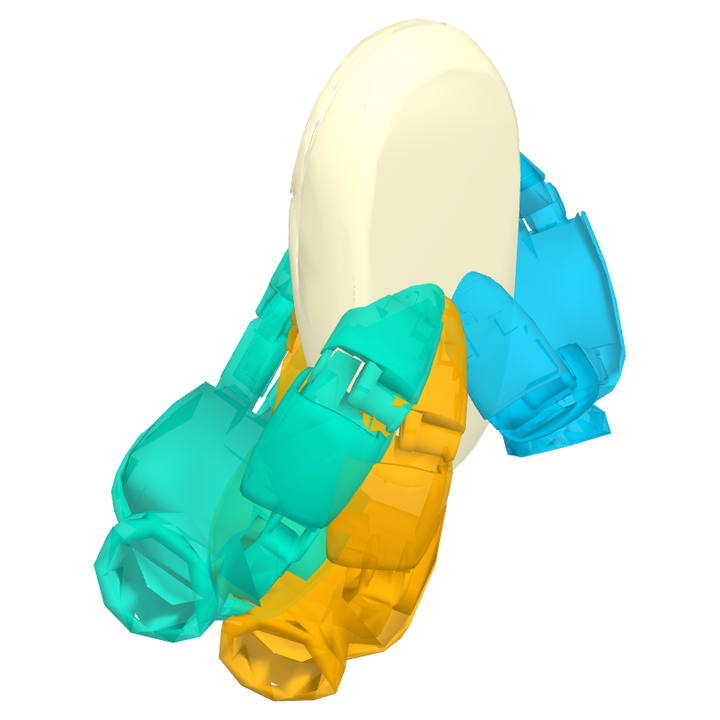} &
        \includegraphics[width=0.315\linewidth]{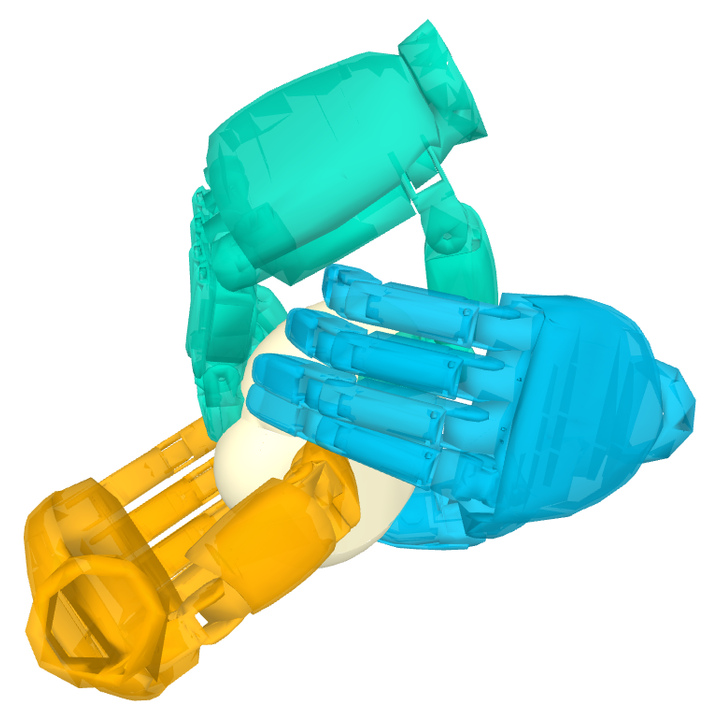} \\
        \includegraphics[width=0.315\linewidth]{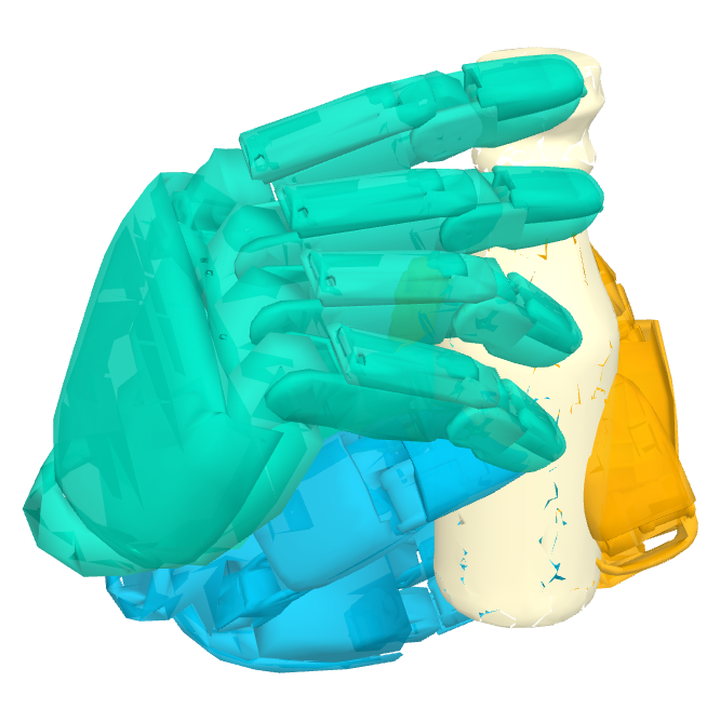} &
        \includegraphics[width=0.315\linewidth]{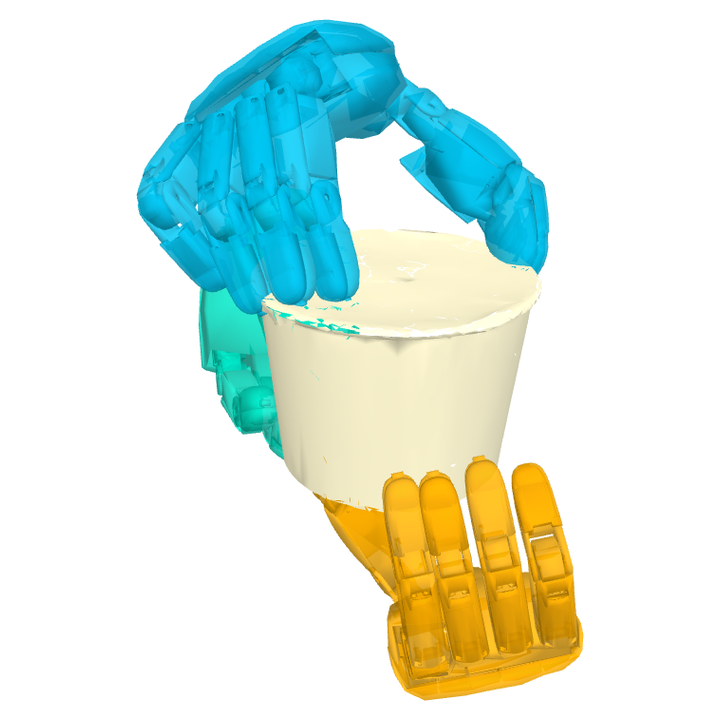} &
        \includegraphics[width=0.315\linewidth]{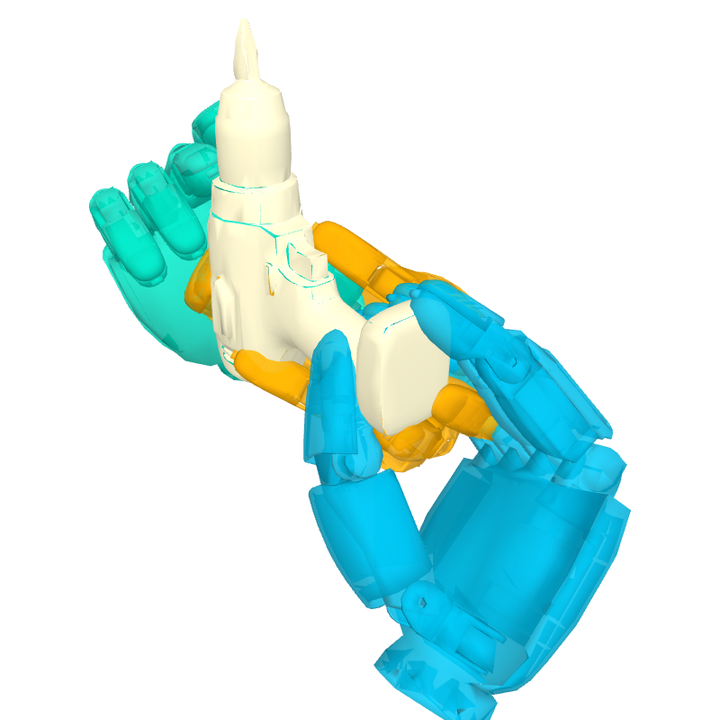}
    \end{tabular}
    \caption{\textbf{Diverse dexterous grasps given diverse human priors.}
    Six accepted \method{} grasps show different hand-role assignments, approach directions, and object-relative support patterns after passing the admission chain. The examples visualize the plausibility-constrained diversity score reported in \cref{tab:grasp_quality,tab:diversity_human}.}
    \label{fig:diversity}
\end{figure}

\paragraph{Grasp diversity evaluations.}
High joint entropy alone can reward implausible or non-contact postures, so we audit diversity after the same physical filters have already been applied. \Cref{fig:diversity} shows that accepted \method{} grasps remain diverse in hand-role assignment, approach direction, and object-relative support, complementing the PCD score in \Cref{tab:grasp_quality,tab:diversity_human}.

\subsection{Comparison Studies}
\label{sec:benchmark_taxonomy}

RQ2 compares taxonomy, optimization, and retargeting baselines under matched XHand contact, FC, IK, and lift filters. Dexonomy is adapted to XHand~\cite{chen2025dexonomy}. Unimanual rows evaluate single functional grasps; bimanual rows add dual-hand coordination, collision-free reachability, and paired lift execution.

Several baselines were originally designed for different hands, static grasp generation, or human-video imitation rather than the exact UR5e+XHand setting. We therefore separate official method scope from our adaptation. All adapted baselines use the same object poses, candidate budget, XHand collision model, force-closure threshold, IK solver, and lift evaluator as \method{}; only the proposal or seed source changes.

\begin{table}[H]
\centering
\scriptsize
\setlength{\tabcolsep}{3pt}
\renewcommand{\arraystretch}{1.08}
\caption{\textbf{Baseline scope and adaptation.} Matched rows share the same XHand evaluator; contextual rows retain their closest available published setup.}
\label{tab:baseline_scope}
\begin{tabular*}{\textwidth}{@{\extracolsep{\fill}}p{0.19\textwidth}p{0.28\textwidth}p{0.31\textwidth}p{0.12\textwidth}@{}}
\toprule
Row & Original method scope & Our adaptation & Type \\
\midrule
Dexonomy-XHand & taxonomy-conditioned grasp templates & map selected templates to XHand, then apply the same FC/IK/lift filters & adapted \\
BODex-XHand & optimization-based dexterous grasp synthesis & replace hand model with XHand and use the same object poses and filters & adapted \\
DexMV-style retargeting & human-video imitation / retargeting pipeline & use MANO or bimanual MANO pose transfer as a seed-only baseline & adapted \\
GeoRT-XHand & geometric neural hand retargeting & train XHand-specific maps; no FC refinement unless stated & adapted \\
UltraDexGrasp & bimanual synthetic trajectory data & closest trajectory-level reference under matched evaluator where possible & contextual \\
\bottomrule
\end{tabular*}
\end{table}

\begin{table}[H]
\centering
\small
\setlength{\tabcolsep}{3pt}
\renewcommand{\arraystretch}{1.12}
\caption{\textbf{Embodiment-aware grasp success.} All methods use identical held-out object-task configurations; FC and IK/lift are generator-level admission metrics and are distinct from policy success in \cref{sec:quantitative}.}
\label{tab:embodiment_grasp_success}
\begin{tabular*}{\textwidth}{@{\extracolsep{\fill}}llccc@{}}
\toprule
Method & Prior / seed & FC (\%) $\uparrow$ & IK/lift (\%) $\uparrow$ & Task match (\%) $\uparrow$ \\
\midrule
\rowcolor{synfill}\multicolumn{5}{@{}l}{\textit{Unimanual grasping}} \\
Dexonomy-XHand~\cite{chen2025dexonomy} & taxonomy & 42.5 & 28.3 & 76.8 \\
BODex-XHand~\cite{chen2025bodex} & optimization & 71.4 & 39.6 & 45.0 \\
DexMV-style retargeting~\cite{qin2022dexmv} & MANO pose & 12.3 & 8.1 & 72.5 \\
GeoRT-XHand~\cite{yin2025geort} & geometric kpts. & 38.9 & 24.7 & 68.2 \\
\rowcolor{dexfill} \textbf{\method{}-Uni} & \textbf{human prior + FC} & \textbf{86.4} & \textbf{72.9} & \textbf{81.7} \\
\midrule
\rowcolor{manfill}\multicolumn{5}{@{}l}{\textit{Bimanual grasping}} \\
BODex-XHand-bimanual~\cite{chen2025bodex} & paired optimization & 58.6 & 35.4 & 41.2 \\
BODex-XHand-uni2bim~\cite{chen2025bodex} & uni $\rightarrow$ bim & 49.7 & 26.8 & 43.9 \\
DexMV-style retargeting-bim~\cite{qin2022dexmv} & bimanual MANO & 10.8 & 6.5 & 74.1 \\
GeoRT-XHand-bim~\cite{yin2025geort} & bimanual kpts. & 36.2 & 21.9 & 69.3 \\
\rowcolor{dexfill} \textbf{\method{}} & \textbf{human prior + FC} & \textbf{84.8} & \textbf{65.8} & \textbf{82.6} \\
\bottomrule
\end{tabular*}
\end{table}

\begin{table}[H]
\centering
\scriptsize
\setlength{\tabcolsep}{2.5pt}
\renewcommand{\arraystretch}{1.08}
\caption{\textbf{Comparing grasping frameworks.} Static pose-only baselines are passed through the same \method{} IK/rollout stack. Bench success requires the standardized static grasp checks under the matched contact model; IK/lift success is a generator-level admission metric and is distinct from the policy success reported in \cref{sec:quantitative}.}
\label{tab:grasp_benchmark}
\begin{tabular*}{\textwidth}{@{\extracolsep{\fill}}lcccccc@{}}
\toprule
& \multicolumn{2}{c}{Setup} & \multicolumn{4}{c}{Evaluation} \\
\cmidrule(lr){2-3}\cmidrule(l){4-7}
Method & Artifact & Bimanual & Pen.\ (mm) $\downarrow$ & FC (\%) $\uparrow$ & Bench success (\%) $\uparrow$ & IK/lift (\%) $\uparrow$ \\
\midrule
Dexonomy-XHand~\cite{chen2025dexonomy} & pose & no & 4.7 & 42.5 & 36.8 & 28.3 \\
DexGraspNet~\cite{wang2022dexgraspnet} & pose & no & 3.4 & 54.8 & 46.2 & 33.9 \\
BODex~\cite{chen2025bodex} & pose & no & 1.4 & 74.6 & 63.5 & 45.7 \\
UltraDexGrasp~\cite{yang2025ultradexgrasp} & trajectory & yes & 1.9 & 70.8 & 62.1 & 58.6 \\
\rowcolor{dexfill} \textbf{\method{}} & trajectory & yes & \textbf{0.6} & \textbf{86.4} & \textbf{78.9} & \textbf{65.8} \\
\bottomrule
\end{tabular*}
\end{table}

\Cref{tab:embodiment_grasp_success,tab:grasp_benchmark} support the mechanism-level diagnosis from \Cref{sec:qualitative}. Taxonomy and direct MANO retargeting preserve task semantics but lose samples at execution filters. Optimization-based baselines improve contacts but lag in task match and IK/lift admission without a human-functional basin. Policy success is evaluated separately in \cref{sec:quantitative}.

\subsection{Grasp Policy Learning}
\label{sec:quantitative}

RQ3 evaluates whether admitted trajectories improve closed-loop policy learning. \Cref{tab:policy_ablation} ablates the data source and point-cloud policy interface while keeping the held-out object split, training budget, and terminal lift criterion fixed.

\begin{table}[H]
\centering
\small
\caption{\textbf{Policy ablation study on the held-out simulated object split.} Each row trains the same policy architecture while changing either the data source or the policy interface. Avg. L2 is the mean joint-space action error over executed chunks; $\Delta$ is the percentage-point difference from the full \method{} policy.}
\label{tab:policy_ablation}
\begin{tabular*}{\textwidth}{@{\extracolsep{\fill}}lccc@{}}
\toprule
Configuration & Success (\%) $\uparrow$ & $\Delta$ & Avg. L2 (rad) $\downarrow$ \\
\midrule
\rowcolor{dexfill} \textbf{Full method (\method{} policy)} & \textbf{80.7} & --- & \textbf{0.474} \\
\midrule
\multicolumn{4}{l}{\emph{Data source ablations}} \\
\quad No human prior (random-init optim.) & 37.1 & $-$43.6 & 0.622 \\
\quad No force closure (retarget-only) & 22.9 & $-$57.8 & 0.893 \\
\quad No pre-validation & 42.9 & $-$37.8 & 0.561 \\
\midrule
\multicolumn{4}{l}{\emph{Policy-interface ablations}} \\
\quad Scene-only point cloud & 45.7 & $-$35.0 & 0.539 \\
\quad MLP pooling without action queries & 40.0 & $-$40.7 & 0.601 \\
\bottomrule
\end{tabular*}
\end{table}

The complete \method{} policy reaches 80.7\% success on held-out simulated objects. Removing force-closure refinement drops success by 57.8 points, and removing the human prior drops it by 43.6 points, showing that policy performance depends on demonstration quality rather than kinematic paths alone. Scene-only observations and MLP pooling also underperform, indicating the value of point clouds and action queries. Protocol variants are reported in \Cref{sec:main_experiment_protocols,tab:policy_protocol,tab:geort_ik_ablation}.

\subsection{Dexterous Prehensile Manipulation}
\label{sec:extended_evidence}

\begin{figure}[H]
    \centering
    \includegraphics[width=\textwidth]{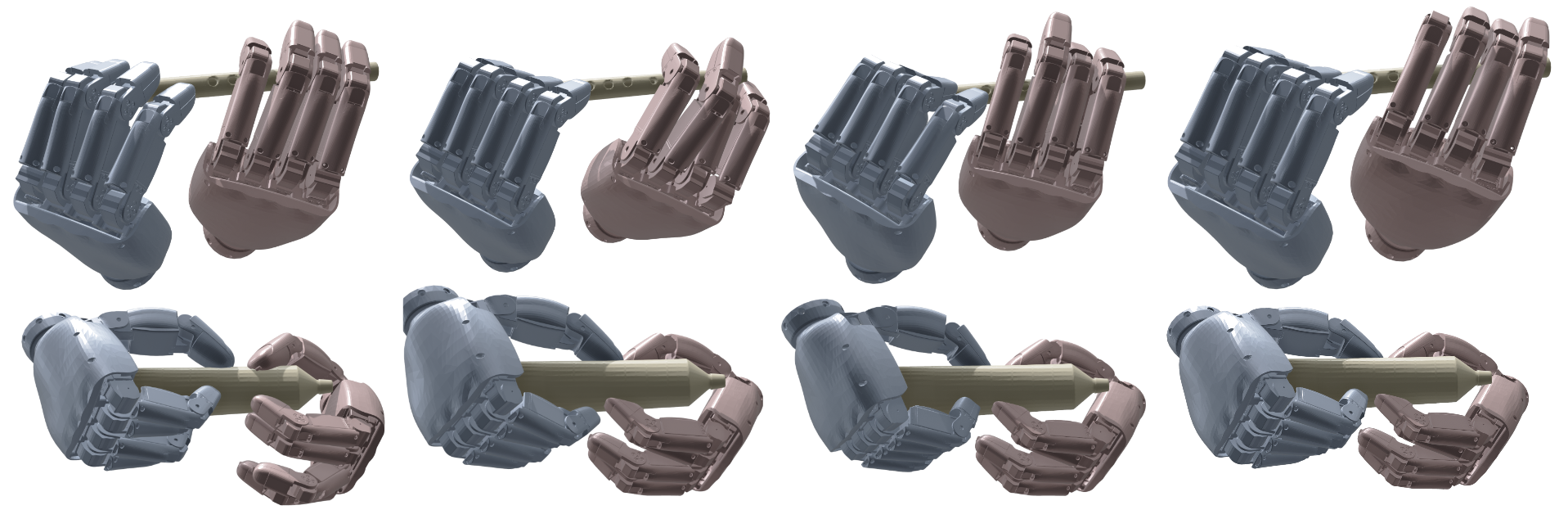}
    \caption{\textbf{In-grasp reconfiguration from validated keyframes.} The simulated sequences show asynchronous bimanual coordination: one hand maintains possession while the other changes object-relative contact or applies a directional manipulation force. These examples correspond to the transition-success evaluation in \cref{tab:ingrasp}.}
    \label{fig:ingrasp_reconfiguration}
\end{figure}

\begin{figure}[H]
    \centering
    \includegraphics[width=\textwidth]{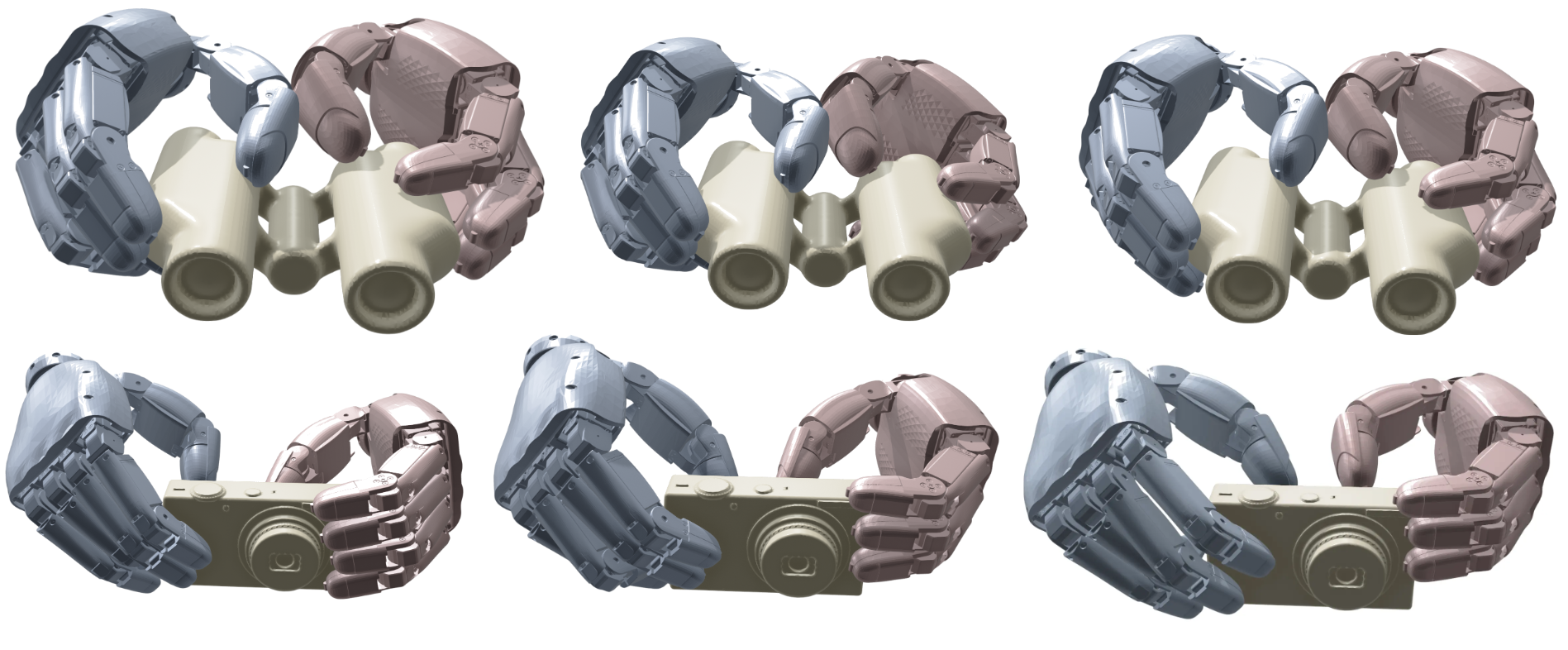}
    \caption{\textbf{Prehensile keyframes.} Accepted \method{} grasps on binoculars and cameras use dual-hand contact patterns conditioned on human priors object geometry, rather than generic power grasps.}
    \label{fig:prehensile_keyframes}
\end{figure}

\begin{figure}[H]
    \centering
    \includegraphics[width=\textwidth]{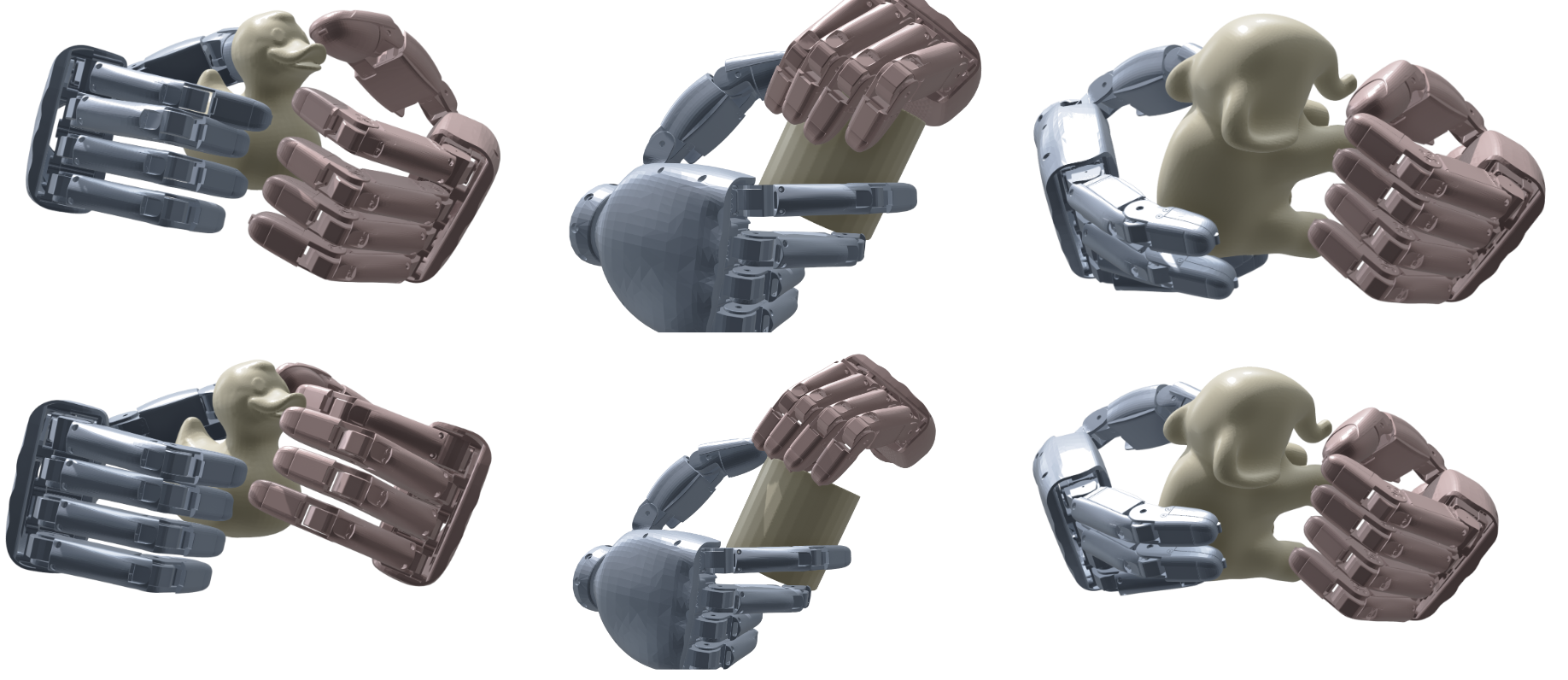}
    \caption{\textbf{Handover bimanual graspss.}
    The examples establish compatible dual contacts on objects to grasp and handover from one hand to another. \Cref{tab:handover_pick_place} evaluates whether these pre-release keyframes maintain possession through left-to-right and right-to-left self-handover rollouts.}
    \label{fig:handover_keyframes}
\end{figure}

\begin{figure}[H]
    \centering
    \includegraphics[width=\textwidth]{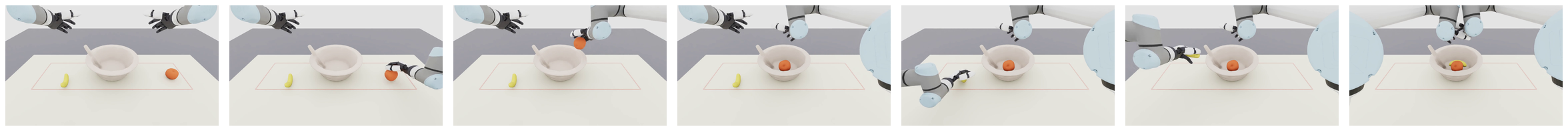}
    \caption{\textbf{Pick-and-place rollout sequence.}
    Temporal frames show approach, grasp, lift, transport to the target region, and terminal release or stabilization. \Cref{tab:handover_pick_place} reports the results where a trial must maintain possession through the placement check.}
    \label{fig:pick_place}
\end{figure}

RQ4 tests whether validated grasp keyframes remain useful beyond vertical lifting. This subsection covers selected in-grasp reconfiguration, self-handover, and pick-and-place rollouts; \Cref{fig:ingrasp_reconfiguration,fig:prehensile_keyframes,fig:handover_keyframes,fig:pick_place} show the corresponding execution traces.

\begin{table}[H]
\centering
\scriptsize
\setlength{\tabcolsep}{2.5pt}
\renewcommand{\arraystretch}{1.08}
\caption{\textbf{Simulated in-grasp manipulation stress test.} Metrics separate initial keyframe stability from contact-maintaining object reconfiguration; this is a selected transition protocol rather than broad manipulation coverage.}
\label{tab:ingrasp}
\begin{tabular*}{\textwidth}{@{\extracolsep{\fill}}lcccc@{}}
\toprule
& Initial state & \multicolumn{3}{c}{Transition} \\
\cmidrule(lr){2-2}\cmidrule(l){3-5}
Method & Keyframe Stability (\%) $\uparrow$ & Success (\%) $\uparrow$ & Slip (cm) $\downarrow$ & Final check (\%) $\uparrow$ \\
\midrule
Retarget-only & 14.6 & 8.3 & 5.8 & 6.2 \\
Static force-closure grasp & 82.1 & 41.7 & 2.9 & 35.4 \\
\rowcolor{dexfill} \textbf{\method{}} & \textbf{87.5} & \textbf{70.8} & \textbf{1.1} & \textbf{66.7} \\
\bottomrule
\end{tabular*}
\end{table}

\Cref{tab:ingrasp} separates static keyframe quality from dynamic transition success. Retarget-only grasps are semantically plausible but not load-bearing; static force-closure grasps stabilize the initial state but slip during reconfiguration. \method{} improves transition success by aligning stabilizing and active hand roles before rollout.

\paragraph{From Reconfiguration to Dynamic Transfer.}
Release transitions are a stricter test because one hand must disengage while the other assumes full possession and supports downstream placement. \Cref{fig:pick_place} illustrates the pick-and-place suite, and \Cref{tab:handover_pick_place} tracks the sequence from pre-release feasibility to terminal possession. We report Keyframe Validity for the pre-release grasp and End-to-End Success for the complete release, transfer, and placement rollout; trial definitions are specified in \Cref{sec:main_experiment_protocols}.

\begin{table}[H]
\centering
\scriptsize
\setlength{\tabcolsep}{2.5pt}
\renewcommand{\arraystretch}{1.08}
\caption{\textbf{Simulated self-handover and pick-and-place stress tests.} Keyframe validity measures pre-release feasibility; end-to-end success measures closed-loop possession through the terminal action. These selected transition protocols are mapped to RQ4 in \cref{tab:rq_map}.}
\label{tab:handover_pick_place}
\begin{tabular*}{\textwidth}{@{\extracolsep{\fill}}llcccc@{}}
\toprule
Suite & Protocol & Trials & Keyframe valid (\%) $\uparrow$ & E2E success (\%) $\uparrow$ & Main failure slice \\
\midrule
\multirow{2}{*}{Bimanual handover}
& self handover, left$\rightarrow$right release & 24 & 87.5 & 70.8 & release slip \\
& self handover, right$\rightarrow$left release & 24 & 83.3 & 66.7 & asymmetric contact \\
\midrule
\multirow{3}{*}{Pick-and-place}
& tabletop pick $\rightarrow$ target zone & 24 & 91.7 & 75.0 & placement overshoot \\
& offhand support $\rightarrow$ placement & 24 & 87.5 & 70.8 & late offhand release \\
& cluttered pickup $\rightarrow$ bin drop & 24 & 75.0 & 58.3 & partial occlusion \\
\bottomrule
\end{tabular*}
\end{table}

The main failures are release slip before the receiving hand has full support, asymmetric bilateral contacts, placement overshoot after possession, and partial occlusion in clutter. Within these selected protocols, validated \method{} keyframes are more reliable than retarget-only and static FC baselines.

\subsection{Real-world Experiments}
\label{sec:real_system_experiments}

Hardware transfer provides the real-system component of RQ4. The bimanual UR5e-XHand platform uses the same fused point-cloud and online replanning interface as simulation (\Cref{sec:main_experiment_protocols}). The quantitative benchmark uses three everyday objects with ten trials each; a trial succeeds only if the system establishes contact, lifts or transports the object, and maintains terminal possession. \method{} succeeds in 25/30 trials (83.3\%; Wilson 95\% CI: 66.4--92.7; 8/10 vase, 8/10 apple, 9/10 spray bottle). Failures come from rim slip, rotational shift at contact, and handle occlusion.

\begin{figure}[H]
    \centering
    \includegraphics[width=\textwidth]{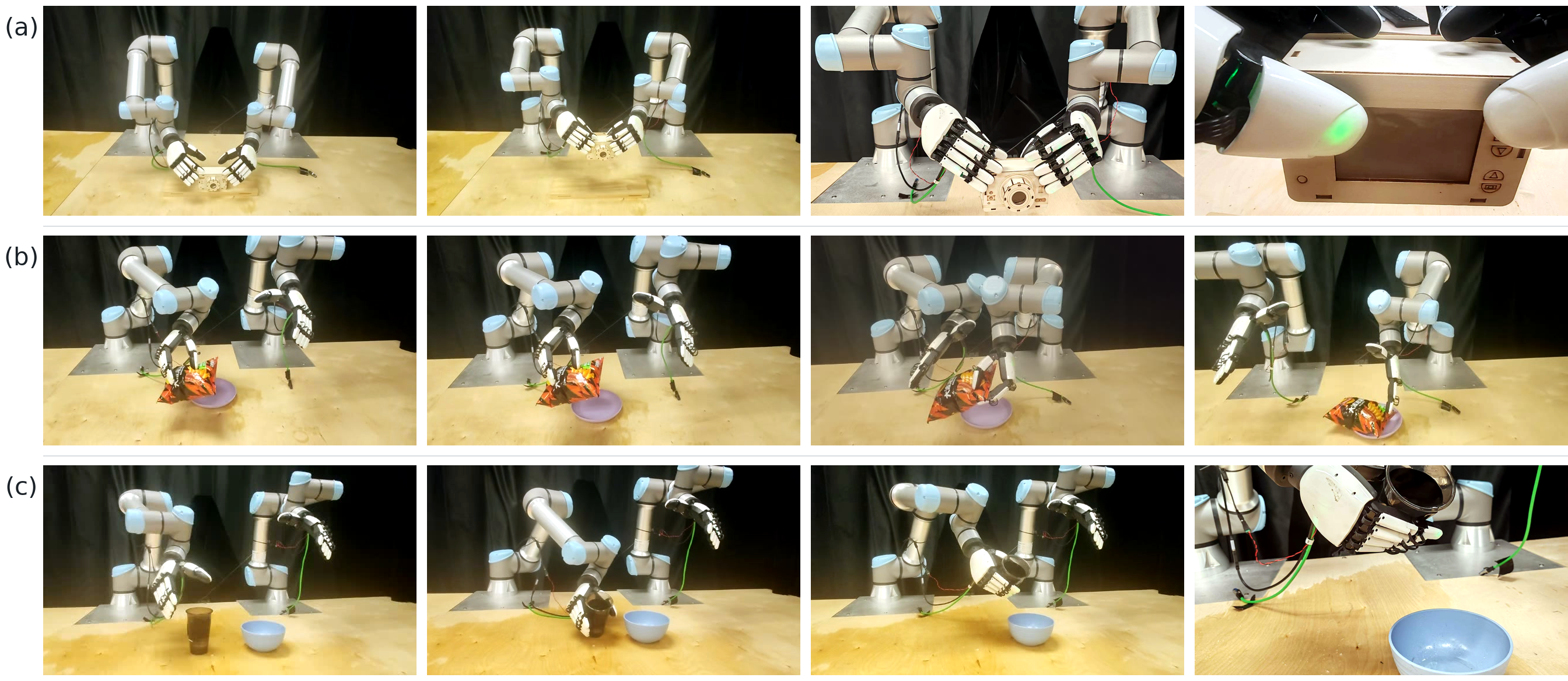}
    \caption{\textbf{Prehensile manipulation.}
    Rows show (a) toy-camera lifting, (b) pick-handover-place, and (c) pouring. These video-keyframe examples test functional terminal states and staged transfer, but they are not included in the quantitative success counts in \cref{tab:hardware}.}
    \label{fig:hardware_functional}
\end{figure}

\Cref{fig:hardware_functional} adds qualitative trials outside the 30-trial count: toy-camera lifting, pick-handover-place, and tilted pouring. These examples test transfer, release, terminal placement, and possession under a functional object pose. \Cref{tab:hardware} isolates the role of demonstration quality under the same hardware reset and evaluation protocol.

\begin{table}[H]
\centering
\scriptsize
\setlength{\tabcolsep}{2.5pt}
\renewcommand{\arraystretch}{1.08}
\caption{\textbf{Data comparison.} Each policy is trained with a different demonstration source and evaluated on the same 30 physical trials: 10 each for vase, apple, and spray bottle. The qualitative functional rollouts in \cref{fig:hardware_functional} are outside these counts.}
\label{tab:hardware}
\begin{tabular*}{\linewidth}{@{\extracolsep{\fill}}lcccc@{}}
\toprule
Data source & Trials & Success $\uparrow$ & Lift valid $\uparrow$ & Dominant failure mode \\
\midrule
Retarget-only data-based policy & 30 & 5/30 (16.7\%) & 6/30 & unstable contact \\
Optimization-only data-based policy & 30 & 11/30 (36.7\%) & 13/30 & poor approach alignment \\
\rowcolor{dexfill} \textbf{\method{} data-based policy} & 30 & \textbf{25/30 (83.3\%)} & \textbf{26/30} & occlusion / slip \\
\bottomrule
\end{tabular*}
\end{table}

The hardware ablation confirms the data-source effect: full \method{} demonstrations produce 25/30 successes, compared with 5/30 for retarget-only data and 11/30 for optimization-only data.

\subsection{Task Generation from Grasp Keyframes}
\label{sec:vlm_agent_exp}

As a final utility test, we feed validated \method{} keyframes to a VLM agent that proposes task-level manipulation specifications. The context packet includes multi-view renders, object metadata, optimized wrist-hand poses, contact regions, hand-role candidates, admission metrics, and the allowed primitive library (\cref{app:vlm_agent}). The agent emits a JSON specification with the functional goal, hand roles, object-relative waypoints, release conditions, and terminal checks. A proposal is accepted only if the deterministic executor converts it to an IK-feasible, collision-free rollout that maintains possession and satisfies the task predicate. This interface generates tasks such as teapot pouring, camera aiming, and flute finger-release variants, turning validated keyframes into a grounded substrate for downstream VLM-based task generation.

\subsection{Cross-Embodiment Experiment}
\label{sec:cross_embodiment_experiments}

We answer whether the same pre-grasp seed distribution helps a different hand by extending the framework to Shadow Hand. We replace the standard initialization of a BODex Shadow Hand optimizer~\cite{chen2025bodex} with a MANO$\rightarrow$Shadow geometric seed while keeping the object manifest, optimizer budget, collision parameters, and force-closure criteria fixed. This seed-only change increases valid grasps from 96/384 to 142/384, raises stability from 44.3\% to 61.5\%, and reduces penetration from 1.8\,mm to 1.2\,mm. The result isolates seed quality in contact-basin search; full morphology-agnostic policy transfer remains future work. Protocol details are in \Cref{app:shadow_cross_embodiment}.

\begin{figure}[H]
    \centering
    \includegraphics[width=.86\textwidth]{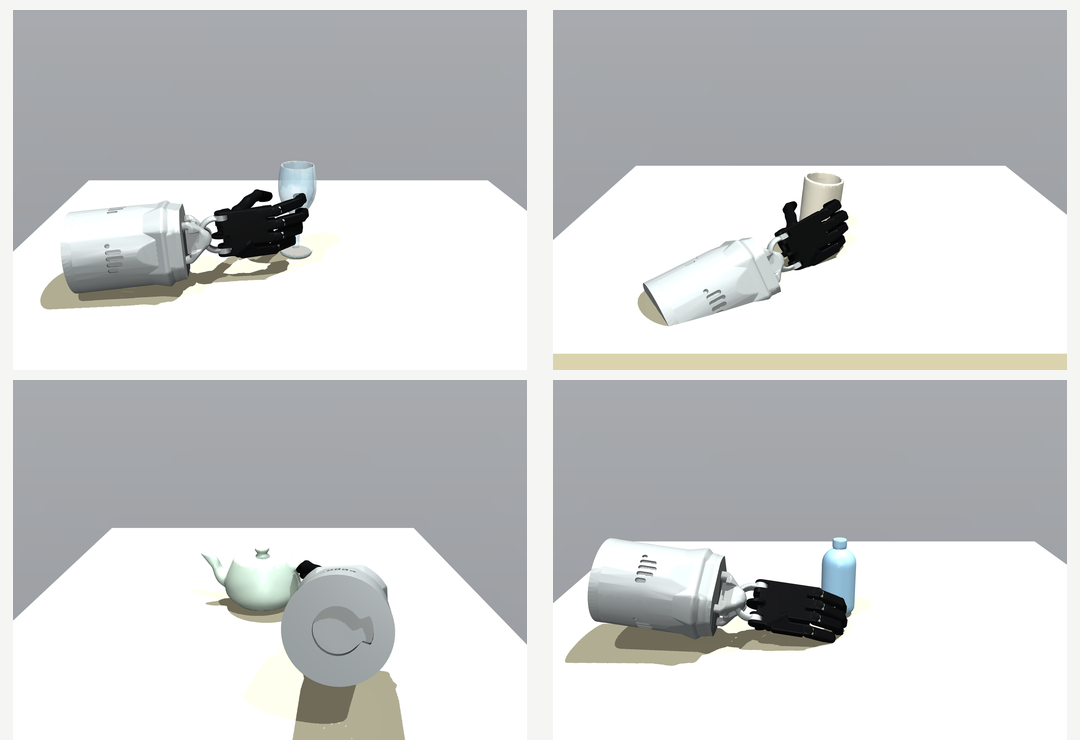}
    \caption{\textbf{Shadow hand grasps with \method{}.}
    MANO$\rightarrow$Shadow seeds initialize the Shadow wrist and fingers near object-relevant contact regions before BODex refinement. \Cref{tab:shadow_seed_diagnostic} changes only the seed distribution under the same object manifest, solver, and restart budget; this is a proposal-basin diagnostic, not cross-embodiment policy transfer.}
    \label{fig:shadow_seed_diagnostic}
\end{figure}

\begin{table}[H]
\centering
\small
\setlength{\tabcolsep}{3.5pt}
\caption{\textbf{Shadow-hand experiment results.} Both rows use the same BODex solver, object manifest, and restart budget; only the seed changes. G1 is $Q_\text{FC}{\times}10^{-2}$ and penetration is in mm.}
\label{tab:shadow_seed_diagnostic}
\begin{tabular}{lcccccc}
\toprule
Method & Seed & Retarget err. $\downarrow$ & Valid grasps $\uparrow$ & FC (\%) $\uparrow$ & G1 $\uparrow$ & Pen. $\downarrow$ \\
\midrule
BODex-Shadow random & native BODex & n/a & 96/384 & 44.3 & 3.8 & 1.8 \\
\rowcolor{dexfill} SynManDex-Shadow seeded & MANO$\rightarrow$Shadow & 7.6\,mm & \textbf{142/384} & \textbf{61.5} & \textbf{5.1} & \textbf{1.2} \\
\bottomrule
\end{tabular}
\end{table}

MANO-to-Shadow seeding improves the matched BODex Shadow optimizer under this seed-only diagnostic.

\subsection{Discussions}
\label{sec:limitations}

The experiments point to a natural path for scaling \method{} into a general-purpose bimanual dexterous manipulation data engine. Since each grasp is grounded by physical validations, the same pipeline could potentially be expanded along object diversity, scene complexity, and task semantics without changing the core formulation. In future work, we plan to scale agentic task generation from selected keyframe-conditioned proposals to broader libraries of object functions, hand-role assignments, release schedules, and terminal predicates. In addition, it is promising to increase the data across cluttered scenes, tool-use cases, handovers, and contact-rich in-grasp transitions.

\section{Conclusion}
\label{sec:conclusion}

We presented \method{}, a synthetic pipeline for human-to-robot dexterous grasp generation. By using a digital human diffusion model as a pre-grasp generator, \method{} improves grasp quality and human-likeness over retargeting-only or optimization-only alternatives. Beyond these results, \method{} proposes a practical route to scalable data generation and evaluation. Together, these results suggest that combining human-prior generative models with robot-native optimization can provide a useful foundation for scalable data generation and evaluation for general-purpose bimanual dexterous manipulation.

\clearpage
\bibliographystyle{unsrtnat}
\bibliography{references}

\clearpage
\appendix
\input{appendix}

\end{document}

%% file: table_capability.tex
\definecolor{natgreen}{HTML}{3A7D5A}
\definecolor{natblue}{HTML}{2F6F9F}
\definecolor{natviolet}{HTML}{7356A8}
\definecolor{natamber}{HTML}{A66A2C}
\definecolor{natgray}{HTML}{6B7280}
\definecolor{natfocus}{HTML}{EDF3EF}
\definecolor{natgroup}{HTML}{F1F3F5}
\definecolor{natline}{HTML}{B8C1BB}

\newcommand{\capfilled}[1]{\textcolor{#1}{\rule{0.34em}{0.66em}}}
\newcommand{\capempty}{\textcolor{natgray!35}{\rule{0.34em}{0.66em}}}
\newcommand{\capbar}[2]{
  \ifcase#1
    \textcolor{natgray}{--}
  \or
    \capfilled{#2}\capempty\capempty\capempty\capempty
  \or
    \capfilled{#2}\capfilled{#2}\capempty\capempty\capempty
  \or
    \capfilled{#2}\capfilled{#2}\capfilled{#2}\capempty\capempty
  \or
    \capfilled{#2}\capfilled{#2}\capfilled{#2}\capfilled{#2}\capempty
  \or
    \capfilled{#2}\capfilled{#2}\capfilled{#2}\capfilled{#2}\capfilled{#2}
  \fi}
\newcommand{\phys}[1]{\capbar{#1}{natgreen}}
\newcommand{\sem}[1]{\capbar{#1}{natblue}}
\newcommand{\agent}[1]{\capbar{#1}{natviolet}}
\newcommand{\infercap}[1]{\capbar{#1}{natamber}}
\newcommand{\nocap}{\textcolor{natgray}{--}}
\newcommand{\grouphead}[1]{\rowcolor{natgroup}\multicolumn{8}{@{}l}{\textit{#1}}\\[-1pt]}
\newcolumntype{P}[1]{>{\raggedright\arraybackslash}p{#1}}

\begin{table}[H]
\centering
\scriptsize
\setlength{\tabcolsep}{1.7pt}
\renewcommand{\arraystretch}{1.12}
{\arrayrulecolor{natline}
\begin{tabular*}{\textwidth}{@{\extracolsep{\fill}}P{0.25\textwidth}ccccccc@{}}
\toprule
\textbf{Method family} &
\shortstack{\textbf{Phys.}\\\textbf{guid.}} &
\shortstack{\textbf{Sem.}\\\textbf{guid.}} &
\shortstack{\textbf{Task}\\\textbf{orient.}} &
\shortstack{\textbf{Cross}\\\textbf{embod.}} &
\shortstack{\textbf{Agentic}\\\textbf{compat.}} & \shortstack{\textbf{Bimanual}\\\textbf{manip.}}  &
\shortstack{\textbf{H2R}\\\textbf{data}} \\
\midrule
Physics-first grasp synthesis~\citep{wang2022dexgraspnet,chen2025bodex,zurbrugg2025graspqp} & \phys{5} & \nocap & \infercap{1} & \phys{4} & \agent{1} & \nocap & \nocap \\
Semantic / task-oriented grasp synthesis~\citep{chen2025dexonomy,wei2025afforddexgrasp} & \phys{3} & \sem{5} & \sem{4} & \phys{3} & \agent{1} & \nocap & \sem{2} \\
Bimanual robot grasp/data systems~\citep{shao2024bimangrasp,lin2026bidexgrasp,yang2025ultradexgrasp} & \phys{5} & \sem{1} & \infercap{2} & \phys{2} & \agent{4} & \phys{5} & \agent{2} \\
Human-to-robot manipulation transfer~\citep{li2025maniptrans} & \phys{4} & \sem{2} & \sem{4} & \phys{3} & \agent{5} & \phys{5} & \agent{5} \\
\rowcolor{natfocus}
\textbf{\method{} (ours)} & \textbf{\phys{5}} & \textbf{\sem{5}} & \textbf{\sem{4}} & \textbf{\phys{4}} & \textbf{\agent{5}} & \textbf{\phys{5}} & \textbf{\agent{5}} \\
\bottomrule
\end{tabular*}}
\vspace{3pt}
\caption{\textbf{Comparison of dexterous grasp generation methods.} We summarize and extend the discussions in \cref{sec:related_work} by comparing \method{} with reprensetative existing 4 families of method with 7 dimensions. Each cell reports an ordinal 0--5 capability score: 5 indicates a core contribution with direct validation in the modality required by the column, 3 indicates explicit but partially validated support, 1 indicates incidental relevance, and -- indicates outside scope. 
Color denotes evidence type rather than score magnitude: green for robot-native physical grounding or execution, blue for semantic, task, or human-intent guidance, violet for trajectory, policy, or downstream-agent compatibility, and amber for task relevance inferred from static grasp coverage.}
\label{tab:capability_stack}
\end{table}

%% file: appendix.tex
This supplementary material records the details behind the main-paper results. It contains the full problem formulation and algorithm (\cref{app:problem_algorithm}), method details omitted from the main text (\cref{app:full_method_details}), qualitative diagnostics for unimanual grasps and flute-holding release priors (\cref{sec:supp_unimanual,sec:supp_flute_taxonomy}), and a fixed-manifest Shadow-hand study that evaluates whether MANO priors remain useful under a non-XHand embodiment (\cref{app:shadow_cross_embodiment}). It also gives extended related work, experiment protocols, domain randomization, policy details, additional failure analysis, and the VLM-agent context, prompt, and trajectory proposal schema (\cref{app:vlm_agent}).

\section{Full Problem Definition and Pipeline Algorithm}
\label{app:problem_algorithm}

Let $\mathbf{q}_l = (\mathbf{T}_l, \boldsymbol{\theta}_l)$ denote the configuration of manipulator $l \in \{L, R\}$, where $\mathbf{T}_l \in SE(3)$ is the wrist pose and $\boldsymbol{\theta}_l \in \mathbb{R}^{n_\text{hand}}$ the hand joint configuration.
Given an object mesh $\mathcal{M}$, \method{} formulates human-like dexterous grasping over the bimanual space $\qbi = \qsingle_L \times \qsingle_R$ as a staged proposal-and-filtering problem rather than a single joint solver:
\begin{align}
    \mathbf{h}_0 &\sim p_\theta(\mathbf{h}\mid \mathcal{M}), \nonumber\\
    \mathbf{q}_{\mathrm{init}} &= R_\psi(\mathbf{h}_0,\mathcal{M}), \nonumber\\
    \mathbf{q}^* &= \arg\min_{\mathbf{q}\in\qbi}
    w_c C_{\mathrm{coll}}(\mathbf{q},\mathcal{M})
    +w_f\mathcal{L}_{\mathrm{FC}}(\mathbf{q},\mathcal{M})
    +w_r\|\mathbf{q}-\mathbf{q}_{\mathrm{init}}\|_2^2 .
    \label{eq:app_dual_obj}
\end{align}
Here $\mathbf{h}_0$ is a generated MANO pre-grasp, $R_\psi$ maps it to a robot seed, and the final optimization is performed in robot configuration space.
Human grasp priors are therefore used as initialization rather than as executable labels.
Compared with random initialization, a human-prior seed starts in a functional region of $\qbi$, after which contact, collision, force-closure, IK, and rollout checks decide whether a sample is admitted.
\method{} operationalizes this insight in three stages (\cref{fig:pipeline}): synthesizing human pre-grasps (\cref{sec:pregrasp}), retargeting and force-closure optimization (\cref{sec:retarget_opt}), and trajectory generation with policy training (\cref{sec:traj_policy}).

\begin{algorithm}[t]
\caption{\method{} as proposal, refinement, and executable filtering.}
\label{alg:synmandex}
\begin{algorithmic}[1]
\REQUIRE Object mesh $\mathcal{M}$, number of human-prior samples $N$, lift threshold $\tau_z$
\STATE Sample MANO pre-grasps $\{\mathbf{h}_0^i\}_{i=1}^{N} \sim p_\theta(\mathbf{h}_0 \mid \mathcal{M})$
\FOR{$i=1,\ldots,N$}
    \STATE Retarget $\mathbf{h}_0^i$ to an XHand seed $\mathbf{q}_{\text{init}}^i$ with GeoRT-calibrated geometry (\cref{eq:geort_train,eq:retarget})
    \STATE Refine $\mathbf{q}_{\text{init}}^i$ into $\mathbf{q}^{i*}$ by minimizing collision, force-closure, and seed-regularization losses (\cref{eq:opt})
    \STATE Reject $\mathbf{q}^{i*}$ if floating-hand stability, penetration, or force-closure checks fail
    \STATE Solve GeoRT-guided arm-hand IK on the same UR5e+XHand model used in execution (\cref{eq:geort_ik})
    \STATE Execute approach--close--squeeze--lift; retain the trajectory if $\max_{t \ge t_{\text{lift}}}(z_t-z_0)>\tau_z$
\ENDFOR
\RETURN IK-validated demonstrations $\mathcal{D}$ with source provenance and failure labels
\end{algorithmic}
\end{algorithm}

\section{Method Details}
\label{app:full_method_details}

\begin{figure}[htbp]
    \centering
    \includegraphics[width=0.78\linewidth]{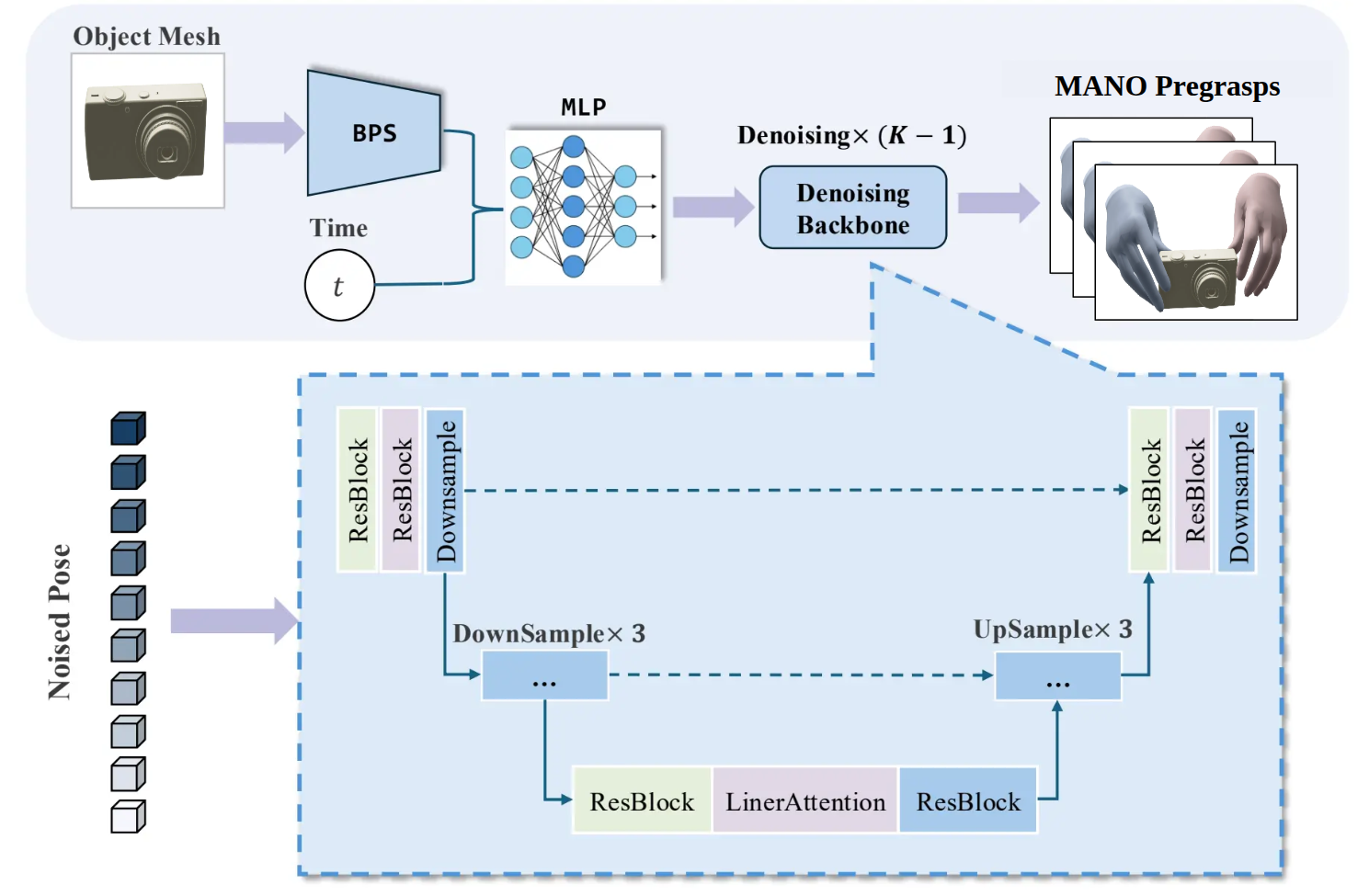}
    \caption{\textbf{Diffusion model architecture.} The object mesh is encoded with a point-based representation and, together with the diffusion timestep, conditions a U-Net denoising backbone that predicts digital human pre-grasp parameters. This figure expands Stage A of \cref{fig:pipeline}; training loss details are given in \cref{eq:app_ddpm_step,eq:app_ddpm_loss}.}
    \label{fig:human_diffusion_architecture}
\end{figure}

\subsection{Digital Human Pre-Grasp Diffusion}
\label{app:diffusion_details}

At each diffusion step, diffusion model estimates the distribution of a less-noisy MANO sample from the current noisy sample and object mesh:
\begin{equation}
    p_\theta(\mathbf{h}_{t-1} \mid \mathbf{h}_t, \mathcal{M}) =
    \mathcal{N}(\mathbf{h}_{t-1};\, \boldsymbol{\mu}_\theta(\mathbf{h}_t,t,\mathcal{M}),\, \boldsymbol{\Sigma}_\theta(\mathbf{h}_t,t,\mathcal{M})).
    \label{eq:app_ddpm_step}
\end{equation}
Using the standard DDPM reparameterization~\cite{ho2020denoising}, the network predicts noise:
\begin{equation}
    \mathcal{L}_\text{diff} =
    \mathbb{E}_{t,\mathbf{h}_0,\boldsymbol{\epsilon}}
    \left[
    \left\|
    \boldsymbol{\epsilon}
    -
    \boldsymbol{\epsilon}_\theta(
    \sqrt{\bar{\alpha}_t}\mathbf{h}_0
    +
    \sqrt{1-\bar{\alpha}_t}\boldsymbol{\epsilon},t,\mathcal{M})
    \right\|_2^2
    \right].
    \label{eq:app_ddpm_loss}
\end{equation}
The model generates a single pre-contact frame. Downstream retargeting, contact optimization, and trajectory generation are responsible for robot feasibility.

\subsection{Retargeting Objective Terms}
\label{app:retarget_details}

\begin{figure}[t]
    \centering
    \includegraphics[width=0.82\linewidth]{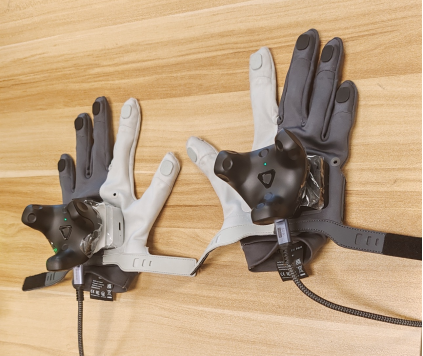}
    \caption{\textbf{Manus pro glove for geometric retargting.}
    Manus Pro glove or MANO pre-grasps provide a functional seed: keypoint and fingertip correspondences map the human approach, wrist pose, and coarse finger coordination into an XHand robot pre-grasp.}
    \label{fig:retargeting_gloves}
\end{figure}

The compact retargeting objective in \cref{eq:geort_train} expands into five geometric criteria. Motion preservation aligns source and target keypoint displacement directions:
\begin{equation}
    \mathcal{L}_{\text{dir}}
    =
    \sum_{i=1}^{K}
    \mathbb{E}_{\mathbf{X}^{H},\boldsymbol{\delta}_i}
    \left[
    1-
    \frac{
    \Delta \mathbf{x}^{R}_i(\mathbf{X}^{H},\boldsymbol{\delta}_i)^\top \boldsymbol{\delta}_i
    }{
    \|\Delta \mathbf{x}^{R}_i(\mathbf{X}^{H},\boldsymbol{\delta}_i)\|_2\|\boldsymbol{\delta}_i\|_2+\epsilon
    }
    \right],
    \label{eq:app_geort_dir}
\end{equation}
where $\Delta \mathbf{x}^{R}_i(\mathbf{X}^{H},\boldsymbol{\delta}_i)=\mathbf{x}^{R}_i(g_\psi(\mathbf{X}^{H}{+}\boldsymbol{\delta}_i))-\mathbf{x}^{R}_i(g_\psi(\mathbf{X}^{H}))$.
Coverage uses a Chamfer proxy against a point-cloud approximation $\mathcal{P}^{R}_i$ of the XHand reachable keypoint space:
\begin{equation}
    \mathcal{L}_{\text{cov}}
    =
    \sum_{i=1}^{K}
    \operatorname{CD}\!\left(
    \{\mathbf{x}^{R}_i(g_\psi(\mathbf{X}^{H}_m))\}_{m=1}^{B},\,
    \mathcal{P}^{R}_i
    \right).
    \label{eq:app_geort_cov}
\end{equation}
Flatness penalizes nonuniform finite-difference response:
\begin{equation}
    \mathcal{L}_{\text{flat}}
    =
    \sum_{i=1}^{K}
    \mathbb{E}_{\mathbf{X}^{H},\boldsymbol{\delta}_i}
    \left[
    \left\|
    \mathbf{x}^{R}_i(g_\psi(\mathbf{X}^{H}{+}\boldsymbol{\delta}_i))
    -2\mathbf{x}^{R}_i(g_\psi(\mathbf{X}^{H}))
    +\mathbf{x}^{R}_i(g_\psi(\mathbf{X}^{H}{-}\boldsymbol{\delta}_i))
    \right\|_2^2
    \right].
    \label{eq:app_geort_flat}
\end{equation}
Pinch preservation keeps robot fingertips close when corresponding MANO fingertips are close:
\begin{equation}
    \mathcal{L}_{\text{pinch}}
    =
    \sum_{(i,j)\in\mathcal{P}_{\text{pinch}}}
    \mathbf{1}\!\left[\|\mathbf{x}^{H}_i-\mathbf{x}^{H}_j\|_2<\tau_H\right]
    \max\!\left(0,\|\mathbf{x}^{R}_i(g_\psi(\mathbf{X}^{H}))-\mathbf{x}^{R}_j(g_\psi(\mathbf{X}^{H}))\|_2-\tau_R\right)^2.
    \label{eq:app_geort_pinch}
\end{equation}
The self-collision term uses the same hand collision model used later for simulation filtering.

\subsection{Force Closure and Kinematics}
\label{app:optimization_details}

For $K$ contact points at positions $\{\mathbf{p}_k\}_{k=1}^{K}$ relative to the object center of mass, the grasp map is
\begin{equation}
    \mathbf{G} =
    \begin{bmatrix}
    \mathbf{I}_3 & \cdots & \mathbf{I}_3 \\
    [\mathbf{p}_1]_\times & \cdots & [\mathbf{p}_K]_\times
    \end{bmatrix}.
    \label{eq:app_grasp_matrix}
\end{equation}
$Q_{\text{FC}}>0$ certifies force closure under the discretized Coulomb friction cone and the bounded unit-force simplex in \cref{eq:fc}. We approximate friction cones as 8-sided polyhedra, normalize edge wrenches before computing the margin, and solve candidate refinements in parallel on GPU.

For GeoRT-IK, the wrist residual is
\begin{equation}
\mathcal{L}_{SE(3)}
=
\|\mathbf{p}^{A}_l(\mathbf{a}_l)-\mathbf{p}^{*}_l\|_2^2
+\lambda_R\|\logm(\mathbf{R}^{*}_l \mathbf{R}^{A}_l(\mathbf{a}_l)^\top)^\vee\|_2^2,
\label{eq:app_ik_pose}
\end{equation}
and pinch preservation maintains the fingertip pair distances from the optimized grasp:
\begin{equation}
    \mathcal{L}^{\text{IK}}_{\text{pinch}}
    =
    \sum_{(i,j)\in\mathcal{P}^{*}_{\text{pinch}}}
    \left(
    \left\|
    \mathbf{T}^{A}_l(\mathbf{a}_l)\mathbf{x}^{R}_i(\boldsymbol{\theta}_l)
    -
    \mathbf{T}^{A}_l(\mathbf{a}_l)\mathbf{x}^{R}_j(\boldsymbol{\theta}_l)
    \right\|_2
    -
    \|\mathbf{y}^{*}_{l,i}-\mathbf{y}^{*}_{l,j}\|_2
    \right)^2.
    \label{eq:app_ik_pinch}
\end{equation}
Solutions are rejected if wrist residual, keypoint residual, collision margin, or joint-limit slack exceed thresholds.

\section{Left Hand Grasp Generation}
\label{sec:supp_unimanual}

\begin{figure}[H]
    \centering
    \includegraphics[width=0.85\linewidth]{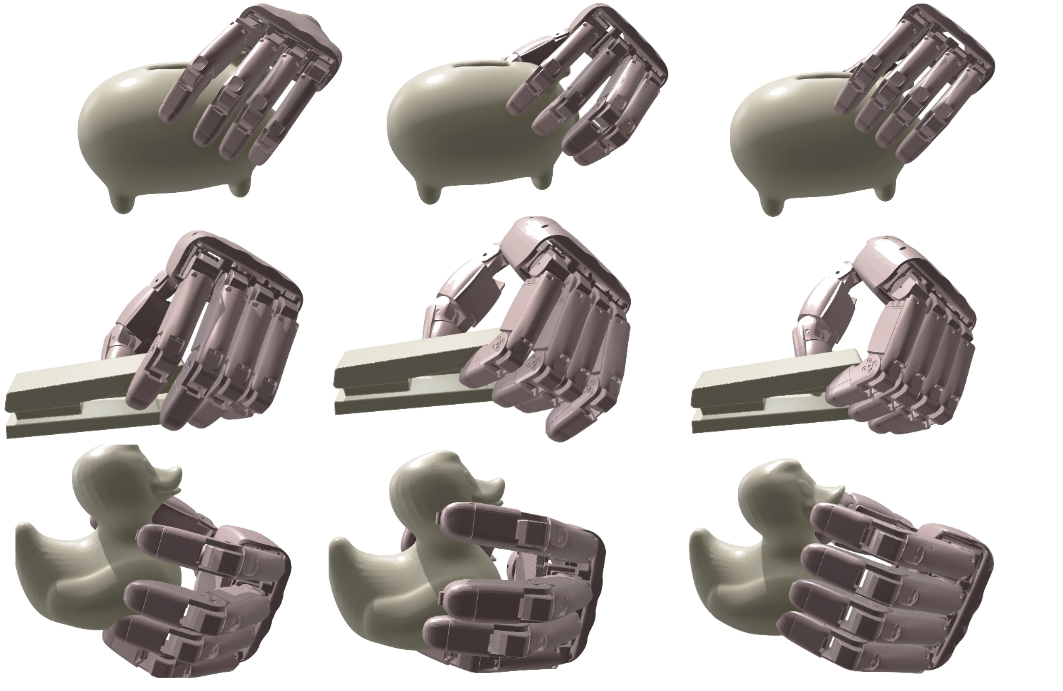}
    \caption{\textbf{Left-hand grasp-generation diagnostic.} Each row shows three accepted left-hand grasps on a piggy bank, stapler, and rubber duck. The examples show that the same human-prior-to-robot-grounding mechanism can operate on left hand setting.}
    \label{fig:unimanual_diagnostic}
\end{figure}

We demonstrated bimanual grasping and right-hand grasping in the main sections, and here we add qualitative results of left-hand grasping. \cref{fig:unimanual_diagnostic} shows that the same human-prior-to-physics mechanism also works in the single-hand regime.
The examples cover rounded, elongated, and irregular objects, producing palmar wrap, precision-power, and stabilizing grasps after force-closure refinement.
This result is kept as an appendix diagnostic: it supports the generality of the representation, while the main empirical claim remains bimanual dexterous grasp generation.

\section{A Grasp-based Flute-Playing Taxonomy}
\label{sec:supp_flute_taxonomy}

The main paper uses \cref{fig:flute_modes} to compare a generic bimanual flute hold against four \method{} flute-holding modalities. Furthermore, to better demonstrate the use of the generated poses, we organize a finger-playing taxonomy, as instrument playing could base on a human-like pose by releasing different fingers. Starting from a canonical two-hand flute grasp, each pose is labeled by the subset of fingers released from the instrument, such as $L{:}\{I,M,R\}, R{:}\{\}$ for releasing the left index, middle, and ring fingers.
This organization reflects a natural property of flute interaction: the hand must maintain instrument support while individual fingers change contact with the keys.
We use this taxonomy as an interpretable visualization of grasp-pose diversity, not as a model of musical performance.

The release labels are generated by parsing the released-finger tags associated with each pose.
For example, releasing the left index, middle, and ring fingers maps to the compact taxonomy key $L{:}\{I,M,R\}, R{:}\{\}$.

\begin{figure}[H]
    \centering
    \includegraphics[width=\textwidth]{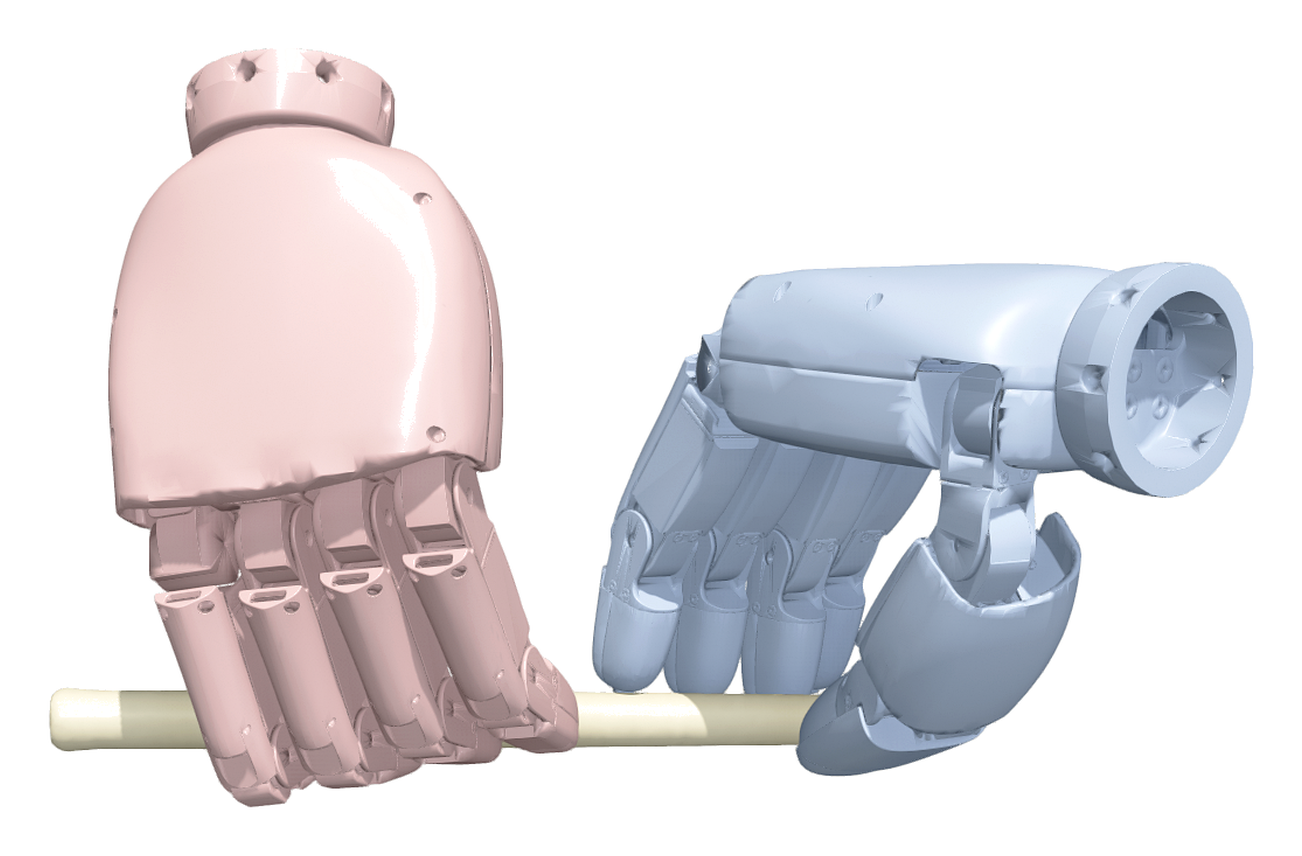}
    \caption{\textbf{Canonical flute-holding support pose.}
    The all-finger-contact pose provides the stable two-hand root configuration from which the release variants in \cref{fig:flute_release_taxonomy,fig:flute_release_variants} are organized.}
    \label{fig:flute_canonical}
\end{figure}

\begin{figure}[H]
    \centering
    \includegraphics[width=\textwidth]{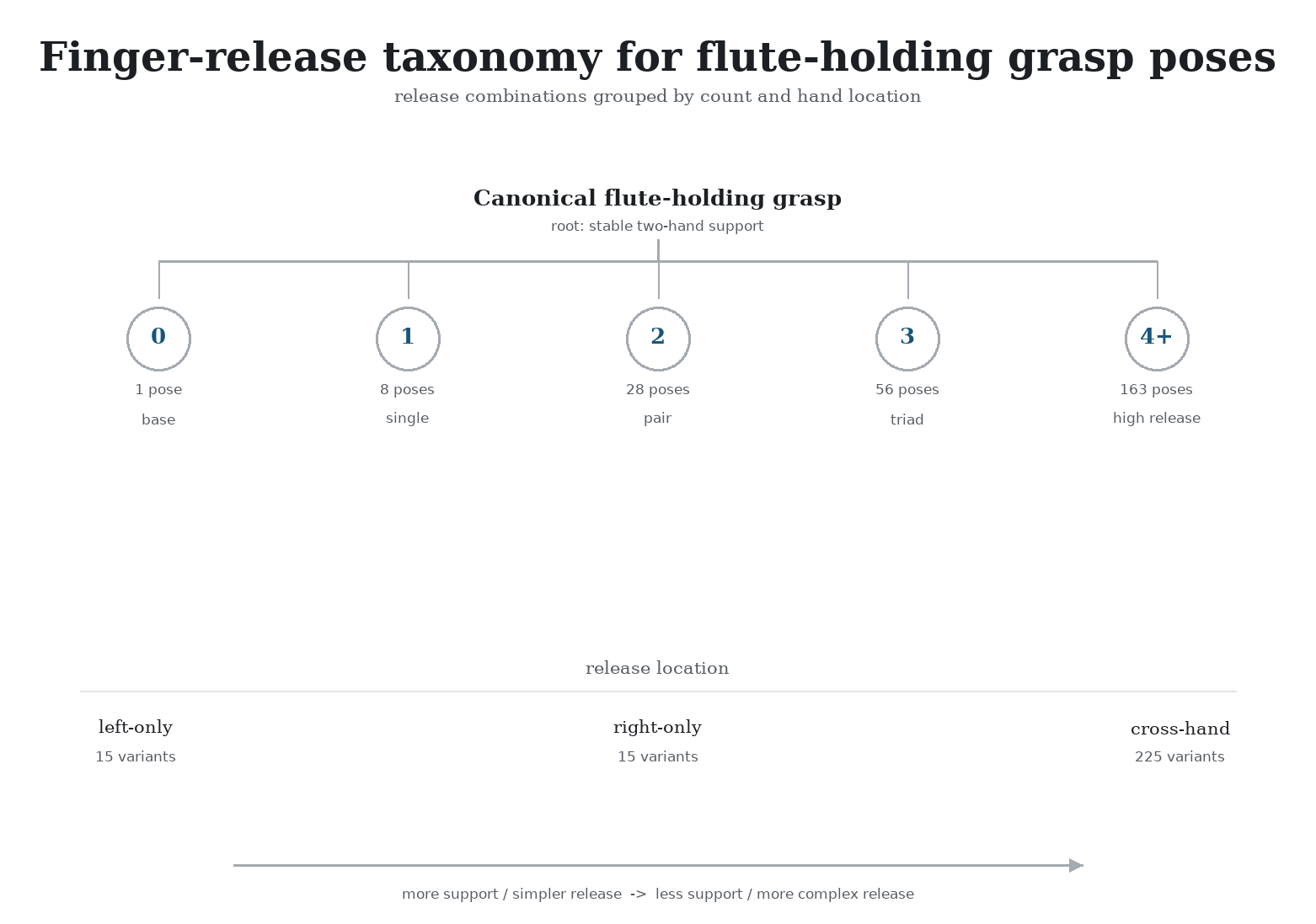}
    \caption{\textbf{Finger-release taxonomy for flute-holding poses.}
    Variants are grouped by release count and by whether released fingers belong to the left hand, right hand, or both hands. Labels such as $L{:}\{I,M,R\}, R{:}\{\}$ denote released index, middle, ring, and pinky fingers; the taxonomy expands the flute example in \cref{fig:flute_modes}.}
    \label{fig:flute_release_taxonomy}
\end{figure}

\begin{figure}[H]
    \centering
    \includegraphics[width=\textwidth]{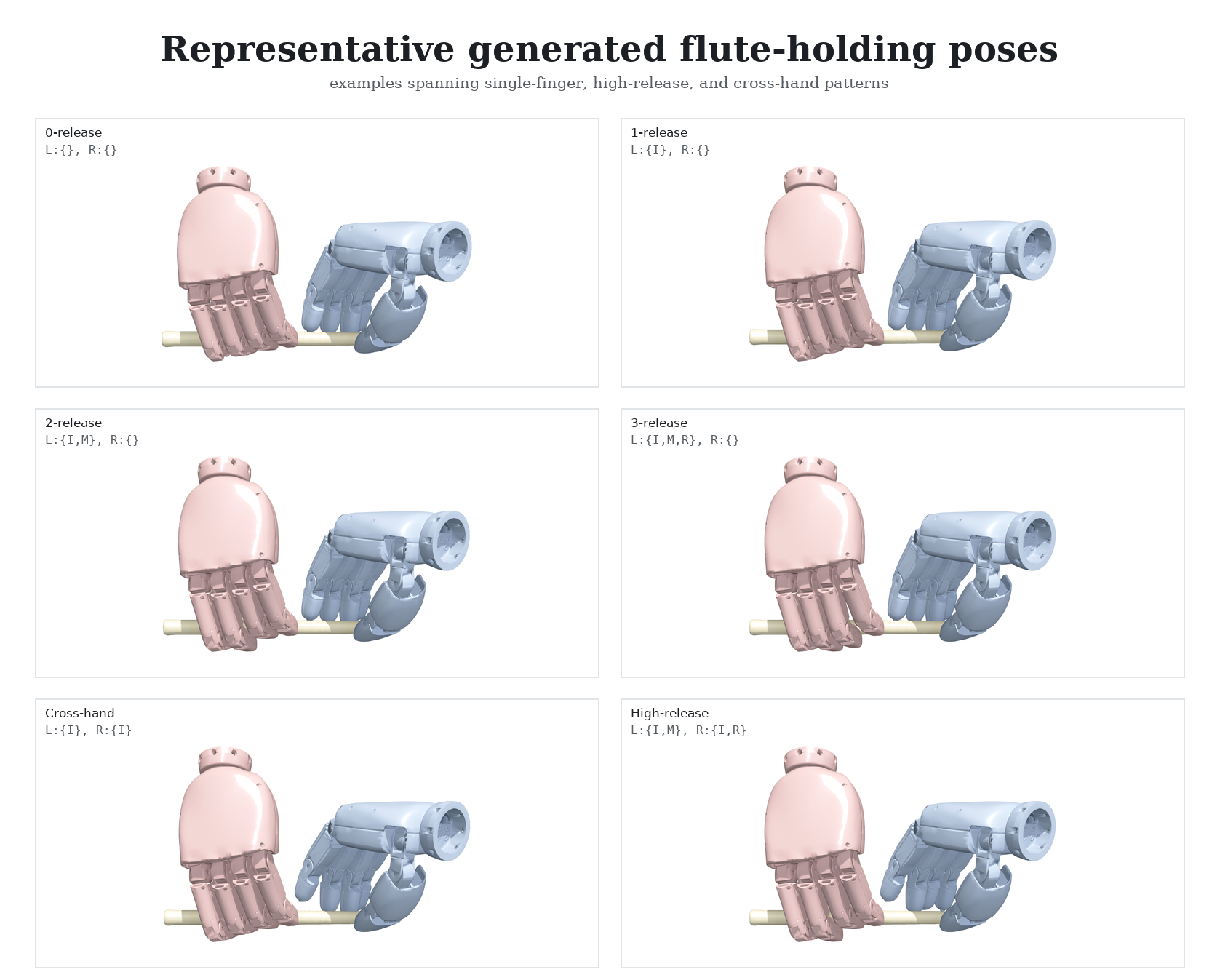}
    \caption{\textbf{Representative flute-holding release variants.}
    Panels sample the 0-release, 1-release, 2-release, 3-release, cross-hand, and high-release groups from the taxonomy in \cref{fig:flute_release_taxonomy}, illustrating structured finger-level diversity around the same supported flute pose.}
    \label{fig:flute_release_variants}
\end{figure}
\clearpage

\section{More Results with the BODex Shadow Hand}
\label{app:shadow_cross_embodiment}

The main paper evaluates \method{} on UR5e+XHand because that is the physical embodiment used for trajectory generation and real-system validation.
However, a stronger mechanistic test asks whether the human-prior proposal is tied to XHand, or whether it transfers to a different dexterous morphology.
We therefore add a matched Shadow-hand study using the native BODex Shadow configuration~\cite{chen2025bodex}.
This fixed-manifest study isolates one variable: \emph{seed distribution}.

\paragraph{Hypothesis.}
If \method{} uses MANO priors as geometry-aware pre-contact proposals, then a MANO pregrasp retargeted to the Shadow hand should improve the same BODex Shadow optimizer relative to its native random/geometric initialization, after controlling object mesh, object pose, scale, candidate budget, solver configuration, and evaluation metrics.
The falsifier is equally clear: if the seeded variant does not improve force closure, penetration, contact consistency, or executable yield under this matched protocol, then the current human-prior mechanism is XHand-specific or loses useful contact geometry under the MANO$\rightarrow$Shadow morphology gap.

\paragraph{Matched protocols.}
We compare two variants on the same GRAB object manifest:
\begin{itemize}
    \item \textbf{BODex-Shadow random}: BODex's native right Shadow-hand grasp synthesis using \texttt{sim\_shadow/fc.yml}, with its default seed generator and a fixed restart budget $B$.
    \item \textbf{SynManDex-Shadow seeded}: the same BODex Shadow solver and the same total restart budget $B$; the first restart is initialized by a MANO prior retargeted to the Shadow hand, and the remaining $B{-}1$ restarts use the same BODex seed generator as the baseline.
\end{itemize}
Both rows use the same object mesh $\mathcal{M}$, object pose $\mathbf{x}_o$, object scale, table configuration, collision geometry, optimizer iteration count, friction coefficient, and force-closure evaluator.
Thus the only controlled difference is whether the candidate set contains a human-prior seed.

\paragraph{MANO-to-Shadow seed construction.}
Let $\mathbf{P}^{H}=\{\mathbf{p}^{H}_k\}_{k=1}^{11}$ be the MANO landmarks used for retargeting: wrist/palm, five distal finger joints, and five intermediate finger joints.
Let $\ell_k$ be the corresponding Shadow links: \texttt{rh\_palm}, distal links \texttt{rh\_thdistal}, \texttt{rh\_ffdistal}, \texttt{rh\_mfdistal}, \texttt{rh\_rfdistal}, \texttt{rh\_lfdistal}, and middle links \texttt{rh\_thmiddle}, \texttt{rh\_ffmiddle}, \texttt{rh\_mfmiddle}, \texttt{rh\_rfmiddle}, \texttt{rh\_lfmiddle}.
We solve a position-retargeting problem over the floating Shadow root and the 22 actuated Shadow finger joints:
\begin{equation}
    (\mathbf{T}^{S}_{0},\boldsymbol{\theta}^{S}_{0})
    =
    \arg\min_{\mathbf{T},\boldsymbol{\theta}}
    \sum_{k=1}^{11}
    \rho\!\left(
    \left\|
    \FK^{S}_{\ell_k}(\mathbf{T},\boldsymbol{\theta})
    -
    \mathbf{p}^{H}_k
    \right\|_2
    \right)
    +
    \lambda_{\mathrm{rt}}
    \left\|
    \boldsymbol{\theta}-\bar{\boldsymbol{\theta}}
    \right\|_2^2,
    \label{eq:shadow_retarget}
\end{equation}
where $\rho$ is a Huber penalty and $\bar{\boldsymbol{\theta}}$ is the previous or neutral Shadow configuration.
Before optimization, we align the MANO palm landmark frame to the Shadow zero-pose palm landmark frame; this avoids the common error of treating the local axes of \texttt{rh\_palm} as if they were MANO axes.
The resulting seed is
\begin{equation}
    \mathbf{s}^{S}_{\mathrm{MANO}}
    =
    \left[
    \mathbf{t}^{S}_{\mathrm{palm}},
    \mathbf{r}^{S}_{\mathrm{palm}},
    \boldsymbol{\theta}^{S}_{0}
    \right]
    \in \mathbb{R}^{29},
    \label{eq:shadow_seed}
\end{equation}
where $\mathbf{r}^{S}_{\mathrm{palm}}$ is represented as a unit quaternion in $wxyz$ order.
BODex then runs its native Shadow-hand objective with either
\begin{equation}
    \mathcal{S}_{\mathrm{rand}}
    =
    \{\mathbf{s}_1,\ldots,\mathbf{s}_{B}\}
    \quad\text{or}\quad
    \mathcal{S}_{\mathrm{human}}
    =
    \{\mathbf{s}^{S}_{\mathrm{MANO}},\mathbf{s}_2,\ldots,\mathbf{s}_{B}\},
    \label{eq:shadow_seed_sets}
\end{equation}
holding $B$ fixed.

\paragraph{Metrics.}
We report both retargeting diagnostics and final grasp quality.
Retargeting diagnostics include mean and maximum landmark residual in \cref{eq:shadow_retarget}.
Final grasp metrics mirror the main paper: number of valid grasps returned, force-closure rate, G1 stability, penetration depth, contact-region count, and, when the BODex UR10e+Shadow motion-generation path is enabled, arm-reachable trajectory success.
A method-blinded VLM-H score can be computed from the same renderer used in \cref{sec:metric_protocol}. The primary diagnostic is physical proposal-basin transfer, with visual preference treated as secondary evidence.

\begin{figure}[t]
    \centering
    \includegraphics[width=0.86\linewidth]{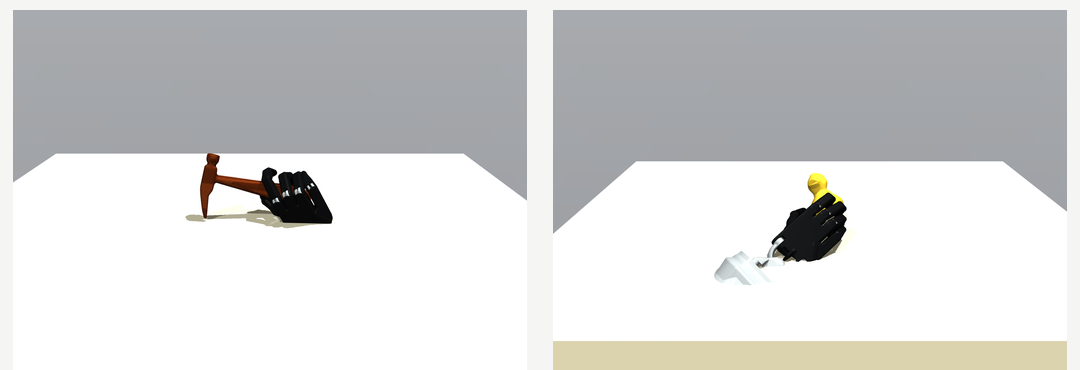}
    \caption{\textbf{Shadow-hand failure cases without human-prior seeding.}
    Random or geometric BODex initialization can satisfy local hand-object contact while producing wrist-table collisions or unusable approach poses. These cases motivate the seed-only comparison in \cref{tab:shadow_seed_diagnostic}.}
    \label{fig:shadow_failures}
\end{figure}

\paragraph{Visual diagnostics.}
\Cref{fig:shadow_seed_diagnostic} gives the human-seeded Shadow-hand visual results under the matched protocol above.
\Cref{tab:shadow_seed_diagnostic} reports the corresponding quantitative diagnostic.
The remaining diagnostic in \cref{fig:shadow_failures} shows why the seed distribution matters for BODex-Shadow: random/geometric initialization can satisfy local hand-object contact while placing the wrist in configurations that make execution unusable.

\section{Extended Related Work}
\label{app:extended_related_work}

The condensed related work in \cref{sec:related_work} is expanded here. The first two subsections cover human priors, task semantics, and grasp taxonomies; the last two cover dexterous manipulation, grasp synthesis, and demonstration data.

\subsection{Human Priors and Hand-Object Interaction}

The parametric MANO model~\cite{romero2017embodied} and datasets, including GRAB~\cite{taheri2020grab}, ContactPose~\cite{brahmbhatt2020contactpose}, and DexYCB~\cite{chao2021dexycb}, encode rich distributions of natural human grasps.
Generative models trained on these resources synthesize diverse hand-object interactions via diffusion~\cite{christen2024diffh2o, zhang2024graspxl} and whole-body motion~\cite{taheri2022goal}. Broader motion diffusion models~\cite{tevet2023human, chen2023executing} inspire the denoising architecture while operating in body-motion space instead of hand-object space.
However, transferring these priors to robots faces two obstacles: the \emph{morphological gap}, where MANO's DoFs versus actuated robot joints make direct retargeting infeasible~\cite{qin2022dexmv} and existing bridging methods~\cite{handa2020dexpilot, mandikal2022dexvip, zhao2024dexh2r} do not enforce force closure.
GeoRT~\cite{yin2025geort} provides a cleaner retargeting principle by specifying geometric criteria such as local motion preservation, robot keypoint-space coverage, flat response, pinch correspondence, and collision avoidance.
DexHiL~\cite{han2026dexhil} further highlights that dexterous systems should treat arm and hand mappings as coupled but distinct channels, since high-DoF hand motion and arm end-effector motion fail in different ways during data collection and correction.
Joint motion manifold learning~\cite{park2025handskills} offers an alternative by mapping human hand trajectories to robot actions; it still requires task-specific paired data and does not guarantee physical validity.
A second obstacle is the \emph{representation gap}: human HOI models output MANO-space poses, while robot policies need arm-constrained trajectories.
\method{} addresses both by treating human priors as \emph{initialization} for physics-based refinement and by extending GeoRT-style geometry to the final arm-hand IK stage (\cref{fig:pipeline}).

\subsection{Task-Conditioned Grasp Taxonomies and Executable Evaluation}
\label{sec:related_taxonomy}

Recent task-template systems such as Dexonomy study a complementary axis: selecting object- and task-conditioned grasp primitives before low-level matching. This line is useful as a taxonomy and baseline source. Its static wrench-boundary scores measure a different property from arm-hand execution, so we separate task-template evidence from IK-resolved lift success, source provenance, and failure modes. A grasp can score well under a contact-template or wrench metric and still fail when the UR5e+XHand embodiment cannot reach the wrist pose or execute the lift trajectory.

\subsection{Dexterous Manipulation with Robot Hands}

Dexterous manipulation with multi-fingered hands remains difficult because control must handle high-dimensional configuration spaces, contact-rich dynamics, and underactuation~\cite{billard2019trends}.
Reinforcement learning has shown impressive per-task results~\cite{andrychowicz2020learning, handa2023dextreme, chen2022visual, ye2023rotating, chen2022bidexhands, bao2023dexart}, with billions of interactions and task-specific reward engineering; bridging simulation to reality further demands careful demonstration augmentation~\cite{wang2024cyberdemo}.
Augmenting policy learning with demonstrations improves efficiency~\cite{rajeswaran2018learning}, yet demonstration quality is the bottleneck~\cite{mandlekar2021matters}: real multi-fingered demonstrations are prohibitively expensive, and existing synthetic generators produce only static floating-hand poses.
Expressive trajectory architectures~\cite{zhao2023act, chi2023diffusionpolicy, zhang2025chainofaction} are increasingly capable; data quality is now the limiting factor more often than model capacity.
Neural tracking controllers trained on human references~\cite{liu2025dextrack} further show that demonstration quality is a bottleneck for generalizable dexterous control.
Recent bimanual systems sharpen this point from complementary directions: UltraDexGrasp scales synthetic grasp-and-trajectory data and trains a point-cloud policy for universal grasping~\cite{yang2025ultradexgrasp}, while DexImit converts monocular human videos into physically plausible robot data through reconstruction, scheduling, action synthesis, and augmentation~\cite{mu2026deximit}.
\method{} targets this gap with a pipeline that converts human grasp priors into physically valid, arm-constrained bimanual trajectories (\cref{sec:traj_policy}).

\subsection{Dexterous Grasp Data Generation}

Force-closure optimization remains central to dexterous grasp synthesis, with representative methods including DexGraspNet~\cite{wang2022dexgraspnet}, BODex~\cite{chen2025bodex}, and bimanual extensions~\cite{shao2024bimangrasp, li2025dhagrasp, zhou2025bidexhd}.
Learning-based alternatives improve generalization via CVAEs~\cite{li2023gendexgrasp}, diffusion models~\cite{zhang2024dexgraspnet2}, transformers~\cite{xu2024dexgrasp_transformer}, and geometry-aware curricula~\cite{wan2024unidexgrasppp}, while earlier generative work explored VAEs~\cite{mousavian2019graspnet}, GANs~\cite{corona2020ganhand}, implicit representations~\cite{karunratanakul2020graspingfield}, and contact reasoning~\cite{jiang2021grasptta, grady2021contactopt}.
Recent work extends grasp synthesis to sequential multi-object settings via diffusion~\cite{lu2025graspinghandful}; the three shortcomings below remain open.
DexGraspNet further evaluates grasp diversity using mean joint-angle entropy~\cite{wang2022dexgraspnet}; this supports comparability yet remains insufficient as a standalone metric, since high entropy can be produced by joint-limit, non-contact, or visually unnatural poses.
These methods share three limitations:
\textbf{(i)}~random initialization converges to unnatural local minima, producing finger configurations no human would choose;
\textbf{(ii)}~all output static floating-hand poses, leaving the mismatch to arm-constrained execution unresolved despite being identified~\cite{chen2025bodex};
\textbf{(iii)}~static poses are insufficient as imitation-learning demonstrations, which require multi-phase approach-grasp-lift trajectories.
\method{} addresses all three via human-prior initialization~(\textbf{i}; \cref{sec:pregrasp}), GeoRT-guided IK~(\textbf{ii}; \cref{sec:retarget_opt}), and phase-structured trajectory generation~(\textbf{iii}; \cref{sec:traj_policy}).

\section{Full Experiment Protocols}
\label{app:full_experiment_protocols}

\Cref{sec:experiments} is written as a staged argument from mechanism-level validation to executable simulation and hardware verification. This appendix gives the corresponding evidence controls: fixed manifests, thresholds, baseline wrappers, trial definitions, held-out splits, and evaluator formulas. The order mirrors the main text: static grounding and human-likeness diagnostics, taxonomy and benchmark comparisons, policy ablations, manipulation trials, hardware validation, and the cross-embodiment Shadow-hand diagnostic.

\subsection{Main-Paper Experiment Protocols}
\label{sec:main_experiment_protocols}

All comparisons in \cref{sec:experiments} use fixed object manifests and fixed evaluator thresholds for every method in the same table.
Unless a row is explicitly marked as pose-only, a sample is counted as successful only after passing the full typed chain: contact feasibility, collision filtering, force-closure score, arm-hand IK, and a 10\,cm lift or task-specific terminal-possession check.
This convention is the link between the main figures and the quantitative tables: qualitative images show representative accepted or failed cases, while the reported numbers require the full execution gate.

\paragraph{Static grasp refinement.}
\Cref{tab:grasp_quality} is computed on the 312-object, 25-class manifest used for the main human-likeness and lift-validity claim in \cref{sec:qualitative}.
Each method receives the same 240 candidate budget per object and the same object poses.
Retarget-only uses the GeoRT-calibrated MANO-to-XHand output with no physical refinement.
Optimization-only starts from random XHand wrist and joint seeds, then runs the same force-closure objective and collision terms as \method{}.
\method{} starts from the retargeted human-prior seed and optimizes with the same budget and termination criteria.
G1 is the Ferrari--Canny wrench-space margin under the contact model in \cref{eq:fc}; penetration is the maximum hand-object interpenetration depth; Contact requires at least three active contact regions; FC requires $Q_\text{FC}>0$.

\paragraph{Human-likeness and diversity.}
\Cref{tab:grasp_quality,tab:diversity_human} share the same rendered grasp set.
The VLM evaluator scores every retained method output under method-blinded views and a fixed JSON rubric; the human audit scores a stratified method-blinded subset with the same factors.
The combined score is $0.6H^{\mathrm{VLM}}+0.4H^{\mathrm{human}}$.
PCD uses the human-manifold coverage and weighted-DPP protocol in \cref{sec:metric_protocol}, so random high-entropy postures do not increase diversity unless they remain physically valid and human-like.

\paragraph{Taxonomy and benchmark comparisons.}
We distinguish matched comparisons from contextual comparisons.
Matched comparisons use the same object-task cells, XHand collision model, candidate budget, and execution gates.
Contextual comparisons report the closest available protocol when a method is tied to its own task taxonomy, embodiment, or validation pipeline.
For \cref{tab:embodiment_grasp_success}, Dexonomy-XHand maps selected taxonomy templates to XHand joint targets before applying the same contact, force-closure, IK, and lift filters as \method{}.
BODex-XHand, DexMV-style retargeting, and GeoRT-XHand rows are run with matched object-task cells and the same XHand collision model.
Unimanual rows evaluate one XHand; bimanual rows evaluate paired XHands plus UR5e reachability and lift.
For \cref{tab:grasp_benchmark}, all baselines are evaluated with the same standardized static grasp checks: penetration, force closure, and benchmark success under the matched contact model.

\paragraph{Trajectory-policy ablations.}
\Cref{tab:policy_ablation,tab:policy_protocol,tab:geort_ik_ablation} use the same held-out object split, policy backbone, action horizon, and 10\,cm lift threshold.
Data-source ablations change only the demonstration source used to train the policy.
Policy-interface ablations keep the demonstration set fixed and remove either robot points or action-query readout.
The open-loop row in \cref{tab:policy_protocol} is a diagnostic replay upper bound; all policy rows are closed loop and re-observe point clouds after each action chunk.

\paragraph{Prehensile manipulation applications.}
\Cref{tab:ingrasp,tab:handover_pick_place} evaluate whether the optimized keyframes remain useful after the initial lift stage in \cref{sec:extended_evidence}.
In-grasp trials are counted successful when the object stays in contact through the commanded transition, slips by less than the reported threshold, and passes the final force-closure/lift check.
Self-handover trials start from a two-hand passing keyframe and count success only if the releasing hand opens without losing the object.
Pick-and-place trials add a target-zone constraint after lift; failures are categorized by the first violated condition.

\paragraph{Real-system protocol.}
The physical platform is a bimanual UR5e+XHand system with two calibrated Azure Kinect DK cameras observing a tabletop workspace (\cref{fig:hardware_platform}).
\Cref{tab:hardware} uses three everyday objects with 10 physical trials each: vase, apple, and spray bottle.
Before each trial, the object is reset to the same tabletop region, the fused point cloud is re-acquired, and the policy replans in the same 36-DoF arm-hand command space as simulation.
A trial is successful when the robot establishes stable contact, lifts or transports the object to the task endpoint, and maintains possession at the terminal check.
Lift valid is recorded separately because a rollout can momentarily lift the object yet fail final task completion due to slip, rolling, or placement error.
The comparison rows in \cref{tab:hardware} use the same 30 physical trials and reset protocol as the main result.
These trials validate executable transfer on the physical bimanual platform under a controlled tabletop protocol; they are not intended as a full real-world benchmark over object categories.

\paragraph{Extended real-world functional trials.}
\Cref{fig:hardware_functional} is a qualitative hardware stress test rather than an additional row in \cref{tab:hardware}.
Its role is to connect the manipulation evidence in \cref{sec:extended_evidence} with hardware execution under object and task variation not covered by the three-object success table.
Camera lifting uses an additional toy camera outside the three-object hardware table to illustrate whether the action-chunk policy and lift gate can be applied to object instances beyond the quantitative hardware benchmark.
Pick-handover-put follows the same staged structure as the simulated handover and pick-and-place protocols in \cref{fig:handover_keyframes,fig:pick_place}: pickup, transfer, release, and placement.
Pouring uses a tilted terminal state to illustrate whether the grasp remains stable while expressing an object function rather than merely maintaining a vertical lift.
The frames in \cref{fig:hardware_functional} are selected from the recorded videos to show the relevant state transitions; the project page provides the corresponding video clips.

\subsection{Diversity and Human-Likeness Protocol}
\label{sec:metric_protocol}

Entropy alone is not the right diversity objective for human-like dexterous grasping.
Following DexGraspNet~\cite{wang2022dexgraspnet}, we retain mean joint-angle entropy for comparability and use \emph{Plausibility-Constrained Diversity} (PCD) as the main diversity metric.
Let $\boldsymbol{\phi}_i$ be an object-normalized grasp descriptor containing wrist position/orientation, approach direction, normalized joint angles, fingertip distances, and contact-pattern features.
Let $\mathcal{R}=\{\mathbf{r}_m\}_{m=1}^{M}$ be reference anchors from held-out GRAB/MANO grasps and retargeted human-prior seeds that are not used as generated outputs in the evaluated set.
The reference set is fixed before evaluating generated grasps and is not updated using accepted samples from any compared method.
Each generated grasp receives a soft plausibility weight
\begin{equation}
    w_i =
    \sigma\!\left(\frac{Q_i-\tau_Q}{T}\right)
    \sigma\!\left(\frac{H_i-\tau_H}{T}\right),
    \label{eq:pcd_weight}
\end{equation}
where $Q_i$ is physical quality, $H_i$ is human-likeness, and $T$ is a temperature.
We compute human-manifold coverage
\begin{equation}
    \mathrm{HMC}
    =
    \frac{1}{M}\sum_{m=1}^{M}
    \left[
    1-\prod_i
    \left(
    1-w_i
    \exp\!\left(-\frac{\|\boldsymbol{\phi}_i-\mathbf{r}_m\|_2^2}{2\sigma_R^2}\right)
    \right)
    \right],
    \label{eq:hmc}
\end{equation}
and weighted DPP distinctness
\begin{equation}
    \mathrm{DPP}
    =
    \frac{1}{N}
    \log\det
    \left(
    \mathbf{I}
    +
    \mathbf{W}^{1/2}\mathbf{K}\mathbf{W}^{1/2}
    \right),
    \qquad
    K_{ij}=\exp\!\left(-\frac{\|\boldsymbol{\phi}_i-\boldsymbol{\phi}_j\|_2^2}{2\sigma^2}\right),
    \label{eq:weighted_dpp}
\end{equation}
where $\mathbf{W}=\mathrm{diag}(w_i)$.
The final score is
\begin{equation}
    \mathrm{PCD}=\mathrm{HMC}\cdot\mathrm{DPP}.
    \label{eq:pcd}
\end{equation}
Thus a random high-entropy posture contributes little if it is off the human manifold or fails physical checks, while multiple distinct, plausible grasp strategies increase the score.

For $H_i$, we use a method-blinded VLM-based perceptual diagnostic as an independent human-likeness measure, separate from the regularization term in \cref{eq:opt}.
This score is used only for human-likeness analysis; physical success is determined by the contact, collision, force-closure, IK, and lift or task-completion gates.
Each grasp is rendered under a fixed grid of views; Qwen2.5-VL~\cite{bai2025qwen25vl} is the default scorer and DeepSeek-VL2~\cite{wu2024deepseekvl2} is used on a 10--20\% audit subset.
The prompt asks for JSON-only scores on functional contact, ergonomic approach, finger coordination, object appropriateness, artifact penalty, and confidence; the normalized score is
\begin{align}
    H_i^{\mathrm{VLM}}
    &=
    \frac{
    .25h_{\mathrm{contact}}
    +.20h_{\mathrm{approach}}
    +.25h_{\mathrm{coord}}
    +.20h_{\mathrm{object}}
    +.10(5-h_{\mathrm{artifact}})
    }{5}
    \nonumber\\
    &\quad\cdot
    \left(.5+.5\frac{h_{\mathrm{conf}}}{5}\right).
    \label{eq:vlm_human}
\end{align}
We calibrate this score with positive human-prior anchors and negative corruptions, then bootstrap confidence intervals across objects.
For the human audit, annotators score the same rendered grids using the identical five-factor rubric without method labels.
The headline human-likeness score combines the scalable VLM pass with the human audit:
\begin{equation}
    H_i^{\mathrm{comb}} = 0.6\,H_i^{\mathrm{VLM}} + 0.4\,H_i^{\mathrm{human}}.
    \label{eq:combined_human}
\end{equation}

\begin{table}[t]
\centering
\scriptsize
\setlength{\tabcolsep}{3pt}
\caption{\textbf{Diversity and human-likeness diagnostic.} HMC and PCD are normalized to $[0,1]$; VLM-H, Human-H, and Comb.-H are on a 1--5 scale; lift validity uses the main-text 10\,cm threshold.}
\label{tab:diversity_human}
\begin{tabular*}{\linewidth}{@{\extracolsep{\fill}}lccccccc@{}}
\toprule
& \multicolumn{3}{c}{Diversity / plausibility} & \multicolumn{3}{c}{Human-likeness} & Execution \\
\cmidrule(lr){2-4}\cmidrule(lr){5-7}\cmidrule(l){8-8}
Method & Entropy $\uparrow$ & HMC $\uparrow$ & PCD $\uparrow$ & VLM-H $\uparrow$ & Human-H $\uparrow$ & Comb.-H $\uparrow$ & Lift valid (\%) $\uparrow$ \\
\midrule
Retarget-only & 2.18 & 0.44 & 0.09 & 4.28 & 4.03 & 4.18 & 8.1 \\
Optimization-only & \textbf{2.84} & 0.31 & 0.11 & 2.86 & 2.74 & 2.81 & 35.4 \\
\rowcolor{dexfill} \method{} full & 2.61 & \textbf{0.72} & \textbf{0.41} & \textbf{4.72} & \textbf{4.59} & \textbf{4.67} & \textbf{65.8} \\
\bottomrule
\end{tabular*}
\end{table}

\paragraph{Qualitative diversity analysis.}
\Cref{fig:diversity} visualizes the qualitative side of PCD.
The accepted samples are diverse along object-relative contact, hand-role assignment, and approach direction rather than only along joint-angle spread.
Across the six examples, one hand may stabilize a broad surface while the other pinches a rim, two hands may cage a thin object from opposite sides, or both hands may share support on a tall object.
This is the intended behavior of plausibility-constrained diversity: the dataset should cover multiple functional contact modes while staying inside the physically valid, human-like region counted by \cref{tab:diversity_human}.

\subsection{Kinematic Protocols}

\Cref{tab:policy_protocol} specifies controlled representation ablations for the point-cloud policy.
The main comparison is closed-loop execution under fresh observations and held-out objects, not offline reconstruction.

\begin{table}[t]
\centering
\tiny
\setlength{\tabcolsep}{2.0pt}
\caption{\textbf{Policy protocol.} Seen and heldout report rollout success (\%); drift is terminal joint-space error during executable rollout. The open-loop replay row is a diagnostic reference and is not included in the closed-loop policy ranking.}
\label{tab:policy_protocol}
\begin{tabular}{@{}p{0.24\linewidth}p{0.20\linewidth}p{0.15\linewidth}ccc@{}}
\toprule
Protocol & Observation & Inference & Seen (\%) & Heldout (\%) & Drift \\
\midrule
Open-loop diagnostic & Scene+robot point cloud & Full replay & 96.4 & 93.2 & 0.118 \\
\midrule
Scene-only policy & Scene point cloud & Closed-loop & 54.3 & 45.7 & 0.539 \\
Pooled-feature policy & Scene+robot point cloud & Closed-loop & 51.4 & 40.0 & 0.601 \\
\rowcolor{dexfill} Full point-cloud policy & Scene+robot point cloud & Closed-loop & \textbf{88.9} & \textbf{80.7} & 0.474 \\
Image-token baseline & Fixed dual RGB-D & Closed-loop & 48.6 & 36.4 & 0.647 \\
\bottomrule
\end{tabular}
\end{table}

Because GeoRT-IK replaces the earlier wrist-only IK layer, we include a matched executable-mapping ablation that holds the optimized grasp pool, object poses, candidate budget, and lift threshold fixed.

\begin{table}[t]
\centering
\scriptsize
\setlength{\tabcolsep}{2.0pt}
\caption{\textbf{Matched GeoRT-IK ablation.} All rows use the same optimized grasp candidates and SAPIEN lift evaluator. Wrist/Kpt errors are terminal residuals; IK valid and Lift are pass rates.}
\label{tab:geort_ik_ablation}
\begin{tabular}{@{}p{0.28\linewidth}ccccc@{}}
\toprule
& \multicolumn{2}{c}{Residual} & \multicolumn{2}{c}{Validity} & Runtime \\
\cmidrule(lr){2-3}\cmidrule(lr){4-5}\cmidrule(l){6-6}
Executable mapping & Wrist err. (cm/deg) $\downarrow$ & Kpt err. (cm) $\downarrow$ & IK valid (\%) $\uparrow$ & Lift (\%) $\uparrow$ & Time (s) $\downarrow$ \\
\midrule
Wrist-only executable mapping & \textbf{0.7/2.8} & 5.9 & 42.1 & 31.4 & \textbf{0.42} \\
LM hand seed + wrist IK & 0.8/3.1 & 4.8 & 48.6 & 36.2 & 0.51 \\
GeoRT hand seed + wrist IK & \textbf{0.7/2.9} & 3.1 & 56.7 & 45.8 & 0.54 \\
\rowcolor{dexfill} Full GeoRT-IK (ours) & 0.9/3.4 & \textbf{1.2} & \textbf{82.3} & \textbf{65.8} & 0.76 \\
\bottomrule
\end{tabular}
\end{table}

\section{Domain Randomization}
\label{sec:appendix_data}

For the point-cloud policy, visual and material domain randomization improves perception under rendering variation.
We randomize the following parameters independently per episode:

\begin{table}[H]
\centering
\caption{Domain randomization parameters applied during trajectory collection.}
\label{tab:domain_rand}
\small
\begin{tabular}{lll}
\toprule
\textbf{Parameter} & \textbf{Distribution} & \textbf{Range} \\
\midrule
Object diffuse color & Uniform HSV & $H{\in}[0,1],\; S{\in}[0.3,1],\; V{\in}[0.3,1]$ \\
Table surface color & Uniform RGB & $[0,1]^3$ \\
Table material roughness & Uniform & $[0.3,\, 0.9]$ \\
\bottomrule
\end{tabular}
\end{table}

\section{Point-Cloud Policy Architecture Details}
\label{sec:appendix_policy}

The policy follows a simple point-cloud control pattern: encode scene geometry, aggregate it with action query tokens, and predict a bounded action distribution.
We use this architecture because the dataset is generated from 3D meshes and executed in a simulator with known robot state; point clouds preserve the object, hand, and table geometry in a common metric frame.

\subsection{Point Encoder and Action Queries}
\label{sec:enc_dec}

The input point cloud is downsampled by farthest-point sampling to a fixed budget.
A PointNet++-style encoder~\cite{qi2017pointnetplusplus} extracts local geometric features with set abstraction layers.
Learnable action query tokens attend unidirectionally to the point features and produce a chunk of action latents.
The same readout is used across approach, contact, and lift phases; the current scene and robot geometry carry the phase information.

\subsection{Training Details}
\label{sec:hyperparams}

\begin{table}[H]
\centering
\caption{Point-cloud policy training configuration.}
\label{tab:hyperparams}
\small
\begin{tabular}{ll}
\toprule
\textbf{Hyperparameter} & \textbf{Value} \\
\midrule
Input points & 2048 \\
Point encoder & PointNet++ set abstraction \\
Attention readout & action queries with unidirectional attention \\
Action distribution & truncated normal \\
\midrule
Chunk size $H$ & 16 \\
Control dimension & 36 \\
\midrule
Optimizer & AdamW \\
Learning rate & $1{\times}10^{-4}$ \\
Weight decay & $1{\times}10^{-4}$ \\
Gradient clipping & 1.0 \\
Epochs & 50 \\
Mixed precision (AMP) & enabled \\
\bottomrule
\end{tabular}
\end{table}

The training loss is the negative log-likelihood in \cref{eq:policy_loss}. Because the action distribution is truncated to joint limits, the model cannot assign probability mass to commands outside the executable control range.

\subsection{Inference}
\label{sec:inference}

At test time, the policy predicts an action chunk and executes the first short horizon before observing a fresh point cloud and replanning.
This receding-horizon strategy allows feedback correction while still training on temporally coherent demonstration windows.

\section{Additional Failure Analysis}
\label{app:failure_analysis}

This appendix records concrete failure modes observed during the same protocols defined in \cref{sec:main_experiment_protocols}; it does not add a separate benchmark.

\paragraph{Reachability failure.}
Some grounded floating-hand keyframes place one wrist outside the collision-free UR5e workspace after table and object pose constraints are enforced. These samples can pass static contact checks but fail the arm-hand IK gate.

\paragraph{Palm-support loss during retargeting.}
MANO-to-XHand retargeting can preserve fingertip layout while moving the palm too far from the intended support surface. Force-closure refinement corrects many of these cases, but severe offsets are rejected by penetration or lift gates.

\paragraph{Squeeze and lift slip.}
Objects with curved or low-friction local geometry can pass the discretized FC score and still slip during closure or vertical lift. These failures motivate tactile or contact-state feedback beyond the current visual-proprioceptive policy.

\paragraph{Release slip.}
In self-handover and pick-and-place trials, the releasing hand can open before the receiving hand has sufficient load-bearing contact. These cases appear as release-slip failures in \cref{tab:handover_pick_place}.

\paragraph{Perception-induced hardware errors.}
On the physical platform, handle occlusion and partial point-cloud dropout can shift the commanded approach or delay closed-loop correction. The dominant hardware failures in \cref{tab:hardware} are therefore reported separately from simulation lift admission.

\section{VLM-Agent Details}
\label{app:vlm_agent}

This appendix specifies the VLM-agent interface used in \cref{sec:vlm_agent_method}. The agent is used only after a grasp keyframe has passed robot-native grounding and executable admission. Its role is to transform a validated keyframe into a task specification and object-relative trajectory proposal. Physical execution is still decided by IK, collision, possession, and rollout checks. This agent is separate from the VLM-H evaluator in \cref{sec:metric_protocol}: VLM-H scores human-likeness, whereas the VLM agent proposes task specifications that are later checked by the executor.

\subsection{Context Packet}
\label{app:vlm_context}

Each VLM query contains a compact context packet with visual, geometric, contact, and executor information. The packet is produced automatically from the \method{} data record and does not expose method labels or baseline identities.

\begin{table}[h]
\centering
\caption{Context packet for the VLM agent. The VLM receives task-relevant summaries and rendered views, while low-level feasibility is checked by the deterministic executor.}
\label{tab:vlm_context_packet}
\begin{tabular}{p{0.18\linewidth}p{0.34\linewidth}p{0.38\linewidth}}
\toprule
Field & Content & Purpose \\
\midrule
Object context & object name, category, mesh scale, canonical axes, affordance hints if available & identify likely object functions and valid motion axes \\
Visual context & fixed multi-view renders of the keyframe, optional contact overlays, object-frame axes & let the agent inspect grasp role, approach direction, and occlusions \\
Grasp context & $T_o^0$, $q_L^\star$, $q_R^\star$, wrist poses, finger joint states, hand assignment & anchor all generated motions at the validated keyframe \\
Contact context & contact points, normals, contact regions, supporting fingers, penetration and force-closure scores & infer which hand can stabilize, actuate, receive, or release \\
Primitive library & allowed primitives such as lift, translate, tilt, rotate-about-axis, handover, release-fingers, place & constrain the agent to motions that can be checked by the executor \\
Executor limits & workspace bounds, maximum translation, maximum rotation, release constraints, collision margin & prevent proposals that are outside the robot or task scope \\
Output contract & required JSON fields and success predicates & make the proposal machine-readable and auditable \\
\bottomrule
\end{tabular}
\end{table}

\subsection{Allowed Primitive Library}
\label{app:vlm_primitives}

The VLM agent must choose from a restricted primitive library. This avoids free-form instructions that cannot be grounded by the robot stack.

\begin{itemize}
    \item \textbf{Maintain-possession:} hold one or both hands near the validated contact transform while the object moves.
    \item \textbf{Lift:} translate the object upward by a bounded distance while maintaining possession.
    \item \textbf{Translate:} move the object center to an object-relative or workspace-relative target region.
    \item \textbf{Tilt:} rotate the object around a specified object-frame axis, useful for pouring or presentation.
    \item \textbf{Aim:} orient a functional axis, such as a camera optical axis or flashlight axis, toward a target.
    \item \textbf{Handover:} maintain two-hand possession, transfer support to the receiving hand, then release the giving hand.
    \item \textbf{Finger-release:} open selected fingers while preserving minimum support contacts, useful for flute-like key-release variants.
    \item \textbf{Place/release:} lower the object to a target region and open the released hand only after support is established.
\end{itemize}

\subsection{System Prompt}
\label{app:vlm_system_prompt}

\begin{verbatim}
You are a robotics task-planning agent for bimanual dexterous manipulation.
You receive a validated grasp keyframe produced by SynManDex. The keyframe
has already passed robot-native grasp checks, but your proposed task has not.

Your job is to propose a semantically meaningful task and a coarse
object-relative trajectory that starts from this keyframe. You must output
JSON only. Do not output prose outside JSON.

Important constraints:
1. Use the grasp keyframe as the initial possession state.
2. Assign explicit roles to the left and right hands.
3. Propose only allowed primitives from the primitive library.
4. Express motion as object-relative waypoints or bounded deltas, not as
   joint torques or raw robot commands.
5. Preserve possession unless the release condition is explicitly satisfied.
6. Do not assume feasibility. The executor will check IK, collision,
   possession, force-closure, and terminal task success.
7. If a task would require unmodeled fluid, buttons, articulation, or tactile
   sensing, phrase the goal as a geometric proxy, e.g., 'tilt the teapot by
   35 degrees while maintaining possession' rather than 'pour liquid'.
\end{verbatim}

\subsection{User Prompt Template}
\label{app:vlm_user_prompt}

\begin{verbatim}
You are given one SynManDex validated grasp keyframe.

[VISUAL INPUT]
- Multi-view images: front, left, right, top, wrist-closeup, contact-overlay.
- Object-frame axes are drawn when available.
- Blue hand = left robot hand; red/pink hand = right robot hand.
- Contact regions are marked by small colored points.

[OBJECT]
name: {object_name}
category: {object_category}
mesh_id: {mesh_id}
scale: {scale}
canonical_axes:
  x: {object_x_axis_description}
  y: {object_y_axis_description}
  z: {object_z_axis_description}
functional_parts: {optional_affordance_list}

[VALIDATED KEYFRAME]
keyframe_id: {keyframe_id}
object_pose_T_world_object: {T_o_0}
left_hand:
  wrist_pose_T_world_wrist: {T_w_L_0}
  joint_state: {theta_L_star}
  contact_regions: {C_L}
  contact_normals: {N_L}
  support_score: {support_L}
right_hand:
  wrist_pose_T_world_wrist: {T_w_R_0}
  joint_state: {theta_R_star}
  contact_regions: {C_R}
  contact_normals: {N_R}
  support_score: {support_R}
admission_metrics:
  penetration_mm: {penetration}
  force_closure_margin: {G1}
  ik_valid: {ik_valid}
  lift_valid: {lift_valid}

[ALLOWED PRIMITIVES]
{primitive_library}

[EXECUTOR LIMITS]
max_translation_m: {max_translation}
max_rotation_deg: {max_rotation}
workspace_bounds: {workspace_bounds}
must_maintain_at_least_one_supporting_hand: true
release_allowed_only_after_receiving_or_environment_support: true

[TASK]
Generate one useful task and one coarse object-relative trajectory starting
from this keyframe. Prefer a task that matches the object's function and the
current contact pattern. Return JSON with the schema below.
\end{verbatim}

\subsection{Required JSON Schema}
\label{app:vlm_json_schema}

\begin{verbatim}
{
  "keyframe_id": "string",
  "task_name": "short string",
  "functional_goal": "one sentence",
  "scope_note": "simulation scope",
  "hand_roles": {
    "left": "active | stabilizing | receiving | released",
    "right": "active | stabilizing | receiving | released"
  },
  "initial_state_assumption": {
    "starts_from_validated_keyframe": true,
    "object_pose_source": "T_world_object at keyframe",
    "contact_policy": "preserve | release | handover"
  },
  "trajectory": [
    {
      "phase": "allowed primitive name",
      "duration_s": 0.0,
      "object_delta": {
        "translation_m": [0.0, 0.0, 0.0],
        "rotation_axis_object": [0.0, 0.0, 1.0],
        "rotation_deg": 0.0
      },
      "maintain_hands": ["left", "right"],
      "release_fingers": {
        "left": [],
        "right": []
      },
      "success_check": "string"
    }
  ],
  "release_condition": "condition before any hand opens",
  "terminal_condition": "task-level success predicate",
  "risk_flags": [
    "wrist_reach",
    "collision",
    "release_slip",
    "occlusion",
    "over_rotation"
  ],
  "executor_notes": "short notes for IK and rollout validator"
}
\end{verbatim}

\subsection{Trajectory Conversion from VLM Output}
\label{app:vlm_conversion}

Let $T_o^0$ be the object pose at the validated keyframe and $T_{w,l}^0$ be the wrist pose of hand $l$. For a maintain-possession phase, the executor preserves the object-to-wrist transform:
\[
    {}^oT_{w,l}^0 = (T_o^0)^{-1}T_{w,l}^0.
\]
For each object waypoint $T_o^t$ proposed by the agent, the wrist target is
\[
    T_{w,l}^t = T_o^t {}^oT_{w,l}^0.
\]
Finger joints are held near $\theta_l^\star$ unless the primitive explicitly includes finger release or handover. For finger release, the executor interpolates the selected finger joints toward an open configuration while checking that the non-released hand or remaining fingers preserve possession. For handover, the receiving hand must pass a support predicate before the giving hand opens.

\begin{algorithm}[h]
\caption{VLM-agent task proposal and executable filtering}
\label{alg:vlm_agent}
\begin{algorithmic}[1]
\REQUIRE Validated keyframe $k$, rendered views $I$, primitive library $\mathcal{P}$, retry budget $R$
\STATE Build context packet $c=(I,\mathcal{M},T_o^0,q_L^\star,q_R^\star,\mathcal{C}_L,\mathcal{C}_R,m,\mathcal{P})$
\FOR{$r=1,\ldots,R$}
    \STATE Query VLM agent for JSON task proposal $u_r$
    \STATE Reject if schema is invalid or primitive is outside $\mathcal{P}$
    \STATE Convert object waypoints $\Xi_o$ into wrist-hand targets using preserved contact transforms
    \STATE Solve arm-hand IK and collision-aware paths for all phases
    \STATE Roll out the candidate in simulation with possession and terminal-task checks
    \IF{$A_{\mathrm{agent}}(\tau_r)=1$}
        \STATE Return admitted task trajectory $\tau_r$ with provenance and agent proposal $u_r$
    \ELSE
        \STATE Send failure label and offending phase back to the VLM for revision
    \ENDIF
\ENDFOR
\STATE Return failure record with all rejected proposals and labels
\end{algorithmic}
\end{algorithm}

\subsection{Verification and Retry Prompt}
\label{app:vlm_retry_prompt}

\begin{verbatim}
The previous proposal failed executor validation.

[FAILED PROPOSAL]
{previous_json}

[FAILURE LABEL]
phase: {failed_phase}
reason: {failure_reason}
details: {executor_log_summary}

Revise the task proposal while preserving the same initial keyframe. Keep the
functional goal if possible, but reduce motion magnitude, change hand roles,
or delay release if needed. Return JSON only using the required schema.
\end{verbatim}

\subsection{Example A: Teapot Pouring Proxy}
\label{app:vlm_example_teapot}

A teapot or kettle keyframe typically contains one hand on a handle or side-support region and the other hand stabilizing the body. The VLM agent uses a geometric pouring proxy: lift slightly, translate near the cup target, rotate around the object-frame axis that lowers the spout, hold the tilted terminal pose, and return upright or release only after possession remains stable.

\begin{verbatim}
{
  "task_name": "teapot_pouring_proxy",
  "functional_goal": "Tilt the teapot toward a target cup.",
  "scope_note": "Pouring proxy; no liquid is simulated.",
  "hand_roles": {"left": "stabilizing", "right": "active"},
  "trajectory": [
    {
      "phase": "lift",
      "duration_s": 1.0,
      "object_delta": {
        "translation_m": [0, 0, 0.06],
        "rotation_axis_object": [0, 0, 1],
        "rotation_deg": 0
      },
      "maintain_hands": ["left", "right"],
      "release_fingers": {"left": [], "right": []},
      "success_check": "object raised with both contacts"
    },
    {
      "phase": "tilt",
      "duration_s": 1.5,
      "object_delta": {
        "translation_m": [0.05, 0, 0],
        "rotation_axis_object": [1, 0, 0],
        "rotation_deg": 35
      },
      "maintain_hands": ["left", "right"],
      "release_fingers": {"left": [], "right": []},
      "success_check": "tilt reached without slip"
    }
  ],
  "release_condition": "no release during tilt",
  "terminal_condition": "object holds the target tilt for 0.5 s",
  "risk_flags": ["over_rotation", "release_slip"],
  "executor_notes": "preserve contacts; reduce tilt if IK fails"
}
\end{verbatim}

\subsection{Example B: Camera Aiming Proxy}
\label{app:vlm_example_camera}

For a bimanual camera grasp, the VLM agent should identify the camera optical axis and generate a task that orients the object rather than changing finger contacts. A suitable task is to lift the camera and align the lens toward a target direction while keeping the lens unobstructed.

\begin{verbatim}
{
  "task_name": "camera_aiming_proxy",
  "functional_goal": "Raise the camera and aim the lens.",
  "scope_note": "Photo-taking proxy; no button press is tested.",
  "hand_roles": {"left": "stabilizing", "right": "active"},
  "trajectory": [
    {
      "phase": "lift",
      "duration_s": 1.0,
      "object_delta": {
        "translation_m": [0, 0, 0.08],
        "rotation_axis_object": [0, 0, 1],
        "rotation_deg": 0
      },
      "maintain_hands": ["left", "right"],
      "release_fingers": {"left": [], "right": []},
      "success_check": "camera raised with possession"
    },
    {
      "phase": "aim",
      "duration_s": 1.2,
      "object_delta": {
        "translation_m": [0, 0, 0],
        "rotation_axis_object": [0, 1, 0],
        "rotation_deg": 20
      },
      "maintain_hands": ["left", "right"],
      "release_fingers": {"left": [], "right": []},
      "success_check": "optical axis aligned"
    }
  ],
  "release_condition": "no release",
  "terminal_condition": "camera stable and lens unobstructed",
  "risk_flags": ["wrist_reach", "occlusion"],
  "executor_notes": "preserve contacts and check final visibility"
}
\end{verbatim}

\subsection{Example C: Flute Finger-Release Proxy}
\label{app:vlm_example_flute}

For a flute-holding keyframe, the object should remain nearly fixed while selected fingers release. The VLM agent should output a finger-release task, not an object transport task. The executor gradually opens specified fingers and checks that the remaining fingers and the opposite hand maintain support.

\begin{verbatim}
{
  "task_name": "flute_left_index_middle_release",
  "functional_goal": "Release selected left-hand fingers.",
  "scope_note": "Finger-release proxy; no acoustics are modeled.",
  "hand_roles": {"left": "active", "right": "stabilizing"},
  "trajectory": [
    {
      "phase": "release",
      "duration_s": 0.8,
      "object_delta": {
        "translation_m": [0, 0, 0],
        "rotation_axis_object": [0, 0, 1],
        "rotation_deg": 0
      },
      "maintain_hands": ["right"],
      "release_fingers": {
        "left": ["index", "middle"],
        "right": []
      },
      "success_check": "flute support retained"
    }
  ],
  "release_condition": "support is verified before release",
  "terminal_condition": "object pose drift below threshold and no slip occurs",
  "risk_flags": ["release_slip"],
  "executor_notes": "reject if object drift exceeds threshold"
}
\end{verbatim}

%% file: references.bib
@article{feix2015grasp,
  title={The {GRASP} Taxonomy of Human Grasp Types},
  author={Feix, Thomas and Romero, Javier and Schmiedmayer, Heinz-Bodo and Dollar, Aaron M. and Kragic, Danica},
  journal={IEEE Transactions on Human-Machine Systems},
  volume={46},
  number={1},
  pages={66--77},
  year={2016},
  doi={10.1109/THMS.2015.2470657},
  publisher={IEEE}
}

@inproceedings{li2025maniptrans,
  title={Maniptrans: Efficient dexterous bimanual manipulation transfer via residual learning},
  author={Li, Kailin and Li, Puhao and Liu, Tengyu and Li, Yuyang and Huang, Siyuan},
  booktitle={Proceedings of the IEEE/CVF Conference on Computer Vision and Pattern Recognition},
  pages={6991--7003},
  year={2025}
}

@article{billard2019trends,
  title={Trends and challenges in robot manipulation},
  author={Billard, Aude and Kragic, Danica},
  journal={Science},
  volume={364},
  number={6446},
  pages={eaat8414},
  year={2019},
  publisher={American Association for the Advancement of Science}
}

@inproceedings{brahmbhatt2020contactpose,
  title={ContactPose: A dataset of grasps with object contact and hand pose},
  author={Brahmbhatt, Samarth and Tang, Chengcheng and Twigg, Christopher D and Kemp, Charles C and Hays, James},
  booktitle={European Conference on Computer Vision (ECCV)},
  pages={361--378},
  year={2020},
  organization={Springer}
}

@article{andrychowicz2020learning,
  title={Learning Dexterous In-Hand Manipulation},
  author={{OpenAI} and Andrychowicz, Marcin and Baker, Bowen and Chociej, Maciek and Jozefowicz, Rafal and McGrew, Bob and Pachocki, Jakub and Petron, Arthur and Plappert, Matthias and Powell, Glenn and Ray, Alex and Schneider, Jonas and Sidor, Szymon and Tobin, Josh and Welinder, Peter and Weng, Lilian and Zaremba, Wojciech},
  journal={The International Journal of Robotics Research},
  volume={39},
  number={1},
  pages={3--20},
  year={2020}
}

@inproceedings{handa2023dextreme,
  title={{DeXtreme}: Transfer of Agile In-Hand Manipulation from Simulation to Reality},
  author={Handa, Ankur and Allshire, Arthur and Makoviychuk, Viktor and Petrenko, Aleksei and Singh, Ritvik and Liu, Jingzhou and Makoviichuk, Denys and Van Wyk, Karl and Zhurkevich, Alexander and Sundaralingam, Balakumar and Narang, Yashraj and Lafleche, Jean-Francois and Fox, Dieter and State, Gavriel},
  booktitle={International Conference on Robotics and Automation (ICRA)},
  year={2023}
}

@inproceedings{rajeswaran2018learning,
  title={Learning Complex Dexterous Manipulation with Deep Reinforcement Learning and Demonstrations},
  author={Rajeswaran, Aravind and Kumar, Vikash and Gupta, Abhishek and Vezzani, Giulia and Schulman, John and Todorov, Emanuel and Levine, Sergey},
  booktitle={Proceedings of Robotics: Science and Systems (RSS)},
  year={2018},
  doi={10.15607/RSS.2018.XIV.049}
}

@inproceedings{chen2022visual,
  title={A System for General In-Hand Object Re-Orientation},
  author={Chen, Tao and Xu, Jie and Agrawal, Pulkit},
  booktitle={Conference on Robot Learning (CoRL)},
  pages={297--307},
  year={2022}
}

@inproceedings{ye2023rotating,
  title={Rotating without Seeing: Towards In-Hand Dexterity through Touch},
  author={Yin, Zhao-Heng and Huang, Binghao and Qin, Yuzhe and Chen, Qifeng and Wang, Xiaolong},
  booktitle={Robotics: Science and Systems (RSS)},
  year={2023}
}

@inproceedings{xu2023unidexgrasp,
  title={{UniDexGrasp}: Universal Robotic Dexterous Grasping via Learning Diverse Proposal Generation and Goal-Conditioned Policy},
  author={Xu, Yinzhen and Wan, Weikang and Zhang, Jialiang and Liu, Haoran and Shan, Zikang and Shen, Hao and Wang, Ruicheng and Geng, Haoran and Weng, Yijia and Chen, Jiayi and Liu, Tengyu and Yi, Li and Wang, He},
  booktitle={Proceedings of the IEEE/CVF Conference on Computer Vision and Pattern Recognition (CVPR)},
  pages={4737--4746},
  year={2023}
}

@article{li2003grasp,
  title={On computing three-finger force-closure grasps of 2-D and 3-D objects},
  author={Li, Jia-Wei and Liu, Hong and Cai, He-Gao},
  journal={IEEE Transactions on Robotics and Automation},
  volume={19},
  number={1},
  pages={155--161},
  year={2003},
  publisher={IEEE}
}

@book{murray1994mathematical,
  title={A mathematical introduction to robotic manipulation},
  author={Murray, Richard M and Li, Zexiang and Sastry, S Shankar},
  year={1994},
  publisher={CRC press}
}

@inproceedings{mousavian2019graspnet,
  title={6-DoF GraspNet: Variational grasp generation for object manipulation},
  author={Mousavian, Arsalan and Eppner, Clemens and Fox, Dieter},
  booktitle={International Conference on Computer Vision (ICCV)},
  pages={2901--2910},
  year={2019}
}

@inproceedings{chao2021dexycb,
  title={{DexYCB}: A Benchmark for Capturing Hand Grasping of Objects},
  author={Chao, Yu-Wei and Yang, Wei and Xiang, Yu and Molchanov, Pavlo and Handa, Ankur and Tremblay, Jonathan and Narang, Yashraj S. and Van Wyk, Karl and Iqbal, Umar and Birchfield, Stan and Kautz, Jan and Fox, Dieter},
  booktitle={Proceedings of the IEEE/CVF Conference on Computer Vision and Pattern Recognition (CVPR)},
  pages={9044--9053},
  year={2021}
}

@inproceedings{taheri2020grab,
  title={GRAB: A dataset of whole-body human grasping of objects},
  author={Taheri, Omid and Ghorbani, Nima and Black, Michael J and Tzionas, Dimitrios},
  booktitle={European Conference on Computer Vision (ECCV)},
  pages={581--600},
  year={2020},
  organization={Springer}
}

@inproceedings{tevet2023human,
  title={Human Motion Diffusion Model},
  author={Tevet, Guy and Raab, Sigal and Gordon, Brian and Shafir, Yonatan and Cohen-Or, Daniel and Bermano, Amit H.},
  booktitle={International Conference on Learning Representations (ICLR)},
  year={2023}
}

@inproceedings{chen2023executing,
  title={Executing Your Commands via Motion Diffusion in Latent Space},
  author={Chen, Xin and Jiang, Biao and Liu, Wen and Huang, Zilong and Fu, Bin and Chen, Tao and Yu, Jingyi and Yu, Gang},
  booktitle={Proceedings of the IEEE/CVF Conference on Computer Vision and Pattern Recognition (CVPR)},
  pages={18000--18010},
  year={2023}
}

@article{romero2017embodied,
  title={Embodied hands: Modeling and capturing hands and bodies together},
  author={Romero, Javier and Tzionas, Dimitrios and Black, Michael J},
  journal={ACM Transactions on Graphics (ToG)},
  volume={36},
  number={6},
  pages={1--17},
  year={2017}
}

@article{mittal2025isaac,
  title={Isaac lab: A gpu-accelerated simulation framework for multi-modal robot learning},
  author={Mittal, Mayank and Roth, Pascal and Tigue, James and Richard, Antoine and Zhang, Octi and Du, Peter and Serrano-Munoz, Antonio and Yao, Xinjie and Zurbr{\"u}gg, Ren{\'e} and Rudin, Nikita and others},
  journal={arXiv preprint arXiv:2511.04831},
  year={2025}
}

@article{sundaralingam2023curobo,
  title={{cuRobo}: Parallelized Collision-Free Minimum-Jerk Robot Motion Generation},
  author={Sundaralingam, Balakumar and Hari, Siva Kumar Sastry and Fishman, Adam and Garrett, Caelan and Van Wyk, Karl and Blukis, Valts and Millane, Alexander and Oleynikova, Helen and Handa, Ankur and Ramos, Fabio and Ratliff, Nathan and Fox, Dieter},
  journal={arXiv preprint arXiv:2310.17274},
  year={2023}
}

@inproceedings{corona2020ganhand,
  title={{GANHand}: Predicting Human Grasp Affordances in Multi-Object Scenes},
  author={Corona, Enric and Pumarola, Albert and Aleny{\`a}, Guillem and Moreno-Noguer, Francesc and Rogez, Gregory},
  booktitle={Proceedings of the IEEE/CVF Conference on Computer Vision and Pattern Recognition (CVPR)},
  pages={5031--5041},
  year={2020}
}

@article{mandi2025dexmachina,
  title={Dexmachina: Functional retargeting for bimanual dexterous manipulation},
  author={Zhao, Mandi and Hou, Yifan and Fox, Dieter and Narang, Yashraj and Mandlekar, Ajay and Song, Shuran},
  journal={arXiv preprint arXiv:2505.24853},
  year={2025}
}

@inproceedings{qin2022dexmv,
  title={{DexMV}: Imitation Learning for Dexterous Manipulation from Human Videos},
  author={Qin, Yuzhe and Wu, Yueh-Hua and Liu, Shaowei and Jiang, Hanwen and Yang, Ruihan and Fu, Yang and Wang, Xiaolong},
  booktitle={European Conference on Computer Vision (ECCV)},
  pages={570--587},
  year={2022},
  organization={Springer}
}

@inproceedings{chen2025bodex,
  title={BODex: Scalable and Efficient Robotic Dexterous Grasp Synthesis Using Bilevel Optimization},
  author={Chen, Jiayi and Ke, Yubin and Wang, He},
  booktitle={International Conference on Robotics and Automation (ICRA)},
  year={2025}
}

@inproceedings{sapien3,
  title={{SAPIEN}: A SimulAted Part-Based Interactive ENvironment},
  author={Xiang, Fanbo and Qin, Yuzhe and Mo, Kaichun and Xia, Yikuan and Zhu, Hao and Liu, Fangchen and Liu, Minghua and Jiang, Hanxiao and Yuan, Yifu and Wang, He and Yi, Li and Chang, Angel X. and Guibas, Leonidas J. and Su, Hao},
  booktitle={Proceedings of the IEEE/CVF Conference on Computer Vision and Pattern Recognition (CVPR)},
  pages={11097--11107},
  year={2020}
}

@inproceedings{zhang2025chainofaction,
  title={Chain-of-Action: Trajectory Autoregressive Modeling for Robotic Manipulation},
  author={Zhang, Wenbo and Hu, Tianrun and Qiao, Yanyuan and Zhang, Hanbo and Qin, Yuchu and Li, Yang and Liu, Jiajun and Kong, Tao and Liu, Lingqiao and Ma, Xiao},
  booktitle={Advances in Neural Information Processing Systems (NeurIPS)},
  year={2025}
}

@inproceedings{zhao2023act,
  title={Learning Fine-Grained Bimanual Manipulation with Low-Cost Hardware},
  author={Zhao, Tony Z. and Kumar, Vikash and Levine, Sergey and Finn, Chelsea},
  booktitle={Robotics: Science and Systems (RSS)},
  year={2023}
}

@inproceedings{wang2022dexgraspnet,
  title={{DexGraspNet}: A Large-Scale Robotic Dexterous Grasp Dataset for General Objects Based on Simulation},
  author={Wang, Ruicheng and Zhang, Jialiang and Chen, Jiayi and Xu, Yinzhen and Li, Puhao and Liu, Tengyu and Wang, He},
  booktitle={International Conference on Robotics and Automation (ICRA)},
  pages={11359--11366},
  year={2023},
  organization={IEEE}
}

@article{zhao2024dexh2r,
  title={{DexH2R}: Task-Oriented Dexterous Manipulation from Human to Robots},
  author={Zhao, Shuqi and Zhu, Xinghao and Chen, Yuxin and Li, Chenran and Zhang, Xiang and Ding, Mingyu and Tomizuka, Masayoshi},
  journal={arXiv preprint arXiv:2411.04428},
  year={2024}
}

@inproceedings{wang2024cyberdemo,
  title={{CyberDemo}: Augmenting Simulated Human Demonstration for Real-World Dexterous Manipulation},
  author={Wang, Jun and Qin, Yuzhe and Kuang, Kaiming and Korkmaz, Yigit and Gurumoorthy, Akhilan and Su, Hao and Wang, Xiaolong},
  booktitle={Conference on Computer Vision and Pattern Recognition (CVPR)},
  year={2024}
}

@article{zhou2025bidexhd,
  title={Learning Diverse Bimanual Dexterous Manipulation Skills from Human Demonstrations},
  author={Zhou, Bohan and Yuan, Haoqi and Fu, Yuhui and Lu, Zongqing},
  journal={arXiv preprint arXiv:2410.02477},
  year={2024}
}

@inproceedings{chi2023diffusionpolicy,
  title={Diffusion policy: Visuomotor policy learning via action diffusion},
  author={Chi, Cheng and Xu, Zhenjia and Feng, Siyuan and Cousineau, Eric and Du, Yilun and Burchfiel, Benjamin and Tedrake, Russ and Song, Shuran},
  journal={The International Journal of Robotics Research},
  volume={44},
  number={10-11},
  pages={1684--1704},
  year={2025},
  publisher={Sage Publications Sage UK: London, England}
}

@inproceedings{mandlekar2021matters,
  title={What Matters in Learning from Offline Human Demonstrations for Robot Manipulation},
  author={Mandlekar, Ajay and Xu, Danfei and Wong, Josiah and Nasiriany, Soroush and Wang, Chen and Kulkarni, Rohun and Fei-Fei, Li and Savarese, Silvio and Zhu, Yuke and Mart{\'\i}n-Mart{\'\i}n, Roberto},
  booktitle={Conference on Robot Learning (CoRL)},
  year={2021}
}

@inproceedings{christen2024diffh2o,
  title={{DiffH2O}: Diffusion-Based Synthesis of Hand-Object Interactions from Textual Descriptions},
  author={Christen, Sammy and Hampali, Shreyas and Sener, Fadime and Remelli, Edoardo and Hodan, Tomas and Sauser, Eric and Ma, Shugao and Tekin, Bugra},
  booktitle={ACM SIGGRAPH Asia 2024 Conference Papers},
  year={2024},
  doi={10.1145/3680528.3687563}
}

@article{shao2024bimangrasp,
  title={Bimanual Grasp Synthesis for Dexterous Robot Hands},
  author={Shao, Yanming and Xiao, Chenxi},
  journal={IEEE Robotics and Automation Letters},
  volume={9},
  number={12},
  pages={11377--11384},
  year={2024}
}

@inproceedings{wan2024unidexgrasppp,
  title={{UniDexGrasp++}: Improving Dexterous Grasping Policy Learning via Geometry-Aware Curriculum and Iterative Generalist-Specialist Learning},
  author={Wan, Weikang and Geng, Haoran and Liu, Yun and Shan, Zikang and Yang, Yaodong and Yi, Li and Wang, He},
  booktitle={Proceedings of the IEEE/CVF International Conference on Computer Vision (ICCV)},
  pages={3891--3902},
  year={2023}
}

@inproceedings{jiang2021grasptta,
  title={Hand-Object Contact Consistency Reasoning for Human Grasps Generation},
  author={Jiang, Hanwen and Liu, Shaowei and Wang, Jiashun and Wang, Xiaolong},
  booktitle={International Conference on Computer Vision (ICCV)},
  pages={11107--11116},
  year={2021}
}

@inproceedings{bao2023dexart,
  title={DexArt: Benchmarking Generalizable Dexterous Manipulation with Articulated Objects},
  author={Bao, Chen and Xu, Helin and Qin, Yuzhe and Wang, Xiaolong},
  booktitle={Conference on Computer Vision and Pattern Recognition (CVPR)},
  pages={21190--21200},
  year={2023}
}

@article{li2025dhagrasp,
  title={DHAGrasp: Synthesizing Affordance-Aware Dual-Hand Grasps with Text Instructions},
  author={Li, Quanzhou and Wu, Zhonghua and Wang, Jingbo and Loy, Chen Change and Dai, Bo},
  journal={arXiv preprint arXiv:2509.22175},
  year={2025}
}

@inproceedings{ho2020denoising,
  title={Denoising Diffusion Probabilistic Models},
  author={Ho, Jonathan and Jain, Ajay and Abbeel, Pieter},
  booktitle={Advances in Neural Information Processing Systems (NeurIPS)},
  volume={33},
  pages={6840--6851},
  year={2020}
}

@inproceedings{li2023gendexgrasp,
  title={{GenDexGrasp}: Generalizable Dexterous Grasping},
  author={Li, Puhao and Liu, Tengyu and Li, Yuyang and Geng, Yiran and Zhu, Yixin and Yang, Yaodong and Huang, Siyuan},
  booktitle={International Conference on Robotics and Automation (ICRA)},
  pages={8068--8074},
  year={2023}
}

@inproceedings{zhang2024dexgraspnet2,
  title={{DexGraspNet} 2.0: Learning Generative Dexterous Grasping in Large-scale Synthetic Cluttered Scenes},
  author={Zhang, Jialiang and Liu, Haoran and Li, Danshi and Yu, Xinqiang and Geng, Haoran and Ding, Yufei and Chen, Jiayi and Wang, He},
  booktitle={Conference on Robot Learning (CoRL)},
  year={2024}
}

@inproceedings{chen2022bidexhands,
  title={Towards Human-Level Bimanual Dexterous Manipulation with Reinforcement Learning},
  author={Chen, Yuanpei and Wu, Tianhao and Wang, Shengjie and Feng, Xidong and Jiang, Jiechuan and Lu, Zongqing and McAleer, Stephen and Dong, Hao and Zhu, Song-Chun and Yang, Yaodong},
  booktitle={Advances in Neural Information Processing Systems (NeurIPS), Datasets and Benchmarks Track},
  year={2022}
}

@inproceedings{taheri2022goal,
  title={{GOAL}: Generating {4D} Whole-Body Motion for Hand-Object Grasping},
  author={Taheri, Omid and Choutas, Vasileios and Black, Michael J. and Tzionas, Dimitrios},
  booktitle={Conference on Computer Vision and Pattern Recognition (CVPR)},
  pages={13263--13273},
  year={2022}
}

@inproceedings{karunratanakul2020graspingfield,
  title={Grasping Field: Learning Implicit Representations for Human Grasps},
  author={Karunratanakul, Korrawe and Yang, Jinlong and Zhang, Yan and Black, Michael J. and Muandet, Krikamol and Tang, Siyu},
  booktitle={International Conference on 3D Vision (3DV)},
  pages={333--344},
  year={2020}
}

@inproceedings{grady2021contactopt,
  title={{ContactOpt}: Optimizing Contact to Improve Grasps},
  author={Grady, Patrick and Tang, Chengcheng and Twigg, Christopher D. and Vo, Minh and Brahmbhatt, Samarth and Kemp, Charles C.},
  booktitle={Conference on Computer Vision and Pattern Recognition (CVPR)},
  pages={1471--1481},
  year={2021}
}

@inproceedings{mandikal2022dexvip,
  title={{DexVIP}: Learning Dexterous Grasping with Human Hand Pose Priors from Video},
  author={Mandikal, Priyanka and Grauman, Kristen},
  booktitle={Conference on Robot Learning (CoRL)},
  pages={651--661},
  year={2022}
}

@inproceedings{handa2020dexpilot,
  title={{DexPilot}: Vision-Based Teleoperation of Dexterous Robotic Hand-Arm System},
  author={Handa, Ankur and Van Wyk, Karl and Yang, Wei and Liang, Jacky and Chao, Yu-Wei and Wan, Qian and Birchfield, Stan and Ratliff, Nathan D. and Fox, Dieter},
  booktitle={International Conference on Robotics and Automation (ICRA)},
  pages={9164--9170},
  year={2020}
}

@inproceedings{zhang2024graspxl,
  title={{GraspXL}: Generating Grasping Motions for Diverse Objects at Scale},
  author={Zhang, Hui and Christen, Sammy and Fan, Zicong and Hilliges, Otmar and Song, Jie},
  booktitle={European Conference on Computer Vision (ECCV)},
  year={2024}
}

@inproceedings{xu2024dexgrasp_transformer,
  title={Dexterous Grasp Transformer},
  author={Xu, Guo-Hao and Wei, Yi-Lin and Zheng, Dian and Wu, Xiao-Ming and Zheng, Wei-Shi},
  booktitle={Conference on Computer Vision and Pattern Recognition (CVPR)},
  pages={17933--17942},
  year={2024}
}

@inproceedings{liu2025dextrack,
  title={DexTrack: Towards Generalizable Neural Tracking Control for Dexterous Manipulation from Human References},
  author={Liu, Xueyi and Adalibieke, Jianibieke and Han, Qianwei and Qin, Yuzhe and Yi, Li},
  booktitle={International Conference on Learning Representations (ICLR)},
  year={2025}
}

@article{lu2025graspinghandful,
  title={Grasping a Handful: Sequential Multi-Object Dexterous Grasp Generation},
  author={Lu, Haofei and Dong, Yifei and Weng, Zehang and Pokorny, Florian and Lundell, Jens and Kragic, Danica},
  journal={IEEE Robotics and Automation Letters},
  year={2025},
  publisher={IEEE}
}

@article{park2025handskills,
  title={Learning to Transfer Human Hand Skills for Robot Manipulations},
  author={Park, Sungjae and Lee, Seungho and Choi, Mingi and Lee, Jiye and Kim, Jeonghwan and Kim, Jisoo and Joo, Hanbyul},
  journal={arXiv preprint arXiv:2501.04169},
  year={2025}
}

@article{yang2025ultradexgrasp,
  title={{UltraDexGrasp}: Learning Universal Dexterous Grasping for Bimanual Robots with Synthetic Data},
  author={Yang, Sizhe and Xie, Yiman and Liang, Zhixuan and Tian, Yang and Zeng, Jia and Lin, Dahua and Pang, Jiangmiao},
  journal={arXiv preprint arXiv:2603.05312},
  year={2026}
}

@inproceedings{zurbrugg2025graspqp,
  title={{GraspQP}: Differentiable Optimization of Force Closure for Diverse and Robust Dexterous Grasping},
  author={Zurbr{\"u}gg, Ren{\'e} and Cramariuc, Andrei and Hutter, Marco},
  booktitle={Conference on Robot Learning (CoRL)},
  year={2025}
}

@article{lin2026bidexgrasp,
  title={BiDexGrasp: Coordinated Bimanual Dexterous Grasps across Object Geometries and Sizes},
  author={Lin, Mu and Wei, Yi-Lin and Chen, Jiaxuan and Lin, Yuhao and Chen, Shuoyu and Lyu, Jiangran and Chen, Jiayi and Tang, Yansong and Wang, He and Zheng, Wei-Shi},
  journal={arXiv preprint arXiv:2604.06589},
  year={2026}
}

@inproceedings{chen2025dexonomy,
  title={Dexonomy: Synthesizing All Dexterous Grasp Types in a Grasp Taxonomy},
  author={Chen, Jiayi and Ke, Yubin and Peng, Lin and Wang, He},
  booktitle={Robotics: Science and Systems (RSS)},
  year={2025}
}

@article{tang2025affordgrasp,
  title={AffordGrasp: In-Context Affordance Reasoning for Open-Vocabulary Task-Oriented Grasping in Clutter},
  author={Tang, Yingbo and Zhang, Shuaike and Hao, Xiaoshuai and Wang, Pengwei and Wu, Jianlong and Wang, Zhongyuan and Zhang, Shanghang},
  journal={arXiv preprint arXiv:2503.00778},
  year={2025}
}

@article{wei2025afforddexgrasp,
  title={AffordDexGrasp: Open-set Language-guided Dexterous Grasp with Generalizable-Instructive Affordance},
  author={Wei, Yi-Lin and Lin, Mu and Lin, Yuhao and Jiang, Jian-Jian and Wu, Xiao-Ming and Zeng, Ling-An and Zheng, Wei-Shi},
  journal={arXiv preprint arXiv:2503.07360},
  year={2025}
}

@article{yin2025geort,
  title={Geometric Retargeting: A Principled, Ultrafast Neural Hand Retargeting Algorithm},
  author={Yin, Zhao-Heng and Wang, Changhao and Pineda, Luis and Bodduluri, Krishna and Wu, Tingfan and Abbeel, Pieter and Mukadam, Mustafa},
  journal={arXiv preprint arXiv:2503.07541},
  year={2025}
}

@article{han2026dexhil,
  title={{DexHiL}: A Human-in-the-Loop Framework for Vision-Language-Action Model Post-Training in Dexterous Manipulation},
  author={Han, Yifan and Chen, Zhongxi and Zhao, Yuxuan and Xu, Congsheng and Shao, Yanming and Peng, Yichuan and Mu, Yao and Lian, Wenzhao},
  journal={arXiv preprint arXiv:2603.09121},
  year={2026}
}

@article{mu2026deximit,
  title={{DexImit}: Learning Bimanual Dexterous Manipulation from Monocular Human Videos},
  author={Mu, Juncheng and Yang, Sizhe and Bao, Yiming and Bae, Hojin and Wei, Tianming and Xu, Linning and Li, Boyi and Xu, Huazhe and Pang, Jiangmiao},
  journal={arXiv preprint arXiv:2602.10105},
  year={2026}
}

@article{bai2025qwen25vl,
  title={{Qwen2.5-VL} Technical Report},
  author={Bai, Shuai and Chen, Keqin and Liu, Xuejing and Wang, Jialin and Ge, Wenbin and Song, Sibo and Dang, Kai and Wang, Peng and Wang, Shijie and Tang, Jun and others},
  journal={arXiv preprint arXiv:2502.13923},
  year={2025}
}

@article{wu2024deepseekvl2,
  title={{DeepSeek-VL2}: Mixture-of-Experts Vision-Language Models for Advanced Multimodal Understanding},
  author={Wu, Zhiyu and Chen, Xiaokang and Pan, Zizheng and Liu, Xingchao and Liu, Wen and Dai, Damai and Gao, Huazuo and Ma, Yiyang and Wu, Chengyue and Wang, Bingxuan and others},
  journal={arXiv preprint arXiv:2412.10302},
  year={2024}
}

@inproceedings{qi2017pointnetplusplus,
  title={{PointNet++}: Deep Hierarchical Feature Learning on Point Sets in a Metric Space},
  author={Qi, Charles R. and Yi, Li and Su, Hao and Guibas, Leonidas J.},
  booktitle={Advances in Neural Information Processing Systems},
  year={2017}
}
